\documentclass[10pt,twocolumn,letterpaper]{article}

\usepackage[pagenumbers]{iccv} 

\newcommand{\btheta}{\boldsymbol{\theta}}
\newcommand{\bGamma}{\boldsymbol{\Gamma}}
\newcommand{\bPhi}{\boldsymbol{\Phi}}
\newcommand{\bbeta}{\boldsymbol{\beta}}
\newcommand{\bgamma}{\boldsymbol{\gamma}}

\usepackage{siunitx}

\definecolor{iccvblue}{rgb}{0.21,0.49,0.74}
\usepackage[pagebackref,breaklinks,colorlinks,allcolors=iccvblue]{hyperref}
\usepackage{amsmath}
\usepackage{amssymb}
\usepackage{bm}
\usepackage[table,xcdraw]{xcolor}
\definecolor{best_color}{RGB}{253,155,154}
\definecolor{second_color}{RGB}{254,205,158}
\definecolor{third_color}{RGB}{255,255,163} 

\usepackage{newtxtext,newtxmath}

\def\paperID{6644} 
\def\confName{ICCV}
\def\confYear{2025}

\title{Demeter: A Parametric Model of Crop Plant Morphology from the Real World}

\author{Tianhang Cheng, Akbert Zhai, Evan Z. Chen, Rui Zhou, Yawen Deng, Zitong Li\
\and
Kejie Zhao, Janice Shiu, Qianyu Zhao, Yide Xu, Xinlei Wang, Yuan Shen, Sheng Wang\
\and
Lisa Ainsworth, Kaiyu Guan, Shenlong Wang\textsuperscript{†}
\\
\vspace{5pt}
University of Illinois Urbana-Champaign
}

\begin{document}

\twocolumn[{
\renewcommand\twocolumn[1][]{#1}
\maketitle
\begin{center}
     \includegraphics[width=0.98\textwidth, trim=45 0 45 0, clip]{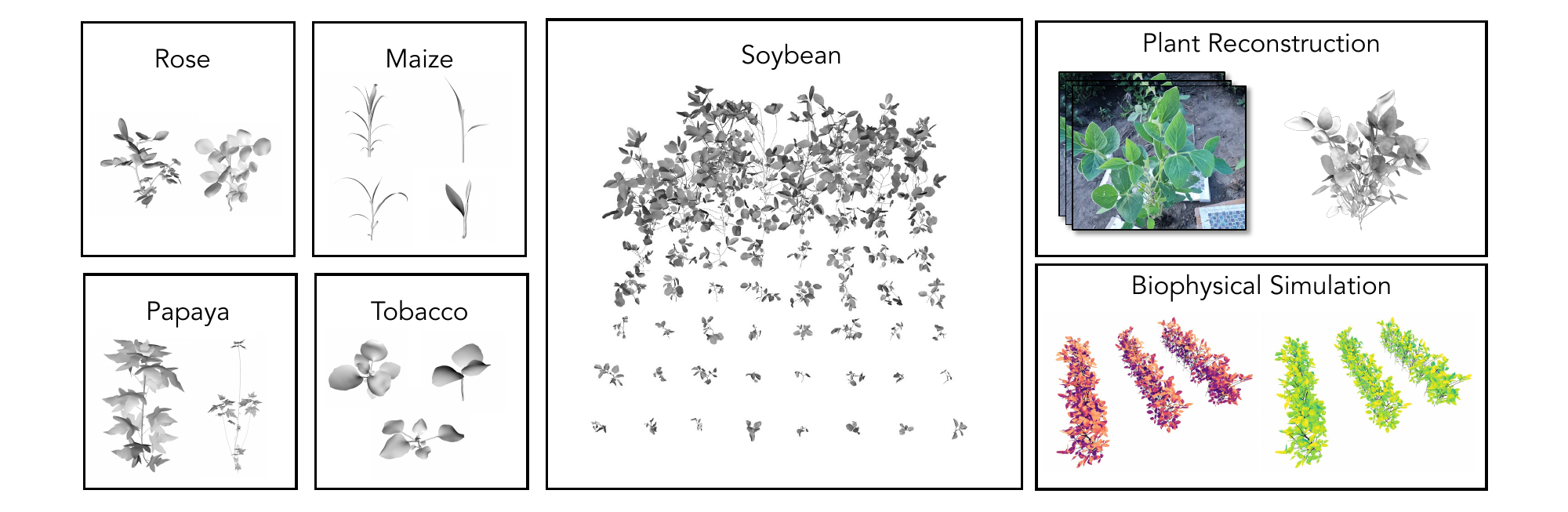}
    \captionof{figure}{
    {\bf Demeter} is a parametric model that is trained on real-world data. It encodes plants into shape, topology, articulation, and deformations, providing a realistic, compact representation that can generalize to many species. 
    }
    \label{fig:teaser}
\end{center}

}]

\begin{abstract}
Learning 3D parametric shape models of objects has gained popularity in vision and graphics and has showed broad utility in 3D reconstruction, generation, understanding, and simulation. 
While powerful models exist for humans and animals, equally expressive approaches for modeling plants are lacking. 
In this work, we present Demeter, a data-driven parametric model that encodes key factors of a plant morphology, including topology, shape, articulation, and deformation into a compact learned representation. 
Unlike previous parametric models, Demeter handles varying shape topology across various species and models three sources of shape variation: articulation, subcomponent shape variation, and non-rigid deformation. 
To advance crop plant modeling, we collected a large-scale, ground-truthed dataset from a soybean farm as a testbed. 
Experiments show that Demeter effectively synthesizes shapes, reconstructs structures, and simulates biophysical processes. 
Code and data is available at our \href{https://Tianhang-Cheng.github.io/Demeter}{project page}.
\end{abstract}

\section{Introduction}
\label{sec:intro}
\begin{figure*}[t]
\centering
\def\arraystretch{0.3}
\setlength{\tabcolsep}{0.5pt}
\resizebox{.85\linewidth}{!}{
\begin{tabular}{ccccc}
&{\small ({\bf a}) + Topology ($\bGamma$)}
&{\small ({\bf b}) $\boldsymbol\Gamma$ + Articulation ($\btheta$)}
&{\small ({\bf c}) $\boldsymbol\Gamma, \boldsymbol\theta$ + Shape ($\bbeta$)}
&{\small ({\bf d}) $\boldsymbol\Gamma, \boldsymbol\theta, \bbeta$ + Deformation($\bgamma$)}

\\
&\includegraphics[width=0.25\linewidth]{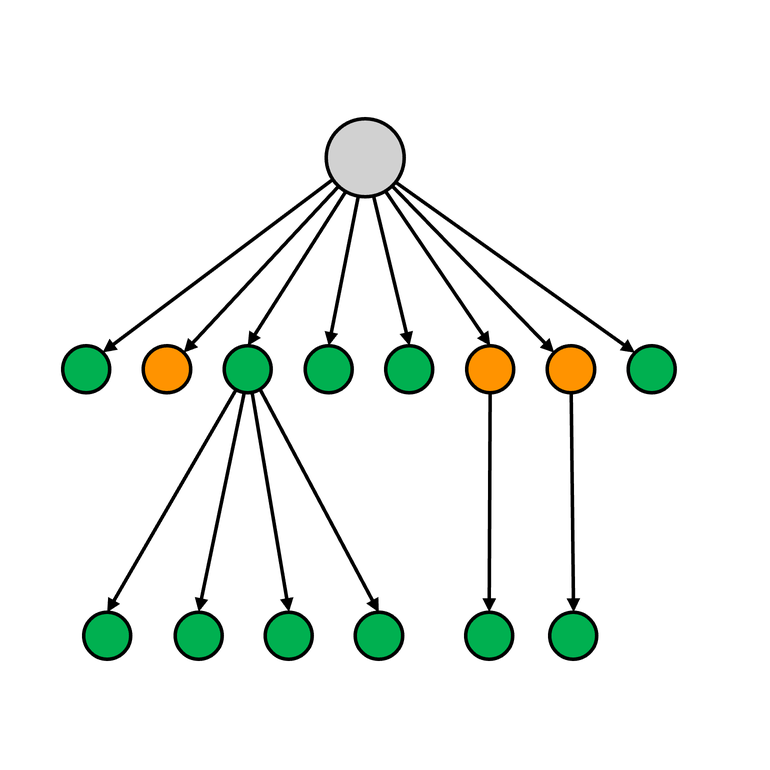}
&\includegraphics[width=0.25\linewidth]{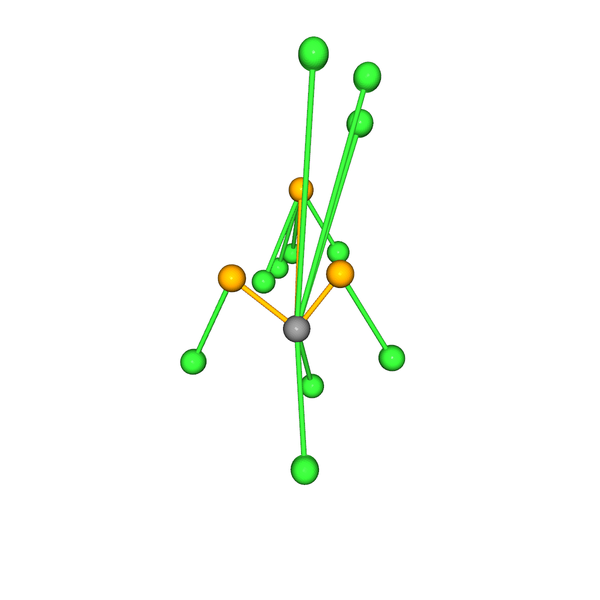}
&\includegraphics[width=0.25\linewidth]{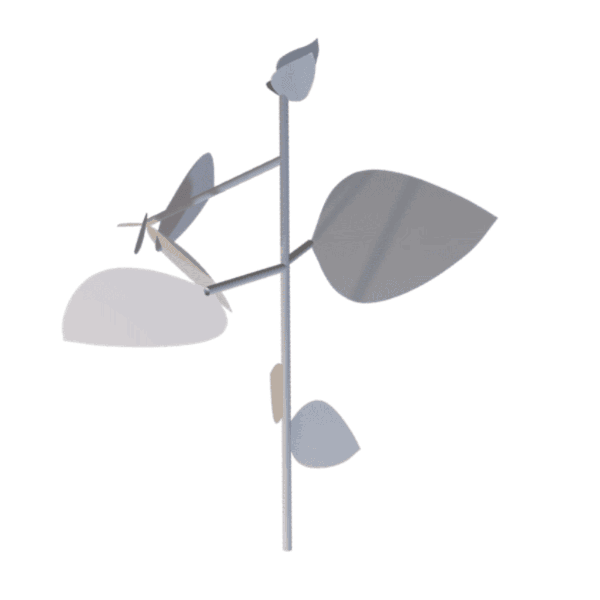}
&\includegraphics[width=0.25\linewidth]{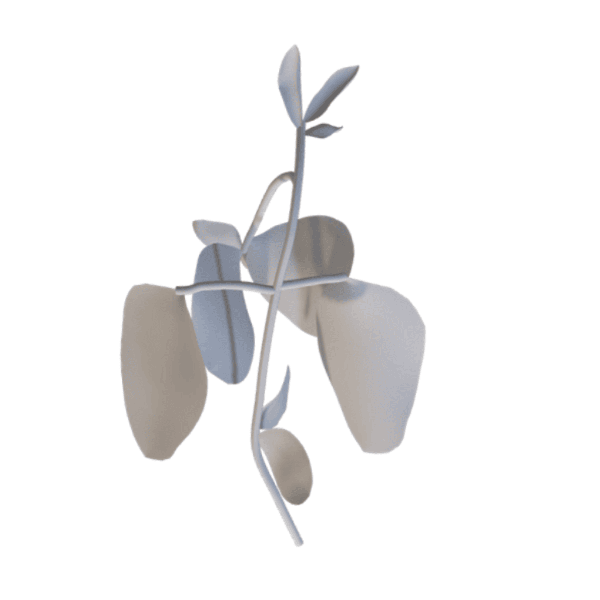}
\end{tabular}
}

\caption{The \textbf{Demeter Model} consists of four parametric components: (a) topology $\bGamma$, a tree-structured graph that stores the binary connection relationship between plant-part nodes; (b) articulation parameters $\btheta$ defining each node’s relative position to its parent, using quaternions for rotation and 1D translation; (c) shape parameters $\bbeta$, representing leaf variance via principal component coefficients, learned from 2D leaf scans; (d) deformation parameters $\bgamma$, learned from 3D leaf point clouds and fixed 2D shape parameters $\bbeta$, controlling the 3D skeletons of leaf and stem. Here the \textcolor{gray}{gray}, \textcolor{orange}{orange} and \textcolor{green}{green} nodes represents root, stem and leaf respectively.
}
\label{fig:params}
\end{figure*}

Crop plants are essential to life on Earth and human well-being. 
Food crops such as rice, maize, and soybeans form the foundation of global agriculture and feed billions of people; industrial crops supply crucial raw materials for manufacturing; tree crops produce oxygen, provide habitats, and help maintain ecological balance.
Reconstructing, understanding, and modeling plants with computer vision opens pathways to enhance crop yields, monitor environmental health, and drive agricultural innovation. A critical step toward this goal is establishing a realistic, flexible, expressive, compact, and preferably data-driven morphological model for plants. While powerful parametric models~\cite{loper2023smpl, li2017learning, zuffi2024varen} exist for human forms and quadrupedal animals,
there is currently no equally expressive model for plants.

Our goal is to bridge the gap by building a compact parametric model capable of representing, reconstructing, animating, and simulating plants of a given species. 
To this end, we present {\bf Demeter}, a novel, data-driven, category-based parametric model that encodes key factors driving variations in 3D plant morphology, including topology, articulation, shape variations, and non-articulated deformations (Fig. ~\ref{fig:params}). 
The articulations of the plant body are modeled in a kinematic tree structure, with nodes representing elements of the plant and edges encoding joint angles that define the relative orientation between connected segments. 
We represent shapes and deformations of each leaf and stem using triangular meshes to ease reconstruction and simulation, where vertex locations are parameterized by Catmull-Rom spline curves. Although these shape and deformation parameters are high-dimensional, we observe—similar to findings in human and animal modeling—that their variations lie primarily in low-dimensional manifolds. Inspired by this, we fit and learn a compact linear space using principal components, significantly enhancing the model's compactness without compromising expressiveness. 

 


The quality of a learned shape model depends on the scale and representativeness of our training data. To capture real-world variations, we collected {\bf 3D soybean} samples from a farm in Illinois over an entire growing season, capturing around 600 plants from multiple genotypes across all life stages—from germination to pod formation. Unlike lab environments, our row-crop farm samples minimize environmental discrepancies. We performed dense 3D reconstructions using a Gaussian splatting pipeline, manually corrected missing parts, and selected over 300 plants for detailed leaf and stem annotation. These annotated reconstructions trained the Demeter model for soybeans. Additionally, we used this dataset to develop benchmarks for 2D/3D semantic segmentation and 3D morphological shape reconstruction from images or point clouds.

We train Demeter for various species across different 2D and 3D datasets, including the aforementioned new soybean data. We rigorously evaluate the learned Demeter both qualitatively and quantitatively, testing its efficacy on tasks such as multi-view and monocular reconstruction, as well as downstream applications like photosynthesis and stomatal conductance simulations. 
Accurate plant morphology, as modeled by Demeter, enables detailed simulations for analyzing crop productivity, demonstrating its practical impact. 
Unlike conventional 3D models, Demeter provides a complete parametric morphological model, supporting diverse biophysical simulations and advancing scientific discovery. 
To foster collaboration, we will open-source a Demeter-based platform, inviting vision and plant phenotyping researchers to create and share models of various species.

Our contributions are: 1) We introduce Demeter, a data-driven parametric model for plant morphology. 2) We demonstrate Demeter's capabilities in reconstruction, and simulation for multiple crop plant species. 3) We present a new dataset with over 300 soybeans, featuring 3D geometry, part segmentation, and topology annotations.


\begin{figure*}[t]
\centering
\def\arraystretch{0.3}
\setlength{\tabcolsep}{0.5pt}

\includegraphics[height=0.26\linewidth]{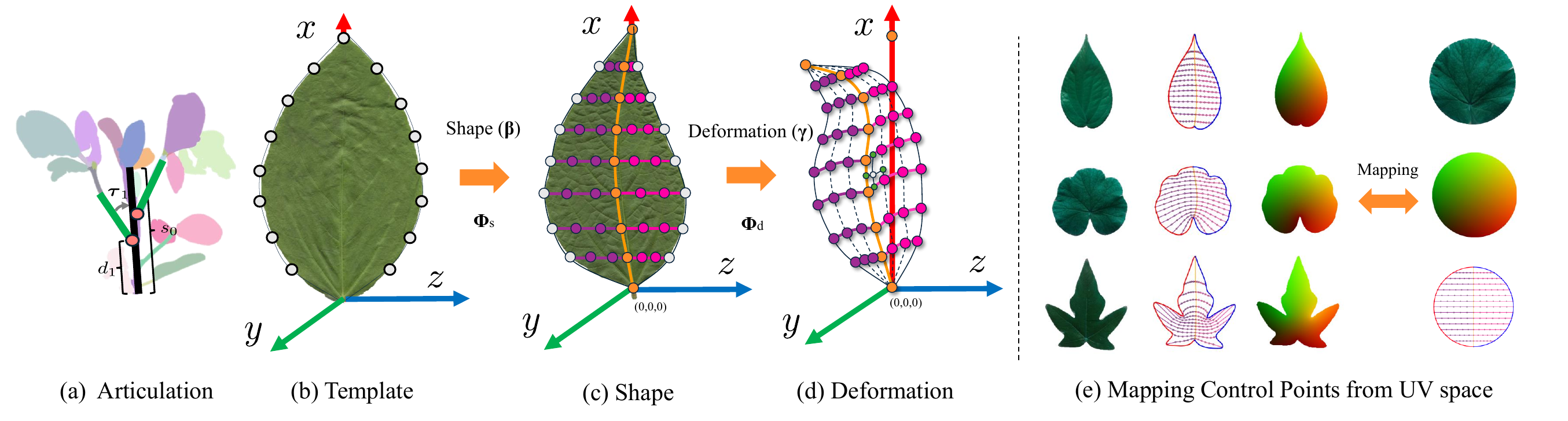}
\caption{\textbf{Demeter model parameters}. (a) Articulation includes scale $s$, which controls the size of the component (e.g., stem length); position $d$, which controls the location of the joint node relative to its parent; and quaternion angle $\boldsymbol{\tau}$, which controls the relative rotation. (b) The mean template learned from 2D leaf scannings. (c) The shape parameter encodes the location of contour points, which defines the inner points of the leaf and serve as joints to deform the leaf and stem. (d) Deformation $\bgamma$ represents a grid of Catmull-Rom control points to manage shape and articulation-independent deformation, such as leaf curling. (e) An illustration of the mapping between the leaf shape control points and canonical UV space.
}
\label{fig:details}
\end{figure*}

\section{Related Works}
\label{sec:related}



\paragraph{Procedural Shape} Procedural models aim to represent geometries by a series of programmatic operations 
to construct the geometry. They have been used extensively in computer graphics to generate building exteriors~\cite{wonka2003instant, muller2006procedural, larive2006wall, muller2007image} and interiors~\cite{lopes2010constrained, marson2010automatic, deitke2022procthor, raistrick2024infinigen}, natural landscapes~\cite{kamal2007parametrically, miller1986definition, belhadj2005modeling, raistrick2023infinite}, and entire cities~\cite{parish2001procedural, kelly2007citygen}. Most existing work on plant models is hand-crafted or procedural, incorporating biological knowledge to simulate the morphology of leaf~\cite{kim2017procedural, miao2013framework, lu2009venation, sun2021wrinkle, wang2025autoregressive} or the real growth of the plant~\cite{prusinkiewicz1999system, lintermann1999interactive, deussen1998realistic, prusinkiewicz2012algorithmic, zhou2023deeptree, lee2023latent}.
While these models yield realistic 3D outputs, their complex symbol-geometry relationship makes inverse modeling (e.g., estimating model parameters from images) challenging. In contrast, our model uses simple geometric transformations of parts (leaves and stems), making it easier to fit sensor observations.

\vspace{-8pt}
\paragraph{Learning-based Parametric Shape}
To overcome the limitations of rule-based models and enhance expressiveness, researchers have used statistical learning to derive parametric models from real-world data, notably in the study of human body models~\cite{allen2003space, allen2006learning, anguelov2005scape, chen2013tensor, hasler2009statistical, loper2023smpl, pavlakos2019expressive, xu2020ghum}. 
The most representative work is SMPL~\cite{loper2023smpl}, which inspired our use of learned linear model. Unlike SMPL, which has a global dependency issue, Demeter learns separate, independent local shapes, offering greater flexibility to handle intra-species variations. The success of learned parametric models has also expanded to domains such as human faces~\cite{li2017learning}, hands~\cite{romero2022embodied}, quadrupedal animals~\cite{deane2021dynadog+, zuffi20173d, zuffi2024varen, cashman2012shape} and leaf~\cite{bradley2013image}. Nevertheless, most current models focus on fixed topology and have not been applied to plant, which is still dominated by procedural or grammar-based models. 
Recent works~\cite{chaurasia2017editable,bradley2013image, hadadi2025procedural, jeong2013simulation} starts to build on the parametric models of leaves but fail to model whole plants, or do not disentangle shape and deformation.
Demeter aims to close this gap, allowing plant morphology to benefit from real-world data and statistical learning.

\vspace{-8pt}
\paragraph{Inverse Parametric Shape Modeling}
Methods for reconstructing complete 3D plant shapes generally rely on inverse parametric modeling, especially inverse procedural modeling (IPM). IPM relies on fitting parameters of a procedural model based on structural similarity with the input observations. This approach has found success in fitting point cloud data~\cite{livny2010automatic, stava2014inverse, Hackenberg2015SimpleTreeE, Guo2020RealisticPP} and interactive user inputs~\cite{lintermann1999interactive, longay2012treesketch, Manfredi2022TreeSketchNetFS}. To tackle reconstruction from single images, Li et al.~\cite{li2021learning} and SVDTree~\cite{Li2024SVDTreeSV} train models (on procedurally generated data) to predict bounding volumes and then fill the volumes using space colonization~\cite{palubicki2009self}. These works focus on tree plants, and often ignore fine-grained leaf details. With Demeter, we aim to provide a plant morphology model that can accurately represent many kinds of variation in plant shapes while being conducive to inverse modeling due to its data-driven, primitive-based structure.

\vspace{-8pt}
\paragraph{General 3D Reconstruction on Plants} A number of works have investigated the use of generic reconstruction methods to model plant shapes. The methods used include multi-view stereo~\cite{wu2020mvs, zhu2020analysing, wang2022three, Wu2022AMP, sun2023soybean} and neural radiance fields~\cite{smitt2023pag, saeed2023peanutnerf, hu2023high, arshad2024evaluating}. 
However, without incorporating shape priors, such methods struggle outside clean lab settings due to plants' thin structures and heavy occlusion. Demeter incorporates prior knowledge and data-driven models to constrain the solution space for 3D reconstruction.


\section{Method}
\label{sec:method}

Our parametric model, Demeter, is illustrated in Fig.~\ref{fig:params}. 
Like parametric models for humans and animals~\cite{loper2023smpl, zuffi2024varen}, 
it decomposes the mesh into shape vectors and deformation vectors. However, Demeter additionally encodes the variable structural topology of plant components. We fit species-specific Demeter models from real-world 3D scans and learn linear bases of shape and deformation. 
In the following sections, we detail the formulation and parameterization of Demeter (Sec.~\ref{sec:method_modeling}), describe the procedure for learning Demeter-Soybean from real-world data (Sec.~\ref{sec:method_learning}), and finally demonstrate how to use learned Demeter models for image-based and point-based 3D reconstruction (Sec.~\ref{sec:method_inference}). 


\subsection{Demeter Plant Model} 
\label{sec:method_modeling}

Demeter is a parametric model that represents a plant's shape as a factorized primitive graph with a set of parameters that uniquely depict its vegetative morphology:
\begin{equation}
(\mathbf{V}, \mathbf{F}) = \mathcal{M_{\mathbf{\Phi}}}(\bGamma, \btheta, \bbeta, \bgamma),
\end{equation}
where $(\mathbf{V}, \mathbf{F})$ are the output vertices and faces, respectively; $\bGamma, \btheta, \bbeta, \bgamma$ represent topology, articulation, shape, and non-articulated deformation, respectively. $\bPhi = \{ {\bPhi_s, \bPhi_d} \}$ is a learned compact PCA basis about shape $\bbeta$ and deformation $\bgamma$ for each species. In essence, Demeter is a function $\mathcal{M}: \mathbb{Z}_2^{|\bGamma|} \times \mathbb{R}^{|\btheta| \times |\bbeta| \times |\bgamma|} \rightarrow \mathbb{R}^{3 \times |\mathbf{V}|} \times \mathbb{Z}^{3 \times |\mathbf{F}|}$ that maps a set of parameters to a triangulated plant mesh.  Specifically, given a Demeter model $\mathcal{M}$, we calculate vertices of the plant using the following equation:

\begin{equation}
\label{eq:main_forward}
{{\bf{v}}_f} = {\rm{T}}(\boldsymbol\theta ;{\rm{ }}\boldsymbol\Gamma) \cdot {\cal D}({\bf{v}}_t +{\cal S}(\boldsymbol\beta ); \boldsymbol\gamma )
\end{equation}
where $\mathbf{v}_t$ is the vertices location of the templates, learned from the mean shape of each organ;  $\bf{v}_f \in V$ is the final location of the deformed plant; $\mathcal{S}(\boldsymbol\beta)$ represents the shape (grid points) offset of each part that $\mathbf{v}_t$ resides in; $\mathcal{D}(\mathbf{v};\boldsymbol\gamma)$ denotes a shape-independent deformation; $\mathrm{T}$ is composed of rigid + scale transformations along the kinematic chain defined by $\bGamma$, parameterized by the articulation $\btheta$. Next, we describe each element of Demeter in detail.





\vspace{-8pt}
\paragraph{Topology} 
Topology describes the connection relationship between plant nodes. Unlike in human or animal models, instances within the same plant species can have significantly different topology.
We represent the plant topology $\bGamma$ as a tree data structure of $n = n_l + n_s + n_o$ nodes, each representing an individual part of the plant, e.g. a leaf or stem. We currently omit other structures (e.g. flowers and fruit) for simplicity. 
Each node $i$ has a type $\mathrm{tp}(i) \in \{\mathrm{leaf}, \mathrm{stem}\}$ and is connected to its parent $\mathrm{pa}(i) \in \mathbb{Z}$ ({except the root}). 

\begin{figure*}[t]
\centering
\def\arraystretch{0.3}
\setlength{\tabcolsep}{0.5pt}
\resizebox{\linewidth}{!}{
\begin{tabular}{cccccccccccc}

&{\small Papaya}
&{\small Thevetia} 
&{\small ashanti blood}
&{\small Betel}
&{\small Geranium}
&{\small Ficus}
&{\small 3D Comp. 1}
&{\small 3D Comp. 2}
&{\small 3D Comp. 3}
&{\small 3D Comp. 7}
&{\small 3D Comp. 20} 

\\
\raisebox{10mm}{\rotatebox{90}{{\small $-2\sigma$}}}
&\includegraphics[width=0.15\linewidth]{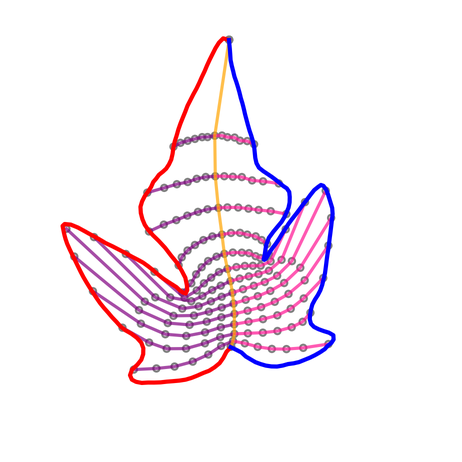}
&\includegraphics[width=0.15\linewidth]{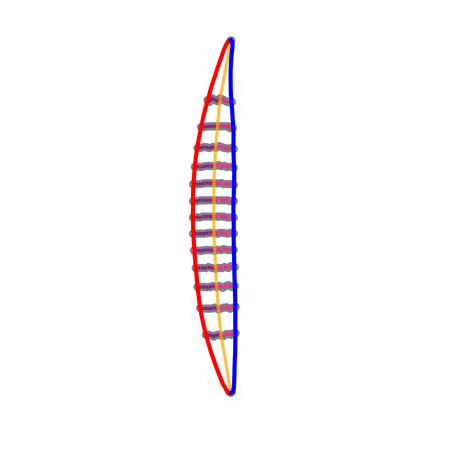}
&\includegraphics[width=0.15\linewidth]{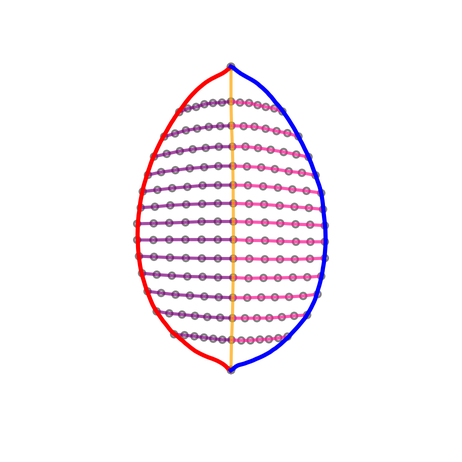}
&\includegraphics[width=0.15\linewidth]{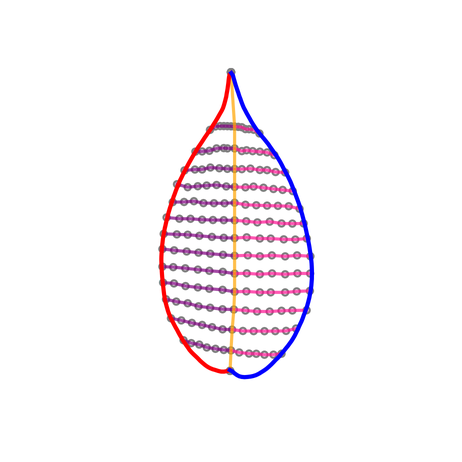}
&\includegraphics[width=0.15\linewidth]{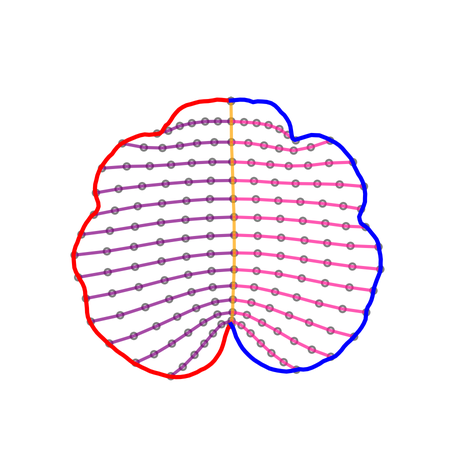}
&\includegraphics[width=0.15\linewidth]{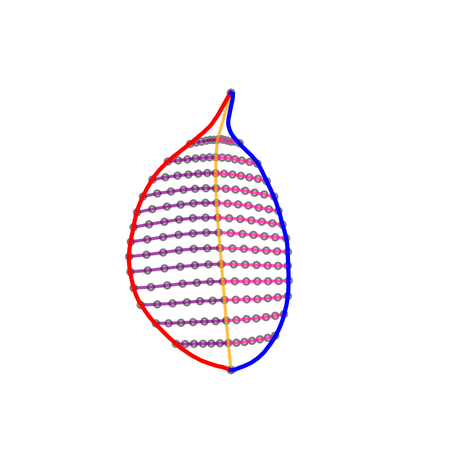}
&\includegraphics[width=0.15\linewidth]{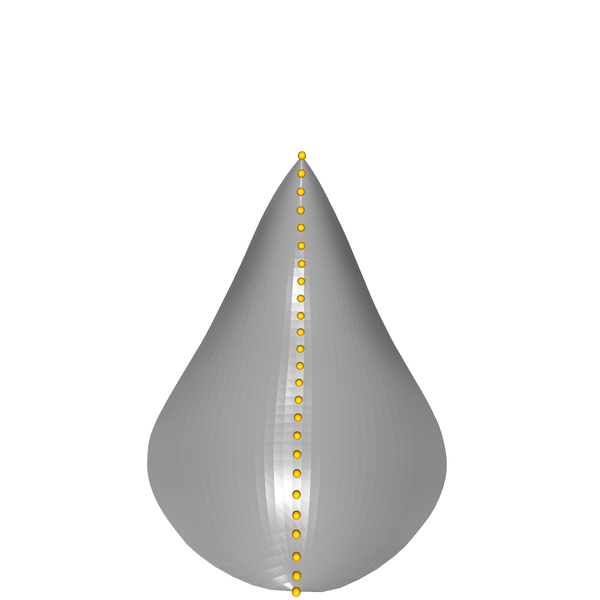}
&\includegraphics[width=0.15\linewidth]{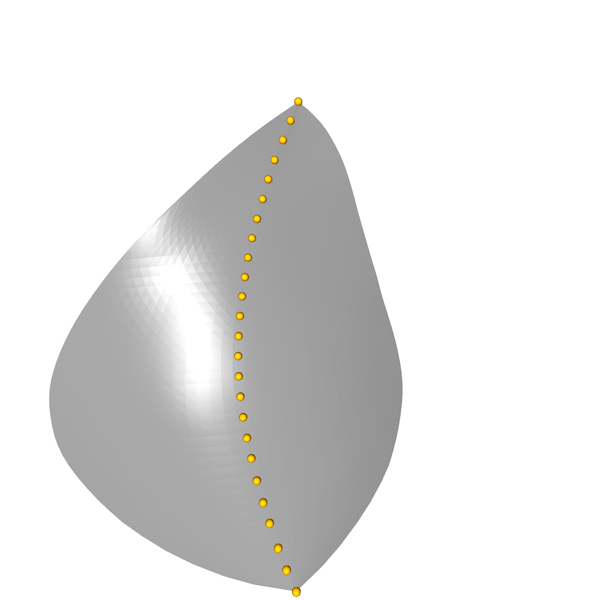}
&\includegraphics[width=0.15\linewidth]{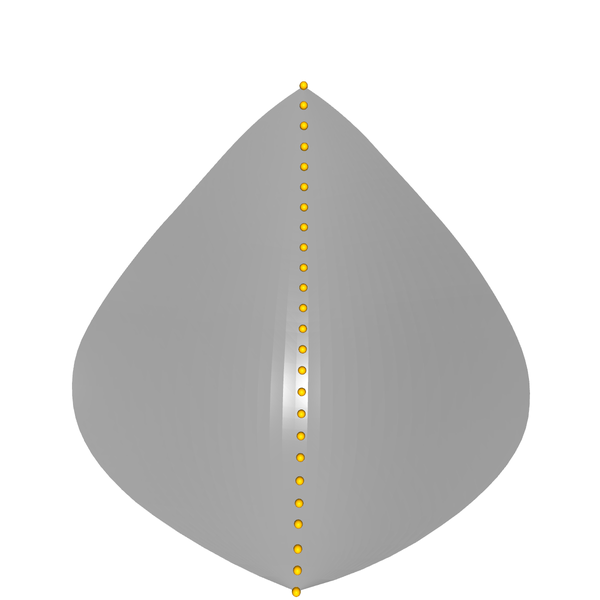}
&\includegraphics[width=0.15\linewidth]{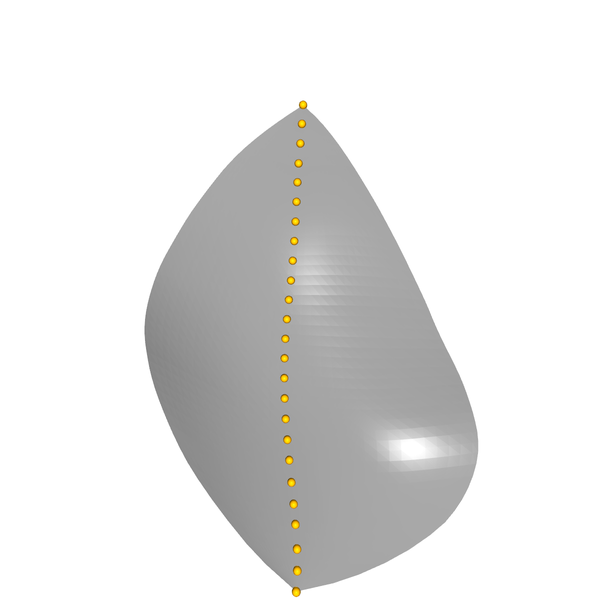}
&\includegraphics[width=0.15\linewidth]{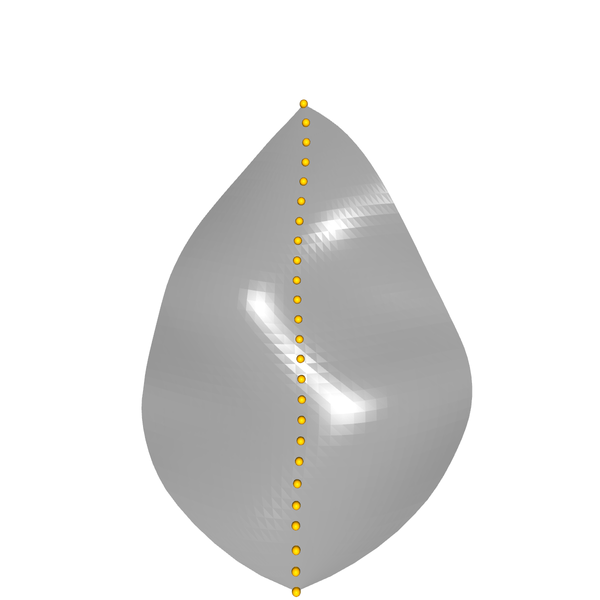}

\\
\raisebox{8mm}{\rotatebox{90}{{\small Mean Shape}}}
&\includegraphics[width=0.15\linewidth]{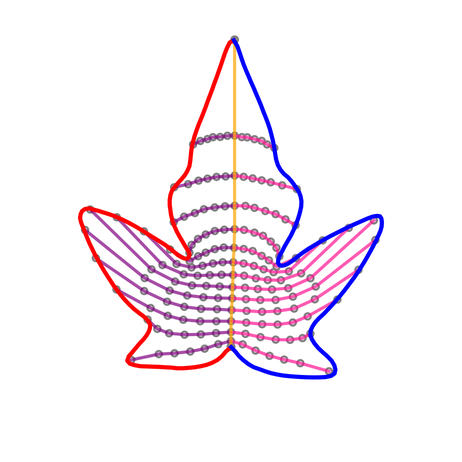}
&\includegraphics[width=0.15\linewidth]{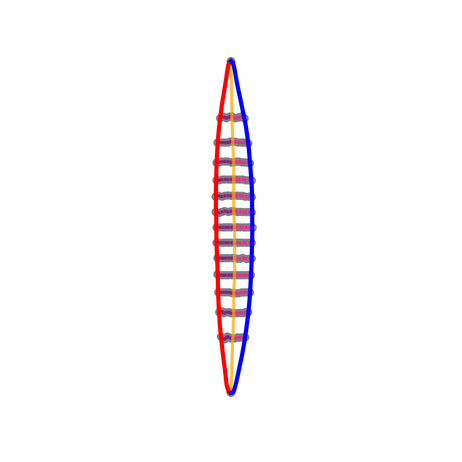}
&\includegraphics[width=0.15\linewidth]{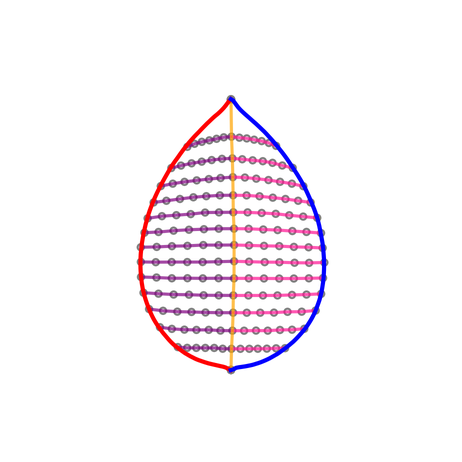}
&\includegraphics[width=0.15\linewidth]{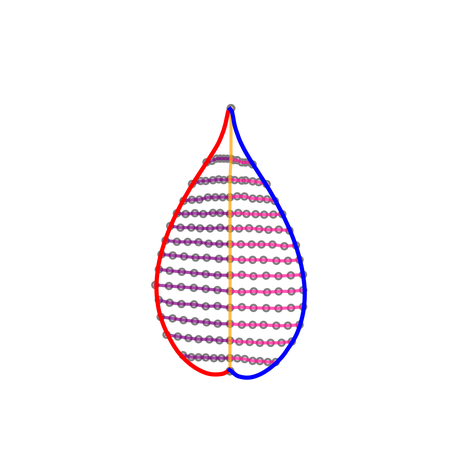}
&\includegraphics[width=0.15\linewidth]{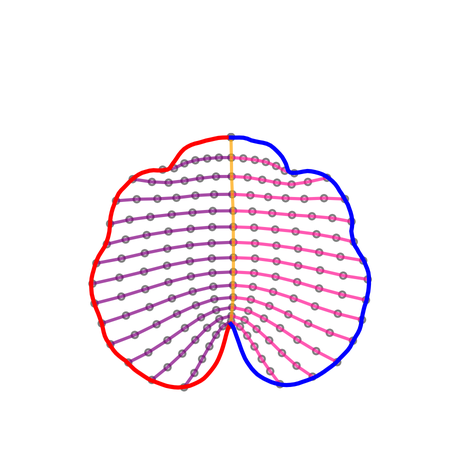}
&\includegraphics[width=0.15\linewidth]{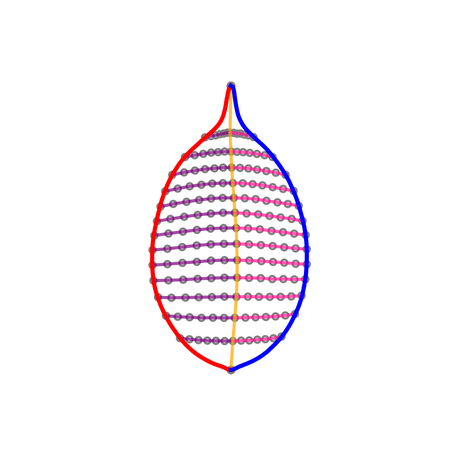}
&\includegraphics[width=0.15\linewidth]{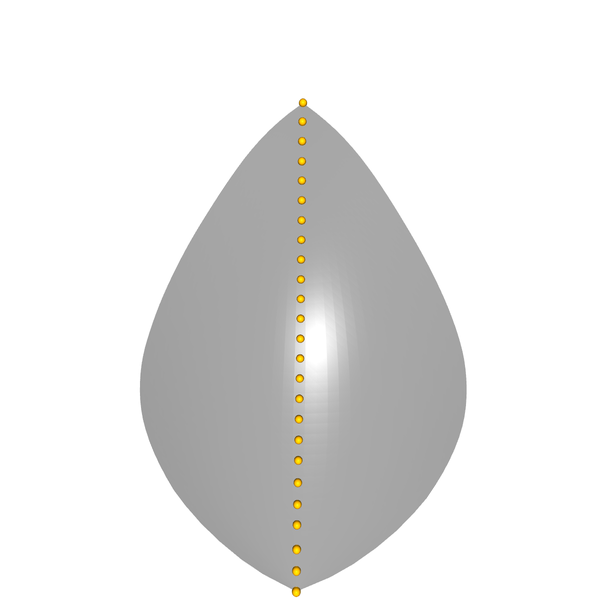}
&\includegraphics[width=0.15\linewidth]{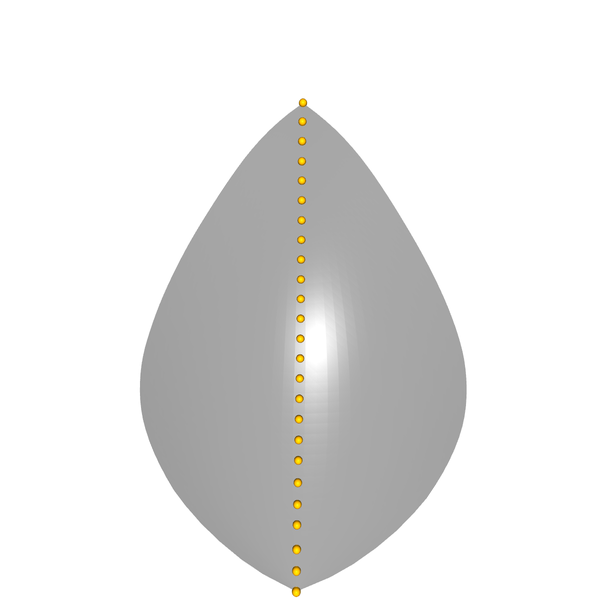}
&\includegraphics[width=0.15\linewidth]{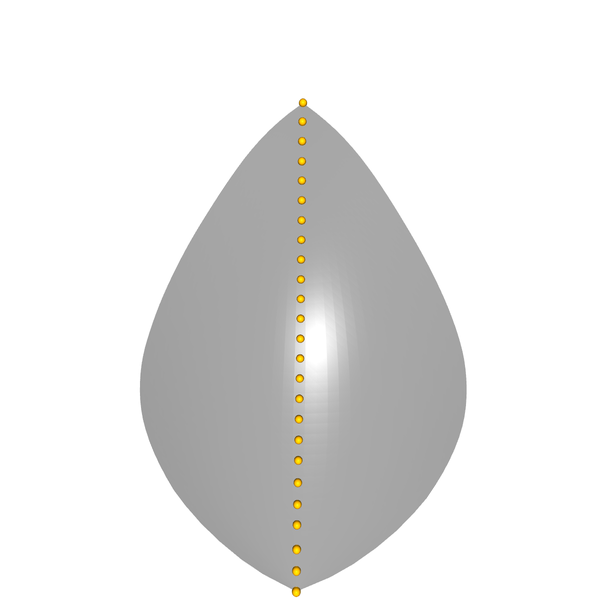}
&\includegraphics[width=0.15\linewidth]{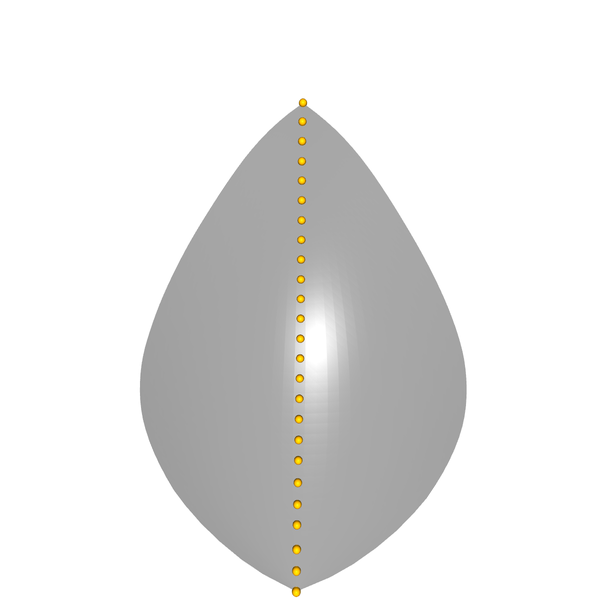}
&\includegraphics[width=0.15\linewidth]{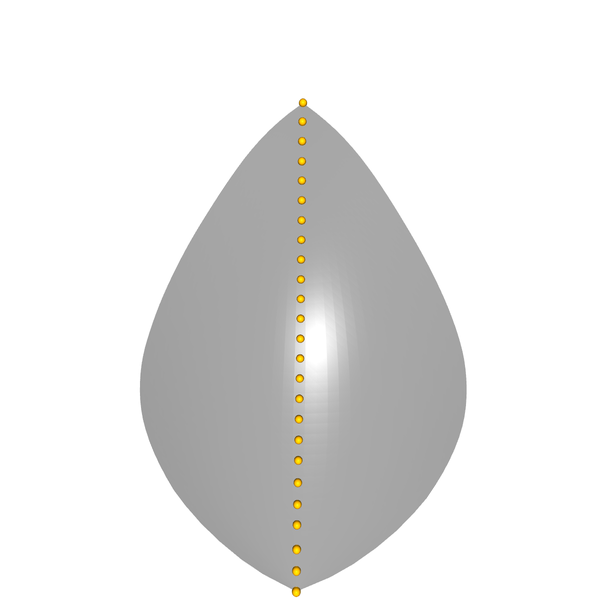}

\\
\raisebox{10mm}{\rotatebox{90}{{\small $+2\sigma$}}}
&\includegraphics[width=0.15\linewidth]{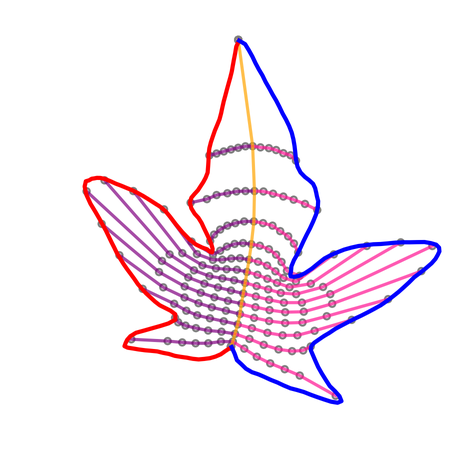}
&\includegraphics[width=0.15\linewidth]{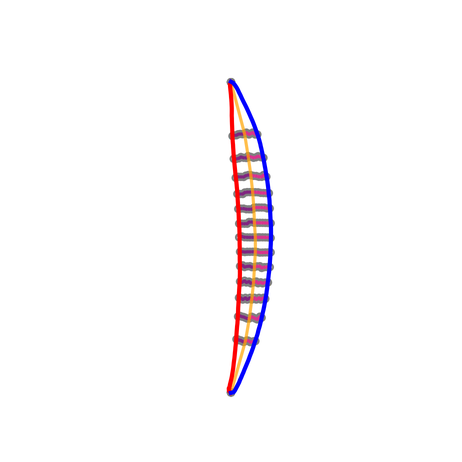}
&\includegraphics[width=0.15\linewidth]{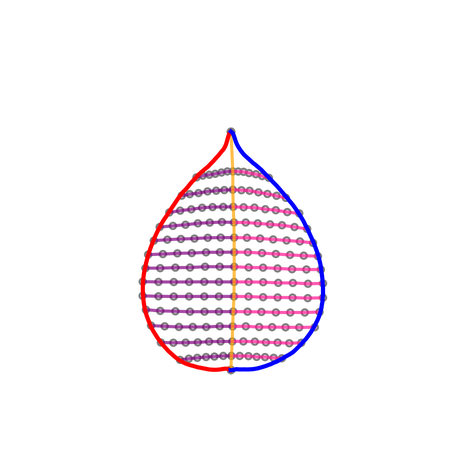}
&\includegraphics[width=0.15\linewidth]{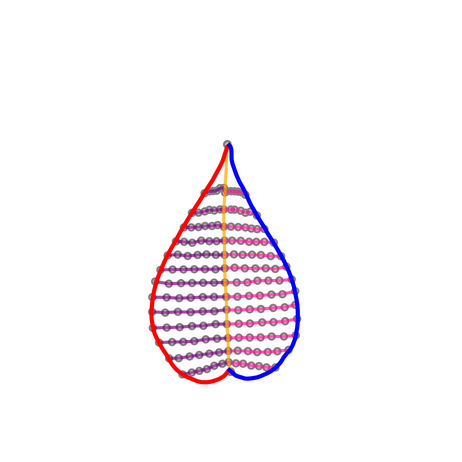}
&\includegraphics[width=0.15\linewidth]{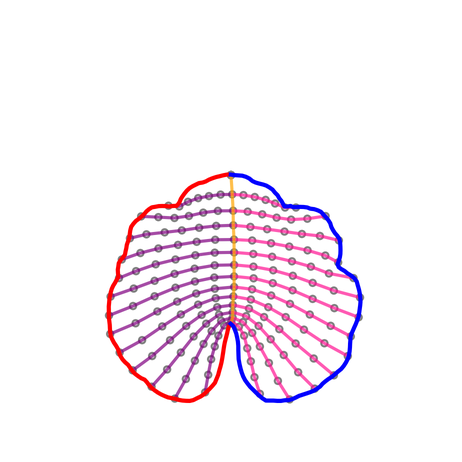}
&\includegraphics[width=0.15\linewidth]{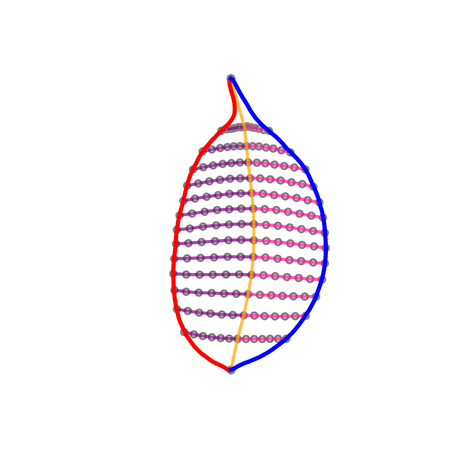}
&\includegraphics[width=0.15\linewidth]{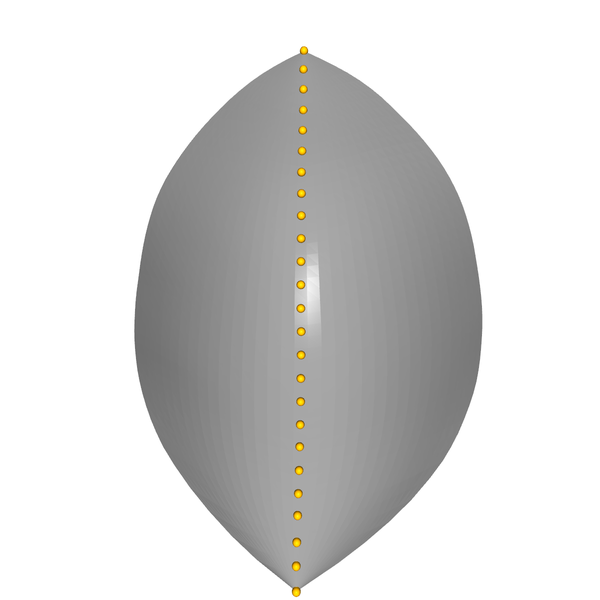}
&\includegraphics[width=0.15\linewidth]{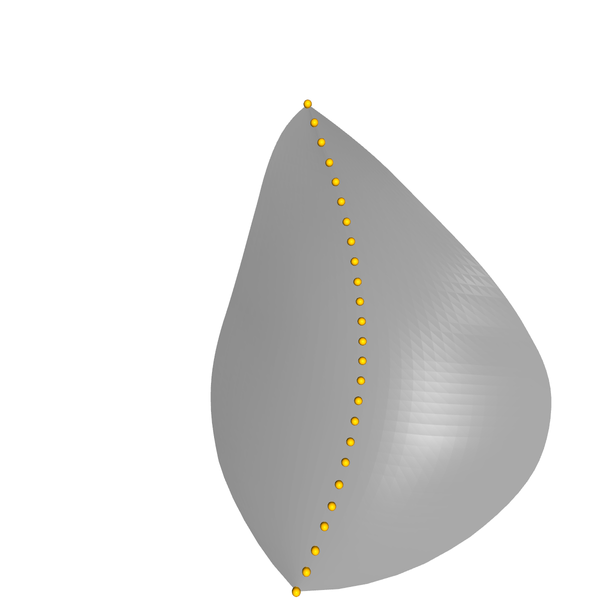}
&\includegraphics[width=0.15\linewidth]{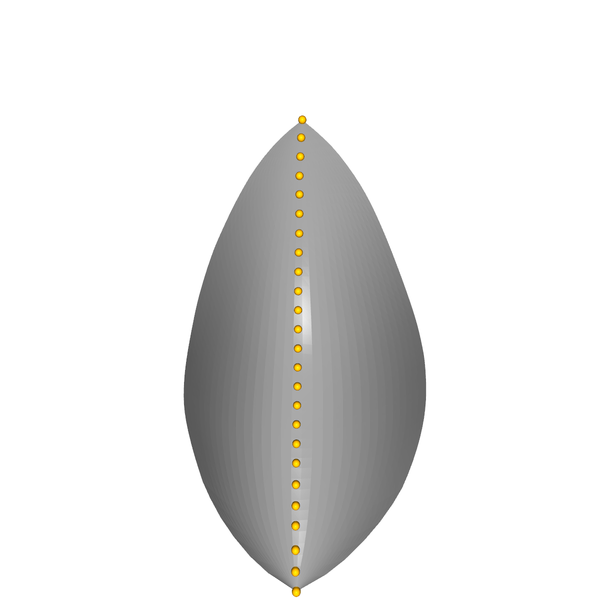}
&\includegraphics[width=0.15\linewidth]{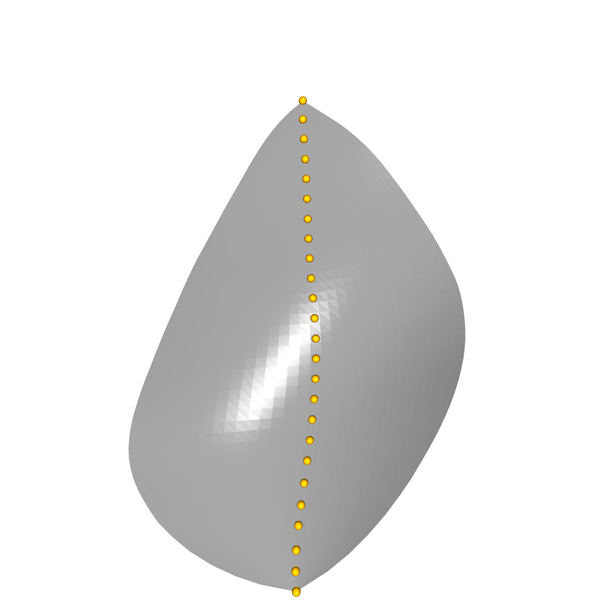}
&\includegraphics[width=0.15\linewidth]{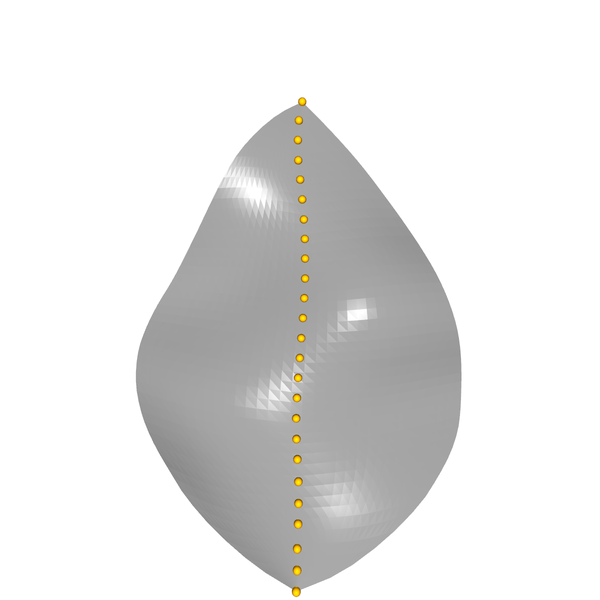}

\end{tabular}

}
\caption{\textbf{PCA coefficients of leaf shape.} We visualize the mean shape and the one of the principal components of leaf for several species in $(-2\sigma, 0, 2\sigma)$. For 3D deformation, we show the first and last few components for soybean.}


\label{fig:leaf_pca_viz}
\end{figure*}

\begin{figure}[t]
\centering
\def\arraystretch{0.3}
\setlength{\tabcolsep}{0.5pt}
\resizebox{\linewidth}{!}{
\begin{tabular}{cccccccccc}

&{\small Input Image}
&{\small One2345++\cite{liu2024one}} 
&{\small Meshy \footnote{\text{https://www.meshy.ai}} } 
&{\small Ours}

\\
\raisebox{7mm}{\rotatebox{90}{{\small Soybean}}}
&\includegraphics[width=0.33\linewidth]{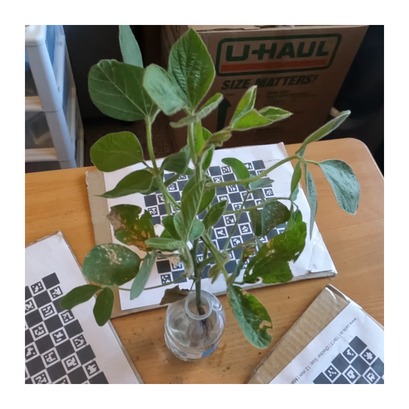}
&\includegraphics[width=0.33\linewidth]{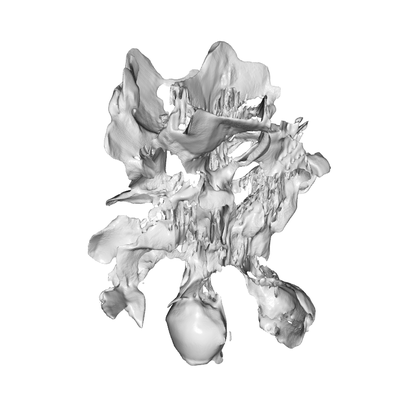}
&\includegraphics[width=0.33\linewidth]{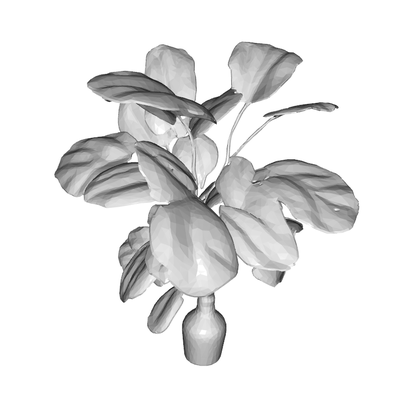}
&\includegraphics[width=0.33\linewidth]{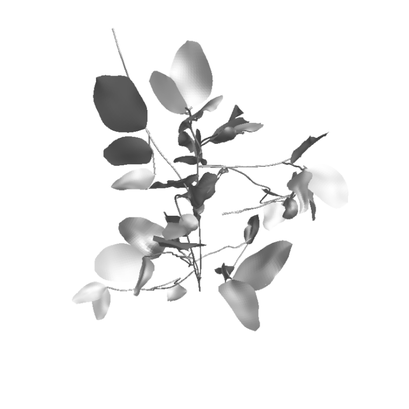}

\\
\raisebox{7mm}{\rotatebox{90}{{\small Maize}}}
&\includegraphics[width=0.29\linewidth]{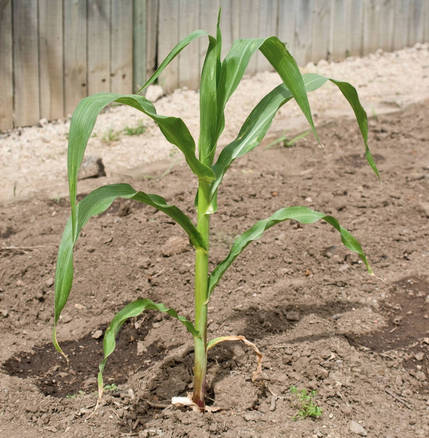}
&\includegraphics[width=0.33\linewidth]{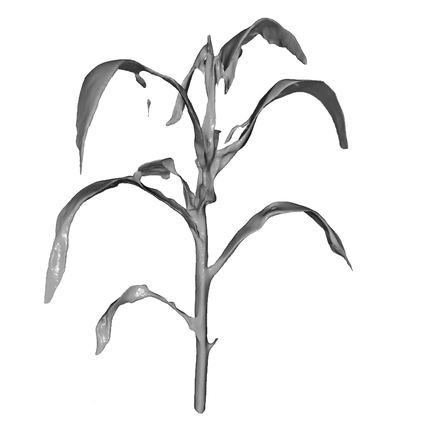}
&\includegraphics[width=0.33\linewidth]{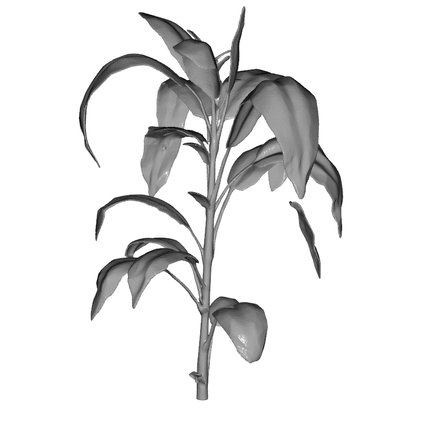}
&\includegraphics[width=0.29\linewidth]{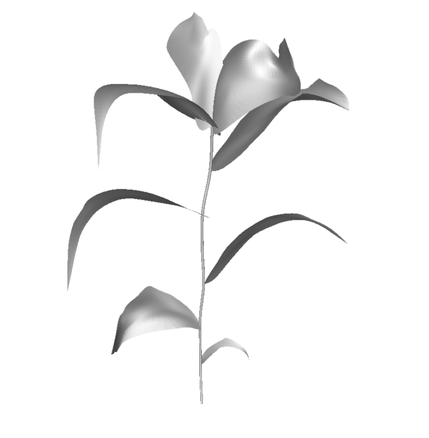}

\end{tabular}

}

\caption{\textbf{Reconstruction from single image}. 
One2345++ generates a coarse shape but fails to capture details. Meshy reconstructs a realistic plant but is misaligned with the input image. In contrast, our model is complete and faithful to input. 
}

\label{fig:2d_recon_all}
\end{figure}

\vspace{-8pt}
\paragraph{Articulation} 

Articulation describes the relative similarity transformations (rotation, translation and scale) between nodes. The articulation is represented as a set of per-node joint state parameters $\btheta =\{\btheta_i\}_{i=1}^n$, where each $\boldsymbol{\theta_i}= (\boldsymbol\tau_i, d_i, s_i)$ represents the rotation quaternion $\boldsymbol\tau_i \in \mathbb{R}^4$, path length $d_i \in [0,1]$ and scale $s_i \in \mathbb{R}^+$ of its canonical space relative to the local coordinate system of its parent stem $\mathrm{pa}(i)$, as shown in Fig.~\ref{fig:details}~(a). The root node serves as the origin. 
For vertex $\mathbf{v}$ belonging to $i$-th component, a forward kinematic chain will be computed,
in which the similarity transform $\mathbf{T}$ of each part is computed through similarity composition from the part to the root: 
\begin{equation}
\label{eq:kinematics}
\mathbf{T}_i(\btheta; \bGamma)  = \prod_{i^\prime \in \textrm{ans}(i)} \rm{T}(\boldsymbol\tau_i^\prime, d_i^\prime, s_i^\prime)
\end{equation}
where $\mathbf{T}(\boldsymbol\tau_i^\prime, d_i^\prime, s_i^\prime)$ is an similarity transform from the parent $\mathrm{pa}(i)$ to current node $i$ and $\mathrm{ans} = \{ i, \mathrm{pa}(i),\mathrm{pa}(\mathrm{pa}(i)),... \}$ denotes the ordered set of ancestor nodes of $i$ (the kinematic chain). 
Details see Fig.~\ref{fig:details}~(a).

\begin{figure}[t]
\centering
\def\arraystretch{0.3}
\setlength{\tabcolsep}{0.5pt}
\resizebox{\linewidth}{!}{
\begin{tabular}{cccccc}

&{\small Topology} 
&{\small Stem Deform}
&{\small Leaf Shape}
&{\small Leaf Deform}

\\
\raisebox{7mm}{\rotatebox{90}{{\small Positive}}}
&\includegraphics[width=0.3\linewidth]{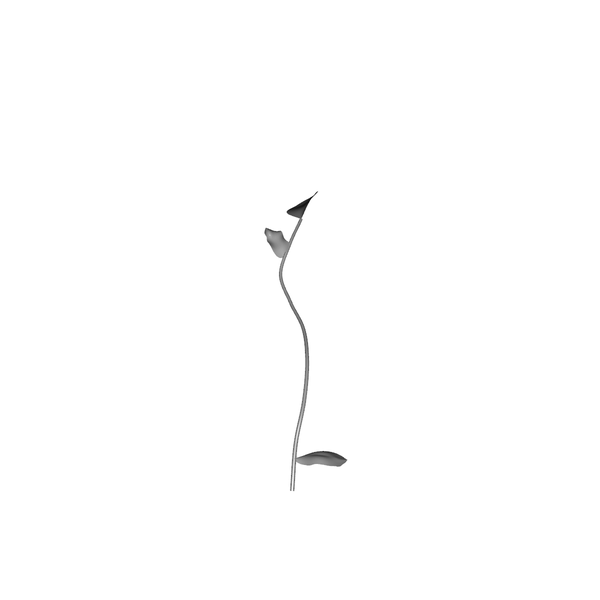}
&\includegraphics[width=0.3\linewidth]{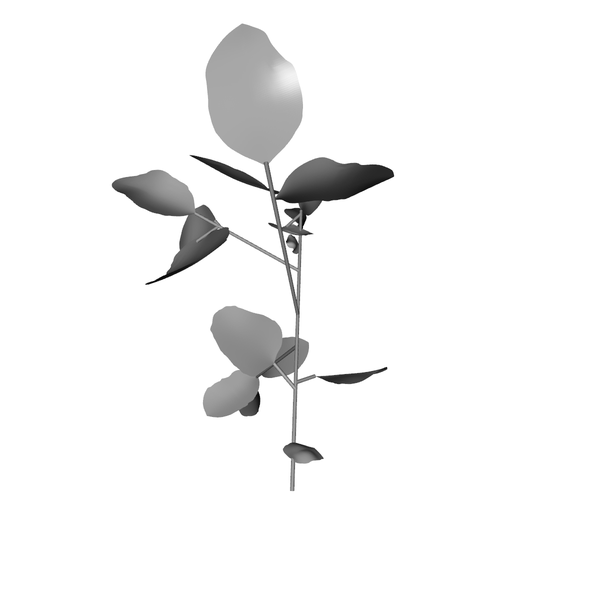}
&\includegraphics[width=0.3\linewidth]{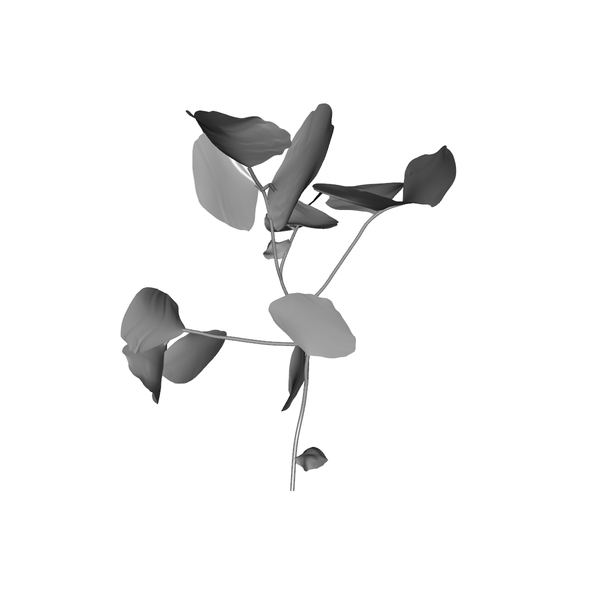}
&\includegraphics[width=0.3\linewidth]{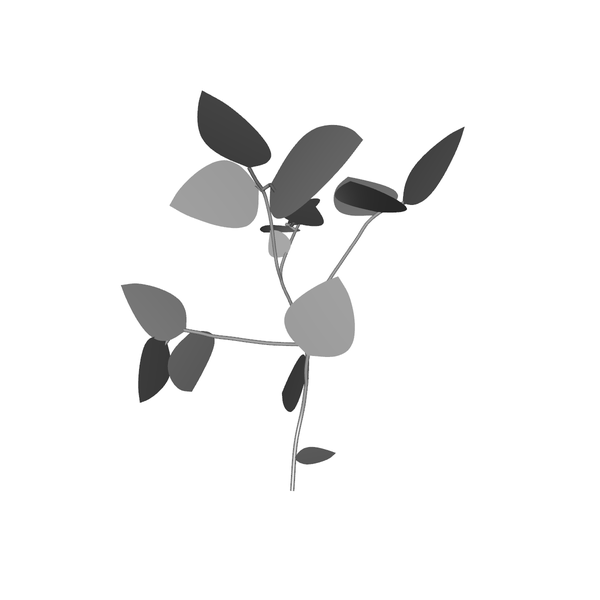}
\vspace{-8pt}

\\
\raisebox{7mm}{\rotatebox{90}{{\small Original}}}
&\includegraphics[width=0.3\linewidth]{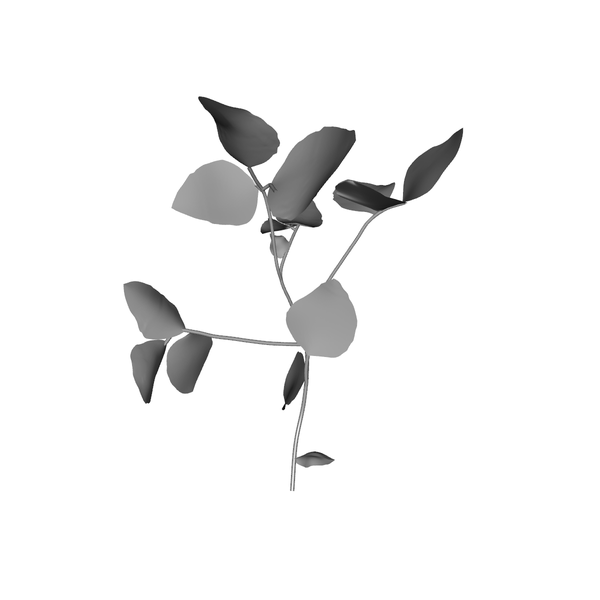}
&\includegraphics[width=0.3\linewidth]{figure/interpolate/new/ref.png}
&\includegraphics[width=0.3\linewidth]{figure/interpolate/new/ref.png}
&\includegraphics[width=0.3\linewidth]{figure/interpolate/new/ref.png}
\vspace{-8pt}
\\
\raisebox{7mm}{\rotatebox{90}{{\small Negative}}}
&\includegraphics[width=0.3\linewidth]{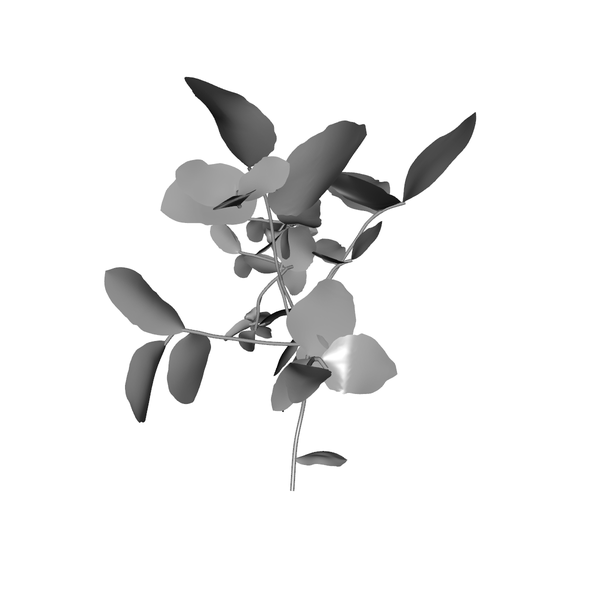}
&\includegraphics[width=0.3\linewidth]{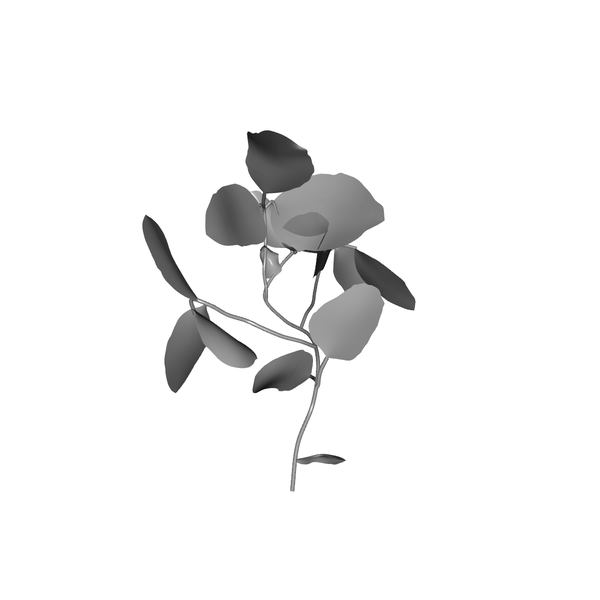}
&\includegraphics[width=0.3\linewidth]{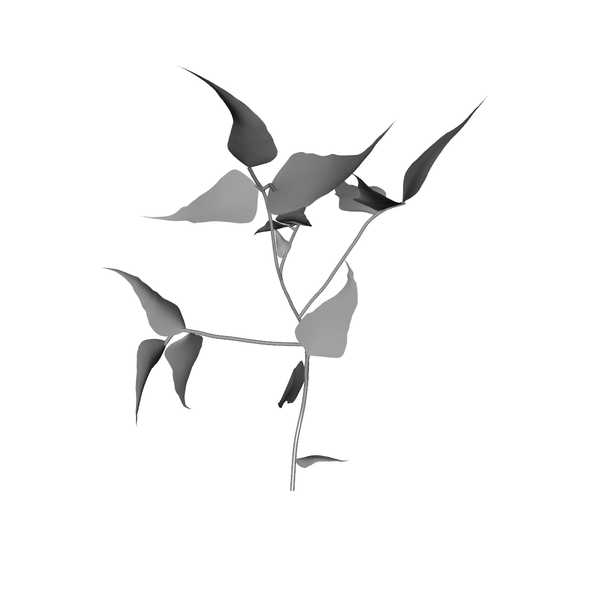}
&\includegraphics[width=0.3\linewidth]{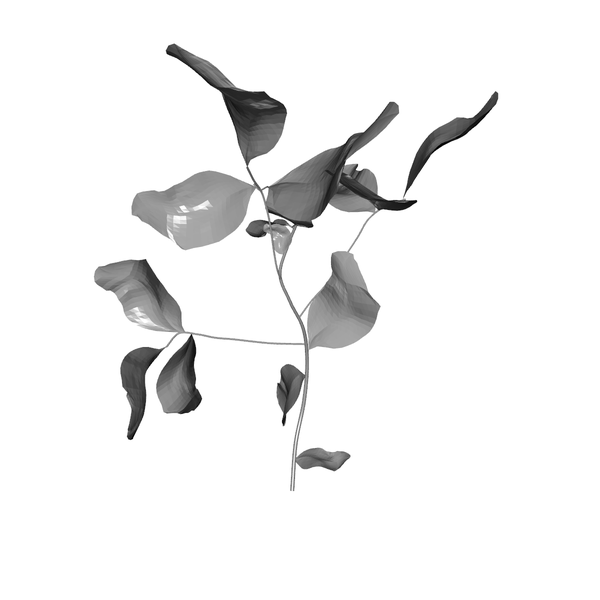}
\vspace{-8pt}

\end{tabular}

}
\caption{\textbf{Parameter Interpolation.} We could explicitly change the overall shape and topology while keeping the identity. We apply the same changes for all nodes for better visualization, but in practice we could manipulate each instance individually.}
\label{fig:interpolate}
\end{figure}

\vspace{-8pt}
\paragraph{Shape}

The shape parameters $\boldsymbol \beta=\{\boldsymbol \beta_i\}_{i=1}^{n_l+n_s}$ define the offsets to the template shape in the canonical space for each organ instance, assuming no 3D deformations. This is conceptually similar to the T-pose for humans, depicting the intrinsic identity-preserving shape factor.
For {\it stems}, the template is parameterized as a unit length cylinder with $m_s$ uniform control points (also as joints here) on the axis. And the shape parameter $\bbeta$ only controls its thickness. 
For {\it leaves}, the template is a $m_{l_1} \times m_{l_2}$ 2D grid enclosed by a left and right contour learned from the mean of the 2D leaf scannings. The grid points of each horizontal grid line serve as the control points of a Catmull-Rom curve~\cite{catmull1974class}, whose control points always lie on the curve, making it easier to interpret and manipulate than other parametric curves, such as Bezier curves~\cite{farin2002handbook}. To further reduce dimensionality, we compress the parameters by learning a linear model $\bPhi_s \in \mathbb{R}^{2m_{l_1}m_{l_2} \times |\bbeta|}$, such that $\bPhi_s^T \bbeta$ represents the offset for two contour curves, where $|\bbeta| \ll 2m_{l_1}m_{l_2}$. We learn the linear model $\bPhi_s$ by running PCA over thousands of 2D scans of leaf (Sec.~\ref{sec:method_learning}). 
This provides a formal definition of shape offset: 
\begin{equation}
\label{eq:shape}
   {\bf{v}}_t +\mathcal{S}(\bbeta) = {\bPhi}_s^{T} \bbeta + {\bf{v}}_t,
\end{equation}
For non-grid surface points, we use the analytical Catmull-Rom spline interpolation based on control points. We refer readers to the supplement for more details.
Fig.~\ref{fig:leaf_pca_viz} illustrates the learned shape model for leaf of several species.

\vspace{-8pt}
\paragraph{Deformation}
Shape and articulation define internal factors of vertex geometry, while external forces like wind, gravity, and contact cause deformations. Modeling these deformations is crucial for biophysical processes (e.g., photosynthesis, growth). Directly controlling vertex locations is inefficient and unrealistic, as real-world deformations typically obey constraints like length and area preservation.

To this end, we turn the joints of leaf into a 2D skeletal structure to drive the 3D deformation. This gives us a fixed topological structure, with one ``main vein" in the middle and multiple horizontal ``sub-veins". These veins comprised a chain structure as shown in Fig.~\ref{fig:details}~(c). 
Moreover, these joints along with the contour points naturally form a quadmesh for each leaf, which we can easily subdivide and triangulate. For stems, we directly connect all joints into a 1D skeleton. 

We could use the joint state along the skeletal structure to encode deformation. For the $j$-th control point on the 2D skeleton, a joint angle $\boldsymbol{\psi}_{j}$ indicates its rotation relative to its parent node. We use forward kinematics to convert the angles into deformed 3D locations:
\begin{equation}
\label{eq:leaf_deformation}
\mathcal{D}(\mathbf{v}_j; \bgamma_i
)  = 
\prod_{j^\prime \in \textrm{ans}(j)} \rm{T}(\boldsymbol\tau_{j^\prime}, d_j, 1) 
\cdot \mathbf{v}_j, 
\end{equation}
where $\mathbf{v}_j$ is the input 3D vertex location before deformation in node $i$, $\mathrm{ans}(j)$ denotes the ancestors of control point $j$, $\mathbf{T}$ is the rigid transformation corresponding to each joint angle, and $d_j$ is the shape-defined skeletal distance between control point $j$ and its parent $d_j$. 
Such deformation enjoys many benefits: 1) strong local rigidity preservation, making leaf area and stem length stay roughly unchanged; 2) ease of calculation, 3) ease of compression due to the spatial smoothness of the joint angles. Fig.~\ref{fig:details}~(c) depicts the deformation process.

Each leaf  has 1 main vein, and each main control points has 2 sub-veins. The deformation parameter count is $2m_{l_1}m_{l_2}$ since each control point has 2 degrees of freedom. Inspired by the success of FLAME~\cite{FLAME:SiggraphAsia2017} in compressing expression-related deformations, we again adopt a linear dimensionality reduction model to represent deformation as $\boldsymbol{\Phi}_d^T \bgamma $, where $\bPhi_d \in \mathbb{R}^{2m_{l_1}m_{l_2} \times |\bgamma|}$ such that $\boldsymbol{\Phi}_d^T \bgamma$ approximates the entire deformation field with $|\bgamma| \ll 2m_{l_1}m_{l_2}$. 
We learn $\bPhi_s$ by running PCA over real-world data, later described in Sec.~\ref{sec:method_learning}. 

\begin{figure*}[t]
\centering
\def\arraystretch{0.3}
\setlength{\tabcolsep}{0.5pt}
{
\begin{tabular}{ccccccc}
&{\small Soybean 1} 
&{\small Soybean 2}
&{\small Soybean 3}
&{\small Maize 1} 
&{\small Maize 2}
&{\small Maize 3}

\\
\raisebox{5mm}{\rotatebox{90}{{\small Input pcd}}}
&\includegraphics[width=0.14\linewidth]{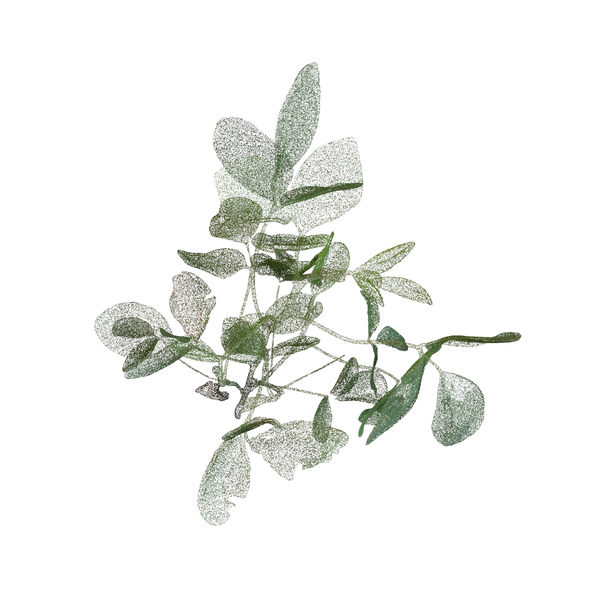}
&\includegraphics[width=0.14\linewidth]{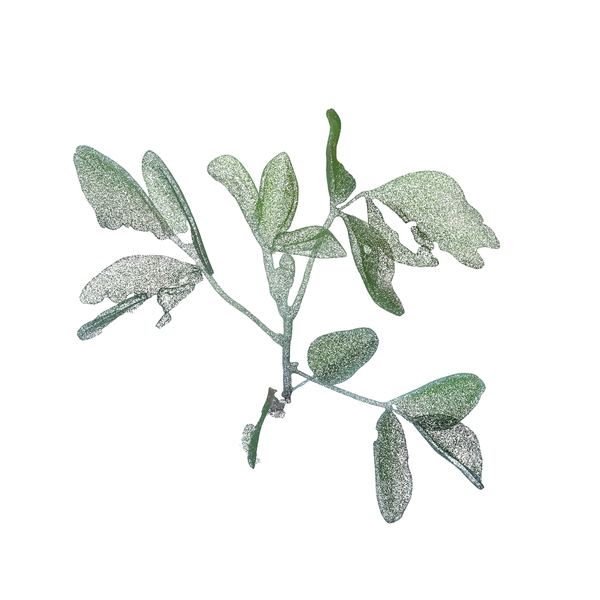}
&\includegraphics[width=0.14\linewidth]{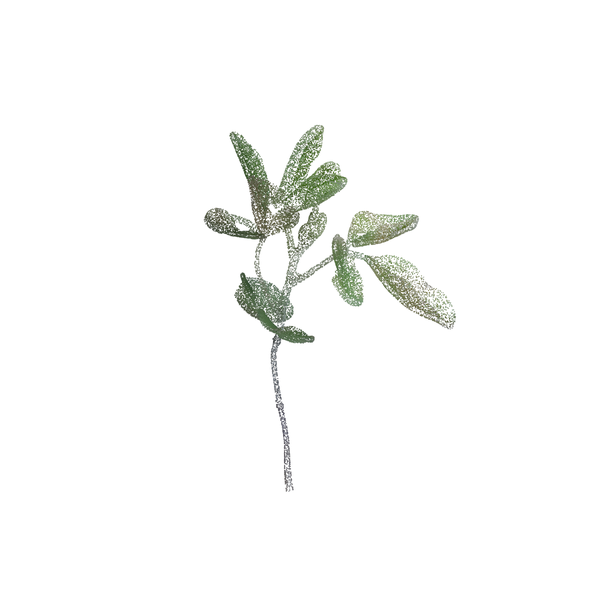}
&\includegraphics[width=0.14\linewidth]{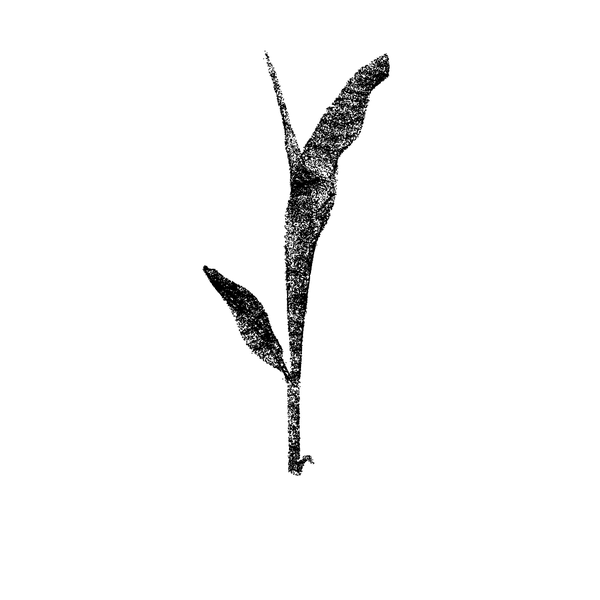}
&\includegraphics[width=0.14\linewidth]{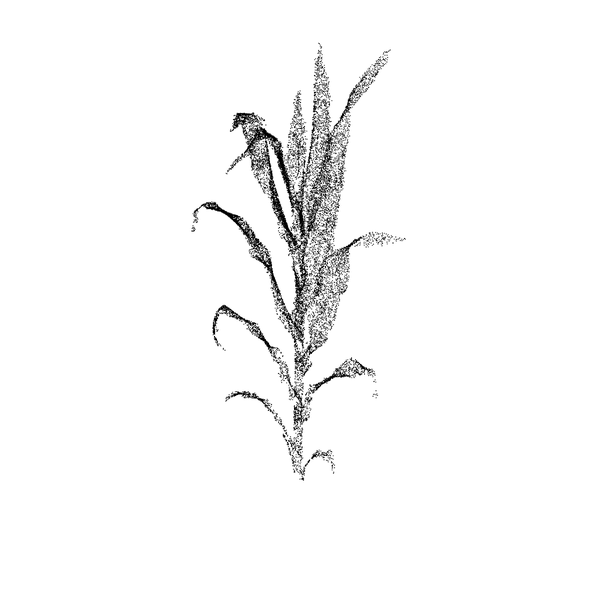}
&\includegraphics[width=0.14\linewidth]{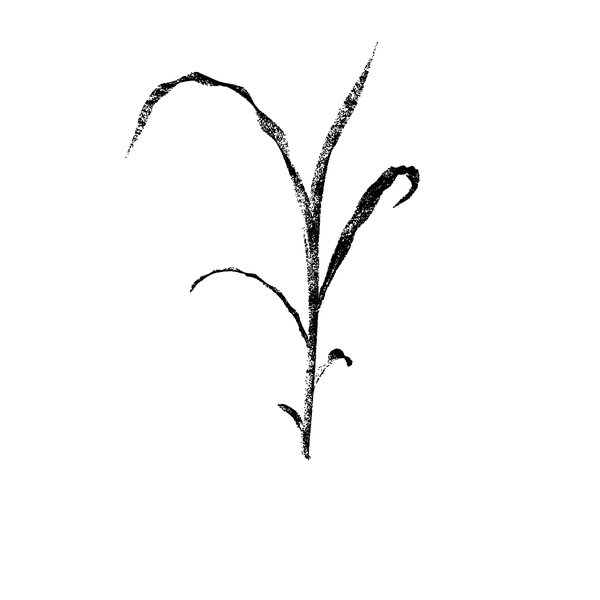}

\\
\raisebox{8mm}{\rotatebox{90}{{\small NKSR}}}
&\includegraphics[width=0.14\linewidth]{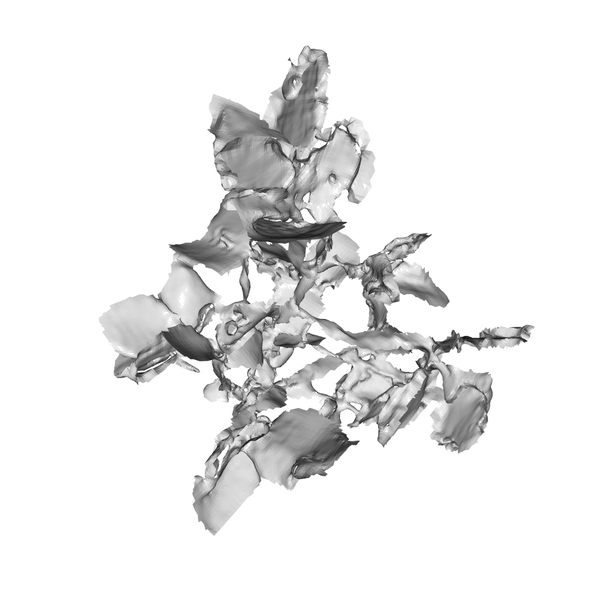}
&\includegraphics[width=0.14\linewidth]{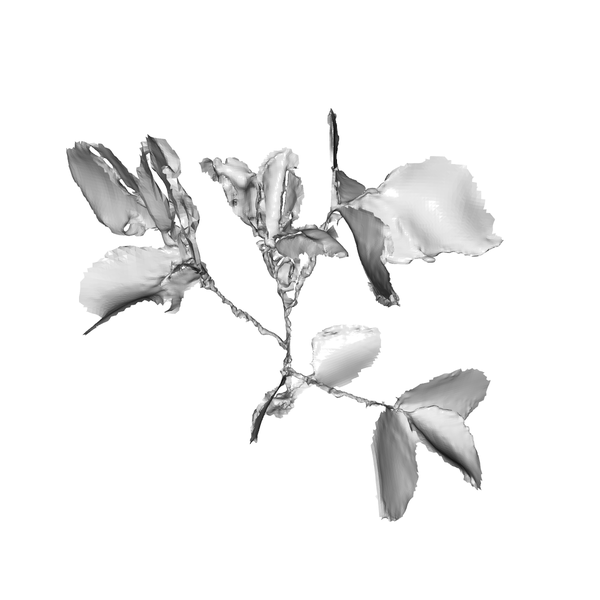}
&\includegraphics[width=0.14\linewidth]{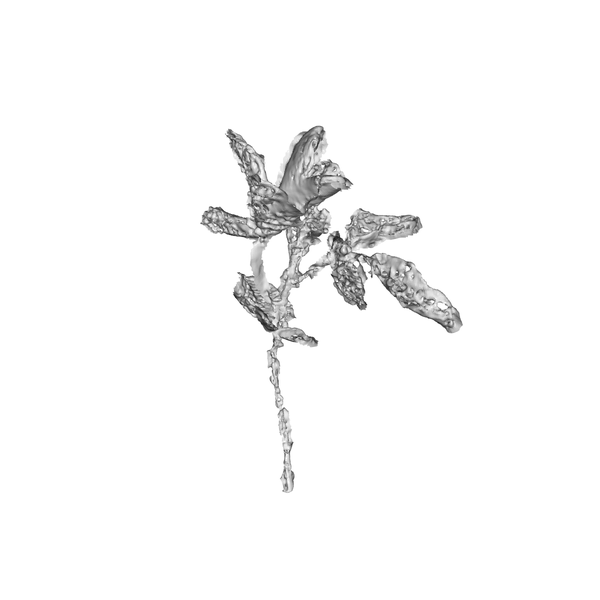}
&\includegraphics[width=0.14\linewidth]{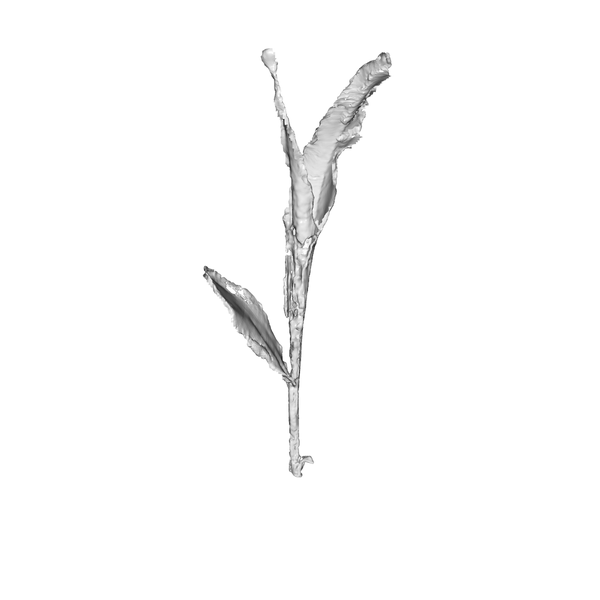}
&\includegraphics[width=0.14\linewidth]{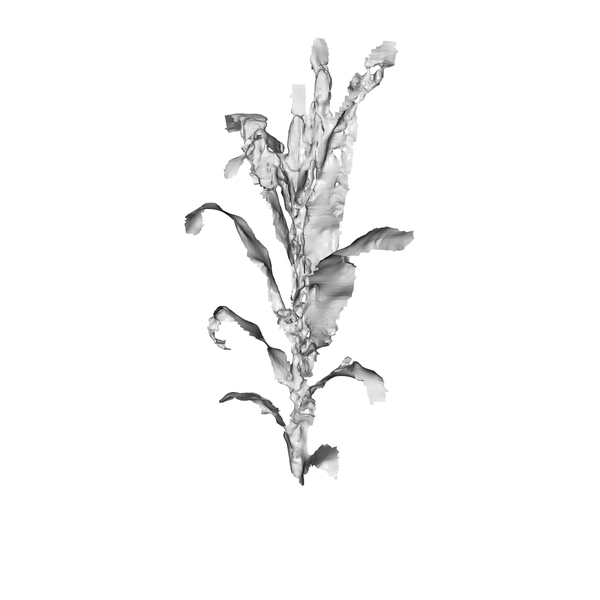}
&\includegraphics[width=0.14\linewidth]{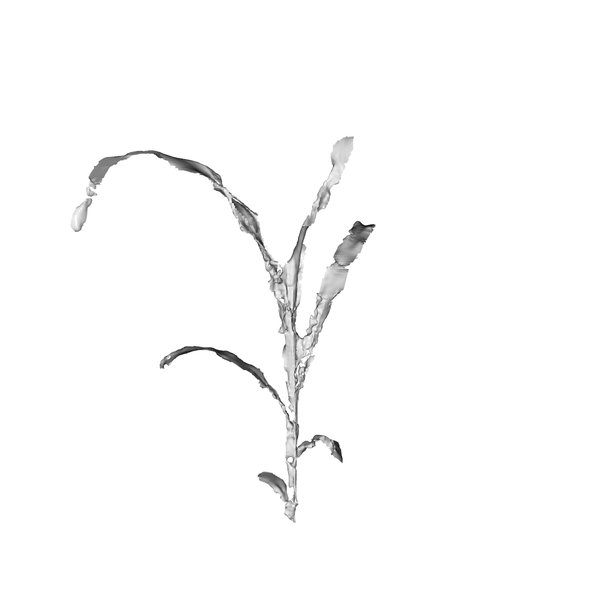}

\\
\raisebox{5mm}{\rotatebox{90}{{\small SimpleProc}}}
&\includegraphics[width=0.14\linewidth]{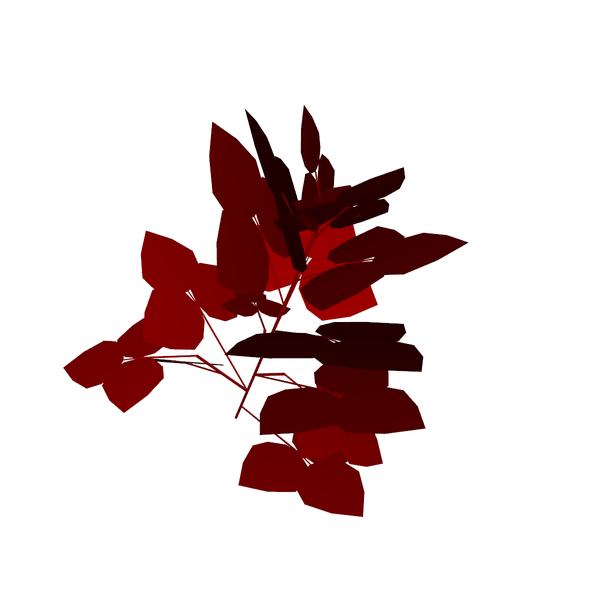}
&\includegraphics[width=0.14\linewidth]{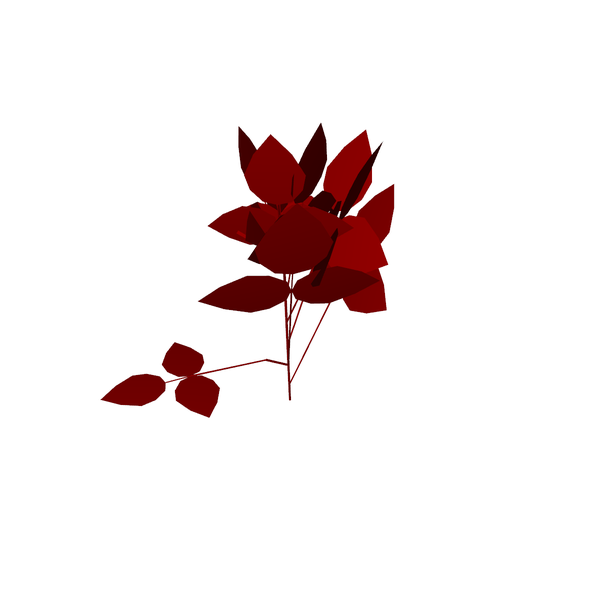}
&\includegraphics[width=0.14\linewidth]{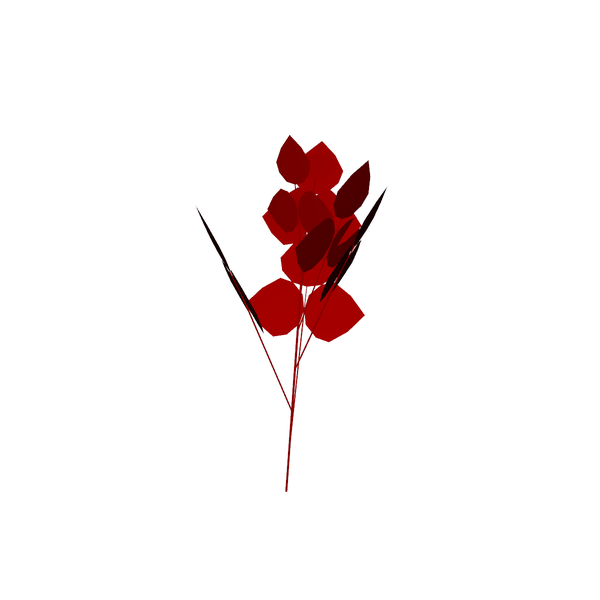}
&\includegraphics[width=0.14\linewidth]{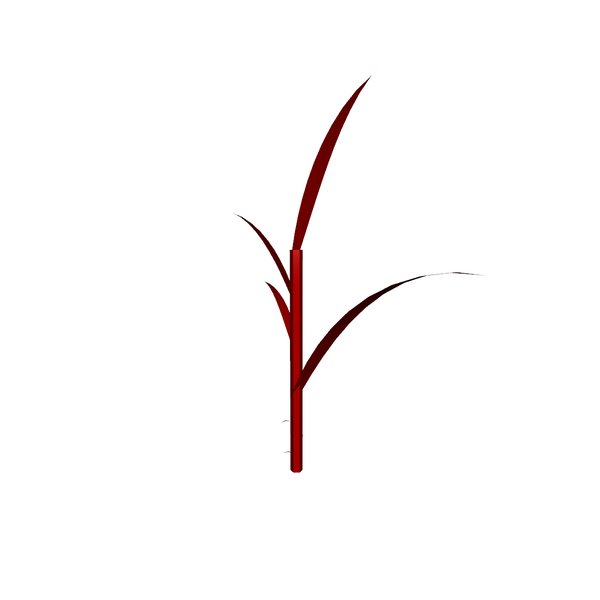}
&\includegraphics[width=0.14\linewidth]{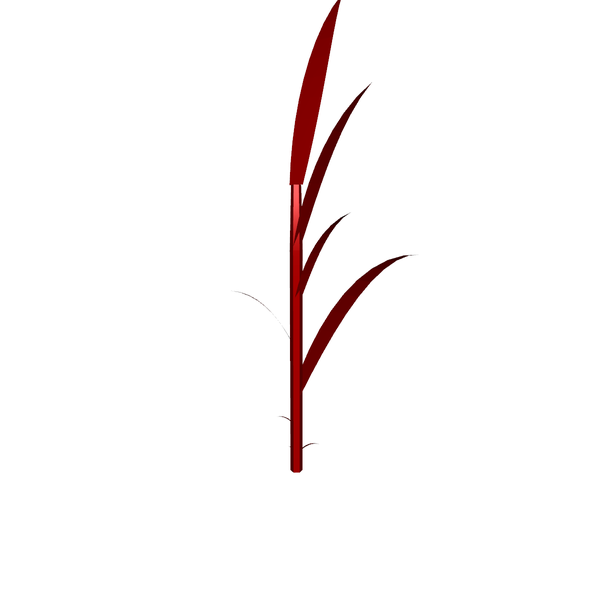}
&\includegraphics[width=0.14\linewidth]{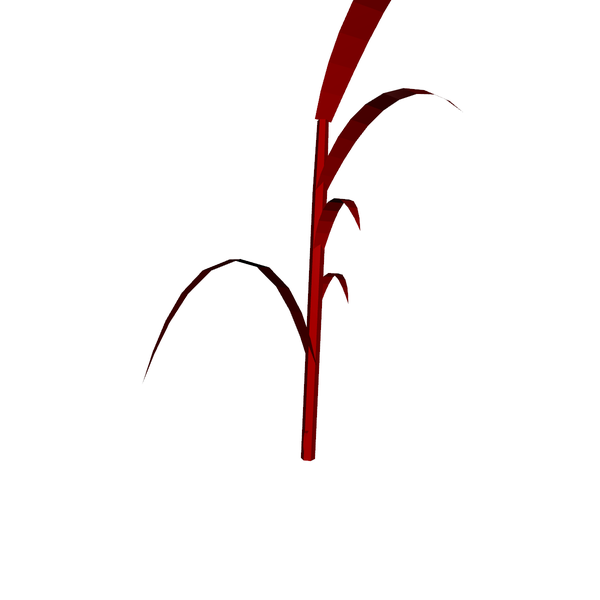}

\\
\raisebox{8mm}{\rotatebox{90}{{\small Ours}}}
&\includegraphics[width=0.14\linewidth]{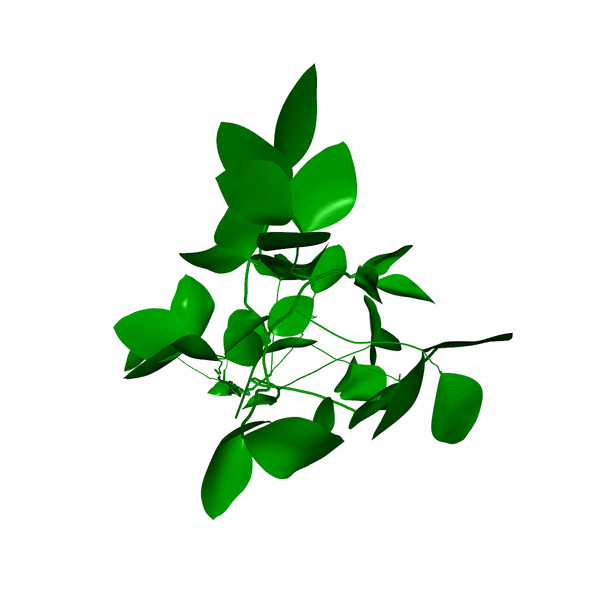}
&\includegraphics[width=0.14\linewidth]{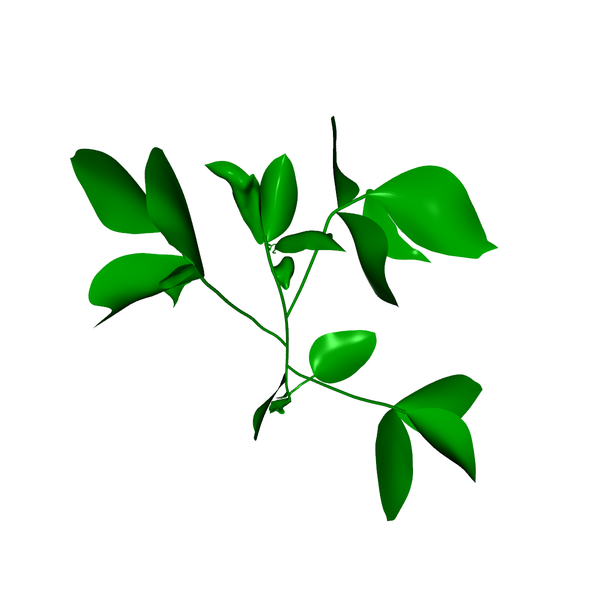}
&\includegraphics[width=0.14\linewidth]{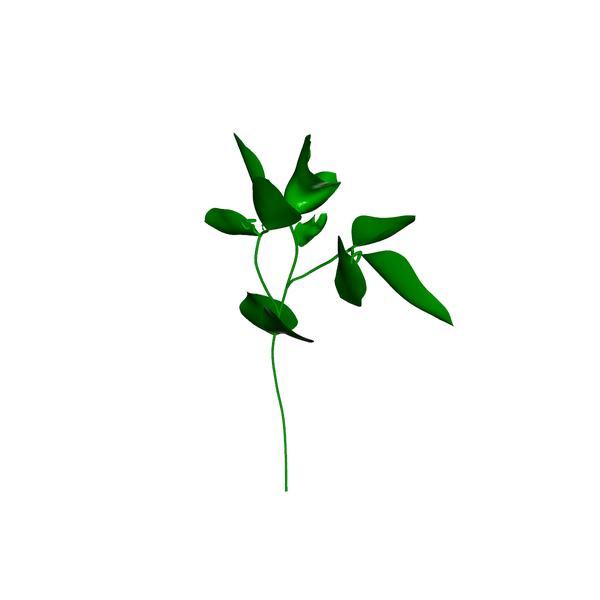}
&\includegraphics[width=0.14\linewidth]{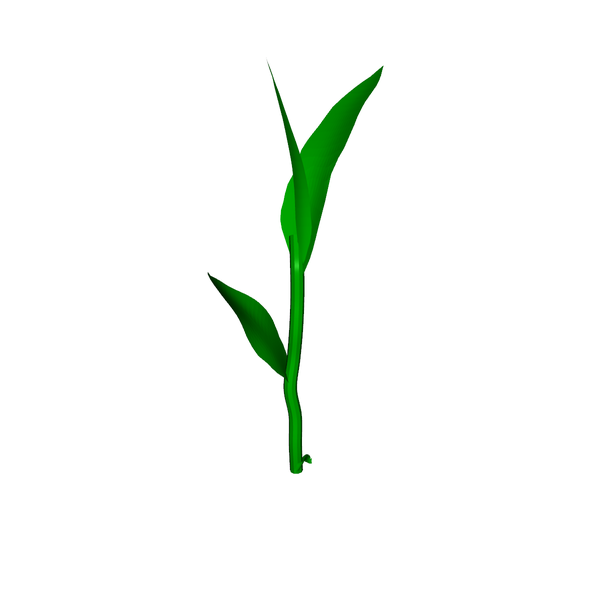}
&\includegraphics[width=0.14\linewidth]{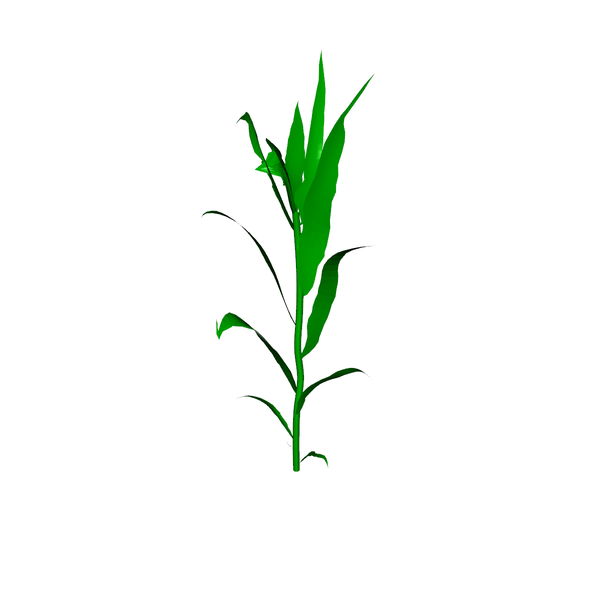}
&\includegraphics[width=0.14\linewidth]{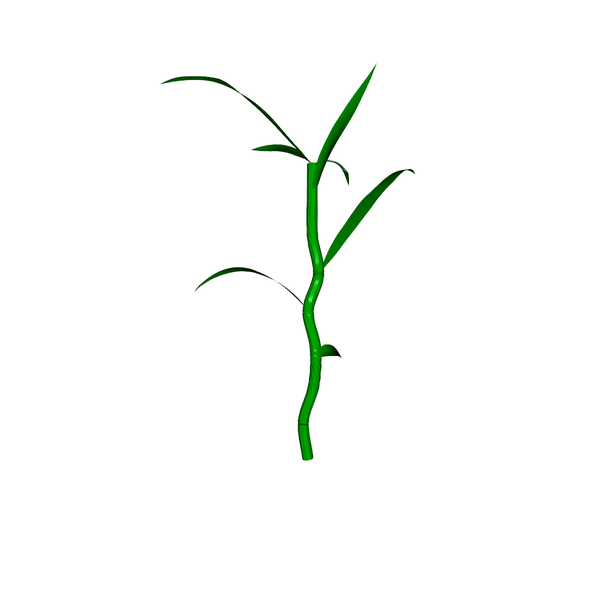}

\end{tabular}
}

\caption{\textbf{Reconstruction from 3D point cloud.} The results show that our model fits the input more accurately than the baselines.}
\label{fig:3d_recon_all}
\end{figure*}

\subsection{Learning Demeter from Real-World Data} \label{sec:method_learning}

\begin{figure*}
    \vspace{-8mm}
    \centering
    \includegraphics[width=.8\textwidth]{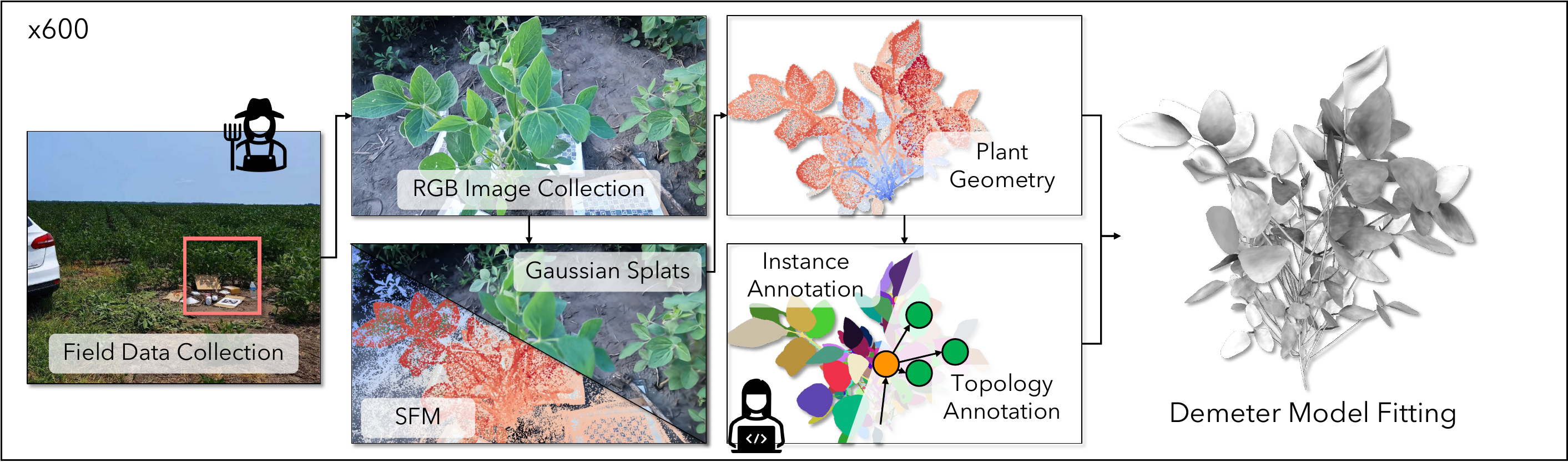}
    \captionof{figure}{{\bf Demeter-soybean data collection process}. We collect around 600 monocular RGB video of soybean in the field, and run a Gaussian Splatting based multiview reconstruction pipeline to get the mesh. After that, we manually annotate the instance segmentation and the topology, and fit this ground-truth data with our Demeter parametric model. A similar process can be adapted to other species.}
    \label{fig:data_processing}
\end{figure*}

Demeter model parameters $\bPhi$ can be learned using real-world plant data captured in the crop field, grounding the learned Demeter model with realistic in-the-wild variations. To demonstrate this, we collect multi-view field data to recover complete geometry, scan 2D leaf images, employ annotators to label each plant component, and finally acquire a clean topology $ \boldsymbol{\Gamma} $ (see Fig.~\ref{fig:data_processing}). The shape parameter basis $ \boldsymbol{\Phi}_s$ and joints $\boldsymbol{\Phi}_j$ is then learned by fitting contours from 2D leaf scans using PCA. The deformation parameter basis $ \boldsymbol{\Phi}_d $ is derived by fitting real-world 3D deformation data from annotated scans, followed by structured joint optimization. 
We refer to our model as Demeter-Soybean. Note that this entire process has little species-specific design and can therefore generalize to other plant species.

 

\vspace{-10pt}
\paragraph{Data Collection and Processing} \label{subsec:data_collection}

Most 3D plant scans for phenotyping are conducted in controlled lab environments, which may lack the realism and variability of real-world field conditions. 
To address this issue, we created a custom annotated 3D soybean dataset with 600 soybean samples of various growth stages from a smart soybean farm. Using these samples, we captured over 600 multi-view RGB videos with a GoPro in both indoor and outdoor settings.

For 3D geometry extraction, we first applied structure-from-motion with Hloc~\cite{sarlin2019coarse, sarlin2020superglue} to determine camera poses. We then used 4DGS~\cite{wu20244d} to reconstruct dynamic plants affected by wind. Although 4DGS captures high-quality renderings and motion, it lacks precision in surface geometry. To address this, we set the $t=0$ and freeze movement to render a set of static multi-view images. We then apply 2DGS~\cite{huang20242d} on these images to extract a high-quality mesh.

The reconstructed mesh sometimes had missing parts, particularly stems, which we manually corrected. We also annotated instance segmentation and topology to provide ground-truth data, as shown in Fig.~\ref{fig:data_processing}. These annotations are essential for training the deformation model and will serve as benchmarks for 3D reconstruction.

\vspace{-10pt}
\paragraph{Learning Shape Basis from 2D Data} \label{sec:learn_from_2d}

We derive the shape parameter basis, $\bPhi_s$ from real-world 2D leaf scans in the FGLIR~\cite{ando2021robust} dataset and Folio dataset~\cite{folio_338}. All images are calibrated to have accurate metric scale.

To extract the contours, we first annotate two key points on each leaf—the \textit{base} (where the leaf connects to the stalk) and the \textit{tip}, and optionally some keypoints in the middle. These annotations enable us to align each leaf along the x-axis. Second, we extract the leaf mask using SAM2~\cite{ravi2024sam} and extract the entire contour from the mask. Third, we divide the contour into two parts according to the keypoints.

Next, we construct the inner points of a leaf by dividing it into a grid with $m_{l_1} \times m_{l_2}$ control points given the contours. For a \textit{simply connected} leaf with general shapes, we first numerically map the control points from canonical UV space to each 2D leaf scanning, and then learn the linear model from these data. Inspired by ~\cite{schneider2015smooth}, we model this problem as a unique mapping $\varphi: \Omega \rightarrow \Omega'$ between UV domain $\Omega'$ and leaf domain $\Omega$. Specifically, we define the boundary mapping $b:\partial \Omega \rightarrow \partial \Omega'$ by mapping the left/right contour of the leaf to the left/right contour of a circle, and solve Laplace equation
\begin{equation}
    \Delta \varphi =0, \text{ s.t. } \varphi|_{\partial \Omega}=b,
\end{equation}
to get the mapping $\varphi$ and its inverse $\varphi^{-1}$ numerically, forming a smooth bijective mapping (Fig.~\ref{fig:details}(e)). Particularly, for some oval leaves, such as soybean, whose contours have at most two intersections with the horizontal lines, we could simply build this mapping by linearly interpolating between each pair of intersections, and the positions of these control points are also linearly dependent on the shape parameter $\bbeta$. Fig.~\ref{fig:details}~(c, e) illustrates how a leaf shape is controlled by such contour curves.

Finally, we perform PCA on these control points, resulting in a linear model $\boldsymbol{\Phi}_s \in \mathbb{R}^{2m_{l_1}m_{l_2} \times |\bbeta|}$ capturing the variation in shape. The PCA components of soybean are visualized in Fig.~\ref{fig:leaf_pca_viz_soybean}.

\vspace{-10pt}
\paragraph{Learning Deformation Basis and Articulation from 3D Data}






Unlike leaf and stem shapes, which can be observed directly from 2D scans, extracting deformation is challenging due to entanglement with articulation and shape factors. Thus, we design a scheme to disentangle these factors through joint optimization, enabling us to learn the articulation and deformation basis $\bPhi_d$. Formally, given the input 3D point cloud collection, $\mathcal{P}=\{\mathcal{P}_{1...n}\}$, where $n$ is the number of nodes, ground-truth annotated topology of each node, $\bGamma_{1 \dots n}$, semantic and instance labels, $\mathbf{tp}_{1 \dots n}$, and pretrained shape basis, $\bPhi_s$, our goal is to estimate the articulation parameters, $\btheta_{1...n}$, and shape parameters, $\bbeta_{1 \dots n}$ for each node. Additionally, we aim to estimate articulation- and shape-independent deformation for each plant, so we can train the deformation PCA model $\bPhi_d$ and estimate $\bgamma_{1 \dots n}$ for each node.

However, joint training is challenging due to nonlinearity and complex forward kinematics. Inspired by prior work~\cite{liu2023building, yang2022banmo}, we adopt a multi-stage approach. In the first stage, we fit shape parameter $\hat{\bbeta}_i$ using shape PCA basis and the raw 3D deformation by minimizing the chamfer distance between the node template and the point cloud $\mathcal{P}_i$ for each instance segmentation. This happens in world coordinates without articulation constraints (e.g., leaves detached from stems). Since these deformed templates are aligned, we learn the PCA basis $\bPhi_d$ from the real-world deformations to compress them into a compact space  $\bgamma_i$ without loss of expressiveness, shown in Fig.~\ref{fig:leaf_pca_viz}. Then we joint optimize $\bgamma_i$ and $\bbeta_i$ again by minimizing the chamfer distance, as shown in Eq.~\ref{eq:learn_deform}.

\begin{equation}
\label{eq:learn_deform}
    \min_{\bbeta, \btheta} 
    \mathrm{CD}(\mathcal{P}_i, \{ {\cal D}({\bf{v}}_t +{\cal S}(\boldsymbol\beta );\bgamma \}_{\forall \mathbf{v}_t} )
\end{equation}

In the second stage, given the initial articulation $\hat{\btheta}$ from instance segmentation, the estimated $\bbeta_i$ and $\bgamma_i$, and annotated topology $\bGamma$, we perform forward kinematic to fit the whole input point cloud $\mathcal{P}$ by Eq.~\ref{eq:learn_articulation} to optimize the articulation and the deformation parameters for stem only, resulting in our final estimation as shown in the right of Fig.~\ref{fig:data_processing}. We optimize the deformation for stem because it needs to compensation for the disconnections when we attach node together.
\begin{equation}
\label{eq:learn_articulation}
    \min_{\bgamma_{\text{leaf}}, \btheta} 
    \mathrm{CD}(\mathcal{P}, \{ {\rm{T}}(\boldsymbol\theta ;{\rm{ }}\boldsymbol\Gamma) \cdot {\cal D}({\bf{v}}_t +{\cal S}(\boldsymbol\beta );\bgamma \}_{\forall \mathbf{v}_t} )
\end{equation}
where $\mathrm{CD}$ is Chamfer distance, $\mathbf{v}_t$ represents canonical template vertex, $\bGamma$ is input topology, $\mathcal{S}(\bbeta)$ is the shape model in Eq.~\ref{eq:shape} and $\mathcal{D}(\mathbf{v}; \bbeta)$ is the deformation function in Eq.~\ref{eq:leaf_deformation} and $T(\btheta, \bGamma)$ is the kinematic articulation in Eq.~\ref{eq:kinematics}.

\subsection{3D Plant Reconstruction with Demeter} \label{sec:method_inference}

Demeter encodes strong morphological priors about the plant. Inspired by the success of SMPL for human reconstruction~\cite{goel2023humans, kanazawa2018end}, 
we investigate the use of Demeter parameters for multi-view and single-view reconstruction. 


\vspace{-8pt}
\paragraph{Multiview Reconstruction} \label{sec:method_inference_3d}





Using multiview images, we first recover the point cloud with 2DGS, like in Demeter training. However, unlike before, there is no ground-truth instance mask or topology annotation available during real-world inference. Hence, we adopt a three-step strategy. First, we perform instance segmentation on the recovered 3D point cloud using a PointTransformer-based network. Next, we infer the topology based on a minimum spanning tree. Finally, we fit parameters as in Eq.~\ref{eq:learn_deform}.

We trained a PointTransformer-V3~\cite{wu2024point} based instance segmentation network. 
Inspired by prior work in 2D~\cite{bai2017deep}, we predict semantic logits for each point and its truncated inverse distance to the instance boundary. Using the inferred (thresholded) instance boundaries, we remove boundary points to isolate connected components. We then run DBSCAN with the same threshold to cluster points into individual leaves and stems. We find this approach effective and visualizations can be found in the supplementary. 

Given the instance predictions, the topology $\boldsymbol\Gamma$ is automatically inferred by finding a minimal spanning tree rooted from the main stem. Finally, we recover the parametric model using the described method for Eq.~\ref{eq:learn_deform}.

\vspace{-8pt}
\paragraph{Single Image Reconstruction} \label{sec:method_inference_2d}

We present a baseline method for single-view plant reconstruction to showcase the usefulness of Demeter for constraining 3D geometry. Given a single RGB plant image, we first apply the pretrained SAM~\cite{ravi2024sam} to get the mask, and use Mask-RCNN~\cite{he2017mask} on the masked image to predict instance segmentation with 3 classes: leaf, stem, and main stem. 

We apply an off-the-shelf depth estimator\cite{yang2024depth} to lift all the instances to 3D, and perform simple filtering (details in supplementary). Then we infer the topology from the partial point cloud via minimal spanning tree, filtering stems without any children. Afterwards, we fit the Demeter model parameters in the same way as before, but constraining the shape $\boldsymbol \beta$ and adopt L1 Chamfer distance for robustness.

\section{Experiments}

We 
evaluate 
the Demeter model across several aspects: fitting (Sec.~\ref{sec:fidelity}), reconstruction (Sec.~\ref{sec:reconstruction}) and interpretability (Sec.~\ref{sec:latent}). Additional training details and quantitative results are in the supp. material.




\subsection{Fitting}
\label{sec:fidelity}


\begin{figure*}
\centering
\def\arraystretch{0.3}
\setlength{\tabcolsep}{0.5pt}
\resizebox{.95\linewidth}{!}{
\begin{tabular}{ccccccc} 
&{\Large Pepper}  
&{\Large Rose 1}
&{\Large Rose 2}
&{\Large Tobacco}
&{\Large Maize}
&{\Large Papaya*}

\\
\raisebox{25mm}{\rotatebox{90}{{\Large Input}}}
&\includegraphics[width=0.24\linewidth]{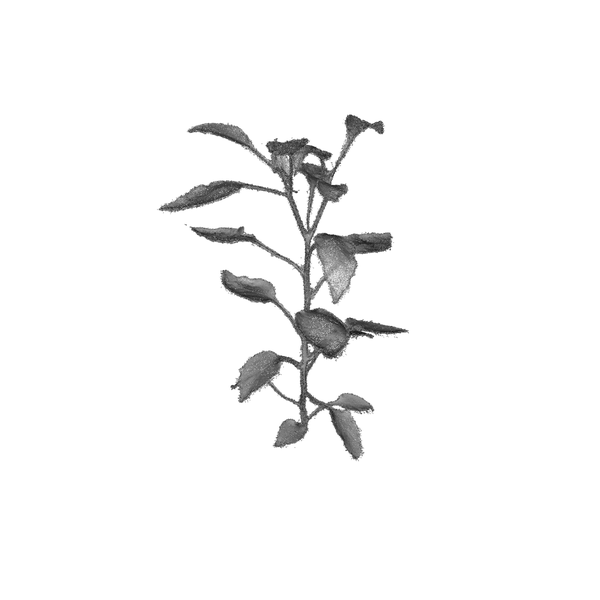}
&\includegraphics[width=0.24\linewidth]{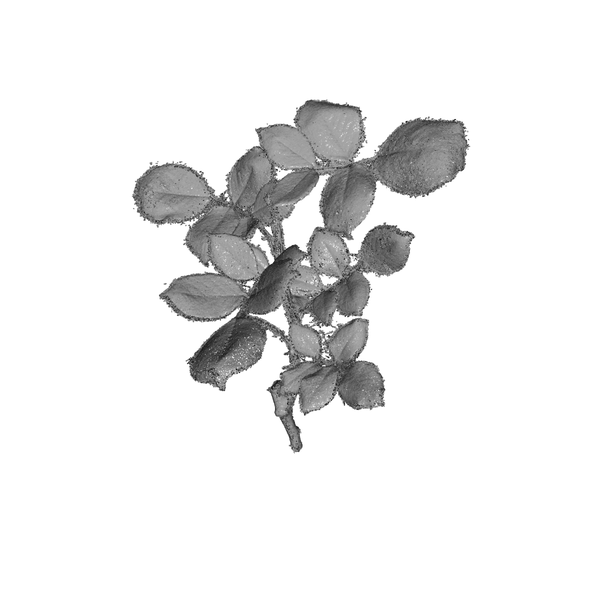}
&\includegraphics[width=0.24\linewidth]{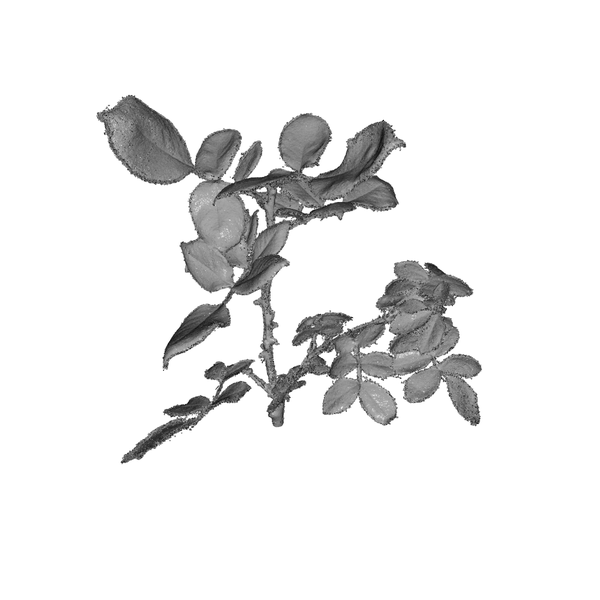}
&\includegraphics[width=0.24\linewidth]{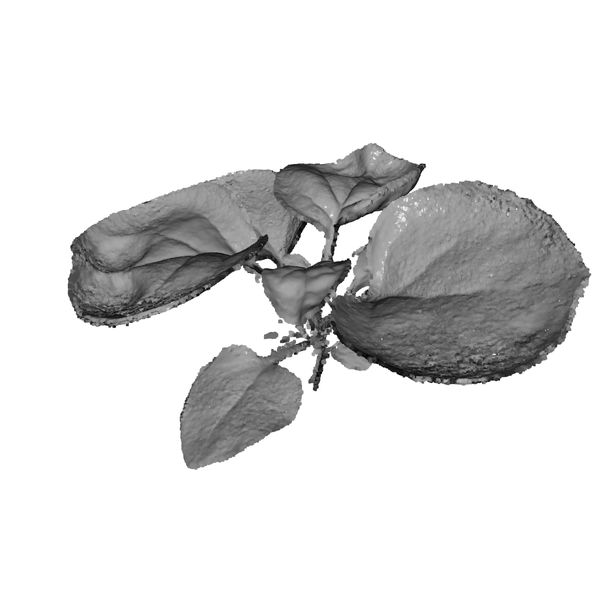}
&\includegraphics[width=0.24\linewidth]{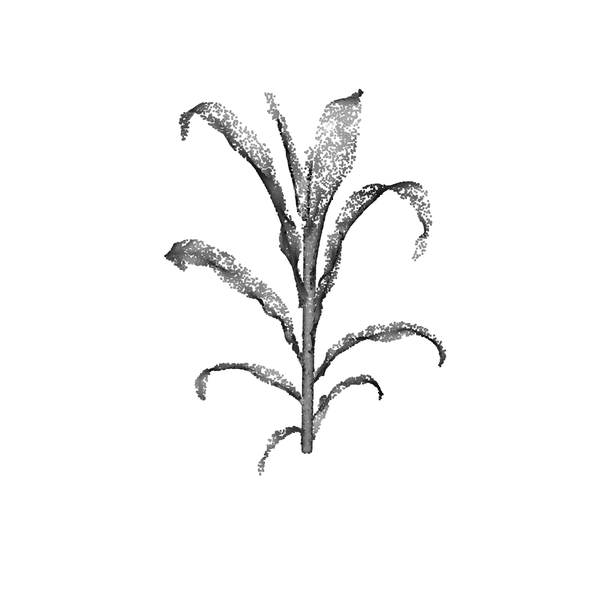}
&\includegraphics[width=0.24\linewidth]{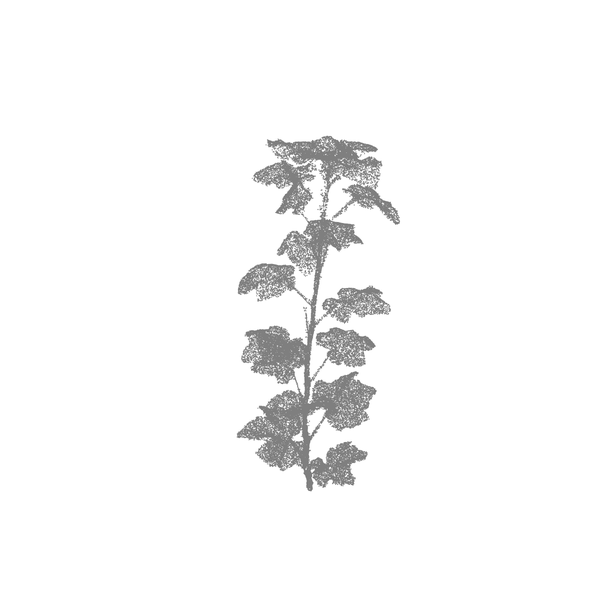}




\\
\raisebox{20mm}{\rotatebox{90}{{\Large Demeter}}}
&\includegraphics[width=0.24\linewidth]{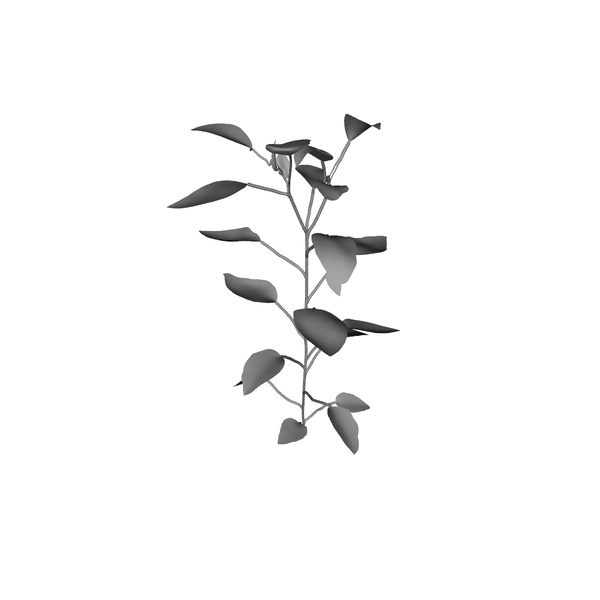}
&\includegraphics[width=0.24\linewidth]{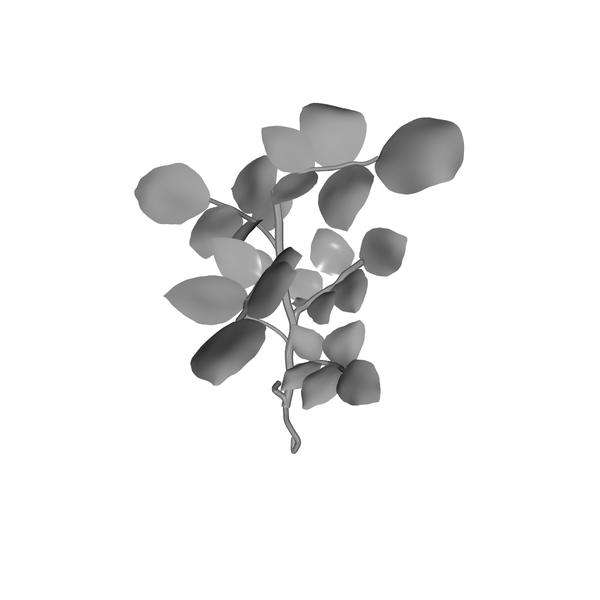}
&\includegraphics[width=0.24\linewidth]{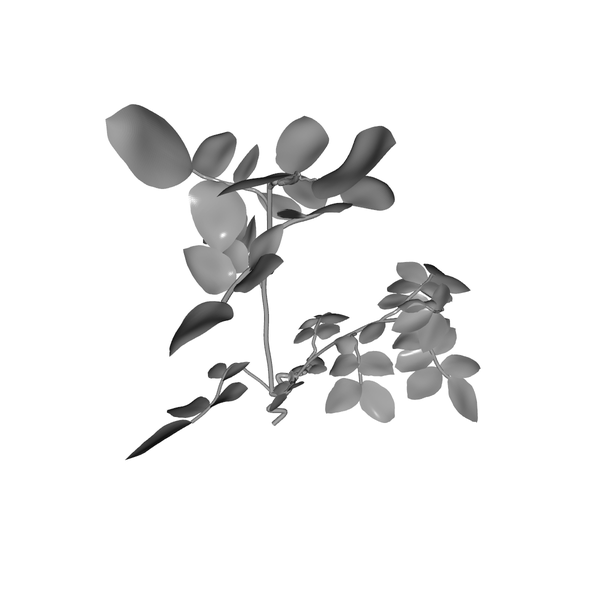}
&\includegraphics[width=0.24\linewidth]{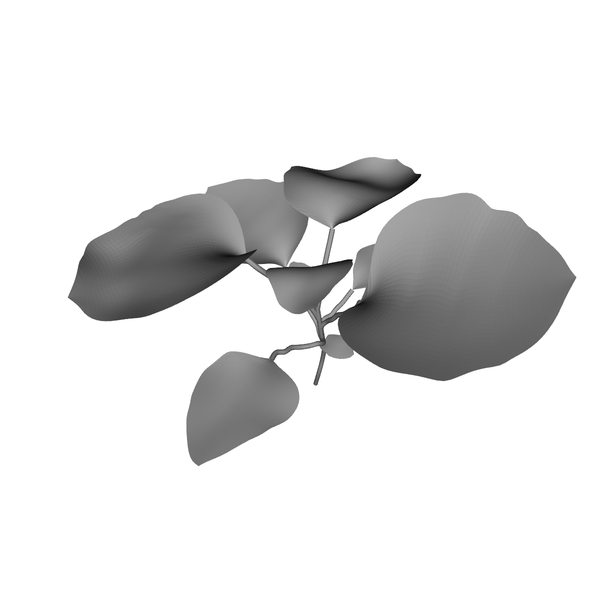}
&\includegraphics[width=0.24\linewidth]{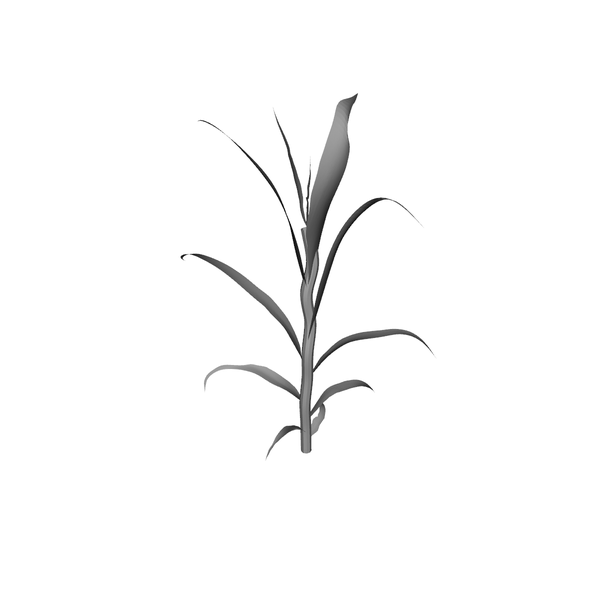}
&\includegraphics[width=0.24\linewidth]{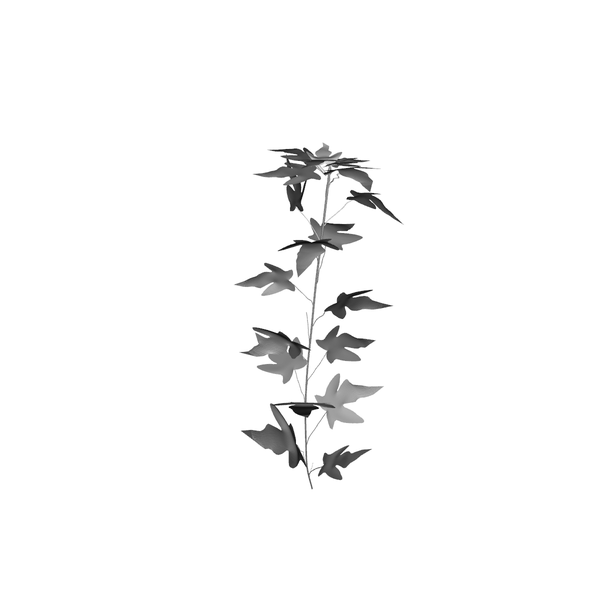}

\end{tabular}
}
\caption{\textbf{Generalization}. 
We learn Demeter on each species (pepper, rose, tobacco, maize) and validate on novel test shapes. Here we use the Papaya model to fit the point cloud of Ribes since we lack 2D leaf scans for Ribes and these two species have similar leaves.}

\label{fig:fitting_general}
\end{figure*}

Demeter can represent plants of different species, capturing individual leaves and stems with high fidelity (Fig.~\ref{fig:fitting_general}). We select Pepper and Rose from the PLANesT3D~\cite{mertouglu2024planest} dataset, and Tobacco from the PLANT3D~\cite{conn2017high} dataset. In Fig.~\ref{fig:3d_recon_all}, we compare Demeter with SimpleProc, a baseline procedural model adapted from CropCraft~\cite{zhai2024cropcraft}. SimpleProc has a compact set of 8 global shape parameters such as average leaf length and node count, and is fitted by Bayesian optimization of the Chamfer distance with the ground truth point cloud. The results show that the Demeter-Soybean is much more expressive and can capture the individual leaf and stem shapes that SimpleProc cannot.

\subsection{Reconstruction}
\label{sec:reconstruction}

\paragraph{Multi-view Reconstruction}



\begin{table}[t]
\centering
\scalebox{0.8}{


\begin{tabular}{cccc}
\hline
\multicolumn{1}{l}{}       & NKSR~\cite{huang2023nksr}                               & SimpProc~\cite{zhai2024cropcraft}         & Ours            \\ \hline
Smooth                 &                  & $\sqrt{}$            & $\sqrt{}$       \\
Disentangle                &                                     & $\sqrt{}$            & $\sqrt{}$       \\
Learnable               & $\sqrt{}$                           & \multicolumn{1}{l}{} & $\sqrt{}$       \\ \hline
Soybean, CD ↓              & \cellcolor[HTML]{FECD9E}{0.0030}  & 0.0376               & \cellcolor[HTML]{FD9B9A}{0.0016} \\
Maize, CD ↓                & \cellcolor[HTML]{FD9B9A}{0.0023}                     & 0.0557         & \cellcolor[HTML]{FECD9E}0.0071           \\
Soybean, size (KB) ↓ & 5785.6                              & \cellcolor[HTML]{FD9B9A}{0.2754}      & \cellcolor[HTML]{FECD9E}{3.3750}         \\
Maize, size (KB) ↓   & 3686.4                              & \cellcolor[HTML]{FD9B9A}{0.2236}      & \cellcolor[HTML]{FECD9E}{1.7656}          \\ \hline
\end{tabular}
}
\vspace{-.5em}
\caption{\textbf{Quantitative 3D reconstruction results.} We report the storage size in kilobytes (KB) and Chamfer distance (CD) in normalized scale. Results show that we fit the input point cloud better than the procedural method~\cite{zhai2024cropcraft}. NKSR outputs large storage size because it stores dense mesh. We also achieve comparable CD while using much less parameters. We highlight the \colorbox{best_color}{best} and \colorbox{second_color}{second best} values.
}
\label{Tab:3d_recon_all}
\vspace{-.8em}
\end{table}
We train PointTransformer-V3 
on 80 cleaned soybean point clouds and 5 maize point clouds for the segmentation module and evaluate the reconstruction pipeline on unseen testing samples. We compare our approach against SimpleProc~\cite{zhai2024cropcraft}, an inverse procedural generation framework based on L-system, and NKSR~\cite{huang2023nksr}, the state-of-the-art neural SDF-based 3D reconstruction pipeline. As shown in Fig.~\ref{fig:3d_recon_all} and Tab.~\ref{Tab:3d_recon_all}, our model performs well on unseen samples, recovering realistic geometry with high fidelity to the input. SimpleProc is compact but misaligned with the input, while NKSR~\cite{huang2023nksr} achieves low error yet produces a triangular mesh with thick, unsmooth leaves and stems and poor connectivity, preventing a biophysically functional plant model.


\paragraph{Single-view Reconstruction}
We trained Mask-RCNN 
on 32K images across 60 plants. The qualitative results of the single view reconstruction are shown in Fig.~\ref{fig:2d_recon_all}. We further evaluate the IOU between the rendered mask and the input as a quantaitive alignment score. From the figure, we see that Demeter achieves better alignment score (IoU: $0.328$) with the input than Meshy (IoU: $0.206$) and Zero12345++~\cite{liu2023one2345++} (IoU: $0.296$). 
In contrast, our baselines produce less accurate shapes, struggling with the complexity of plant structures.

\subsection{Discussions}
\label{sec:latent}
\paragraph{Latent Space Visualization}
The Demeter model allows explicit control of deformation, shape, and topology while preserving plausibility (Fig.~\ref{fig:interpolate}). For topology, we cut branches to create smaller trees or duplicate subtrees to enlarge them. Leaf and shape deformations are adjusted using PCA coefficients, and stem deformation is linearly manipulated from straight to curved.

\begin{table}[]
    \centering
    \resizebox{\linewidth}{!}{
\setlength{\tabcolsep}{0.1em} 
\renewcommand{\arraystretch}{1.}
    \begin{tabular}{cccc}
    \footnotesize Net Photosynthesis & \footnotesize Early Morning (6 AM) & \footnotesize Noon (12 PM)  \\
          
        \includegraphics[width=.3\linewidth, trim={0 0 0 0.5cm}, clip]{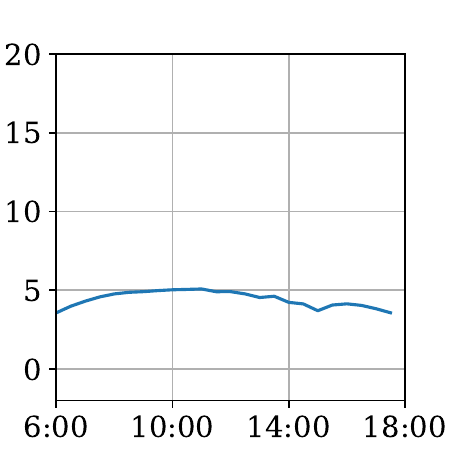}
        &
        \includegraphics[width=.35\linewidth]{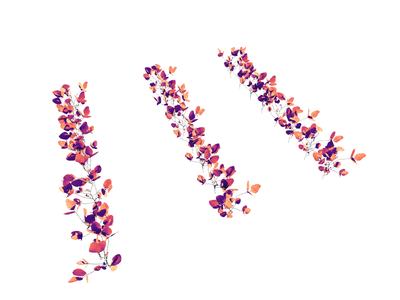}
        & 
        \includegraphics[width=.35\linewidth]{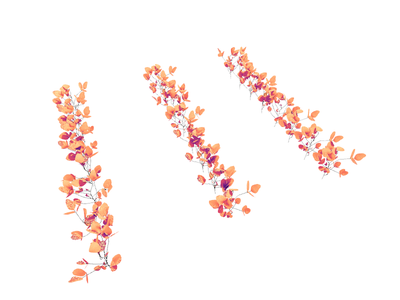}

        \\

       \includegraphics[width=.3\linewidth, trim={0 0 0 0.2cm}, clip]{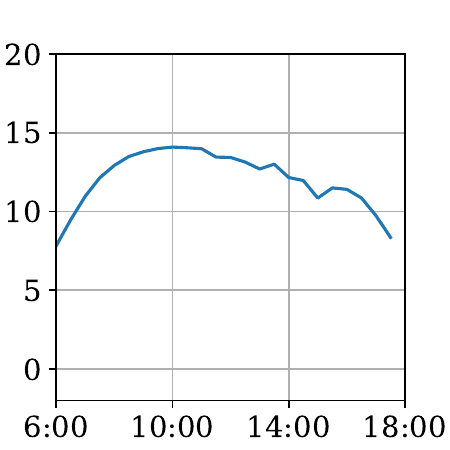}
        &
        \includegraphics[width=.35\linewidth]{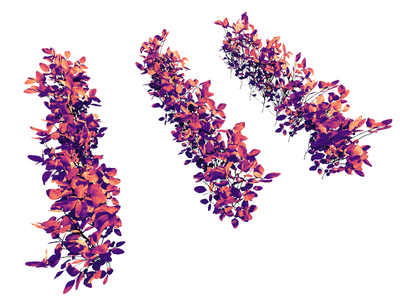}
        & 
        \includegraphics[width=.35\linewidth]{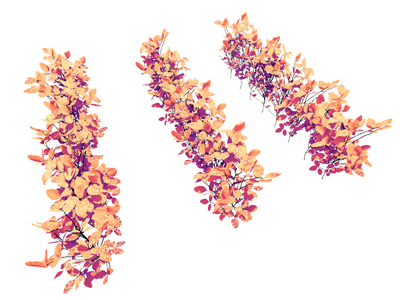}

        \\
    \end{tabular}
    
    }
\captionof{figure}{\textbf{Photosynthesis simulation results}. We perform simulations using Helios~\cite{bailey2019helios} on two soybean canopies generated by repeating Demeter-Soybean models. Left: timeseries of the net photosynthesis rate for the crop canopy over the course of a day, in units of $\mu$molCO$_2$/\SI{}{\meter}$^2$/\SI{}{\second}. Other columns: mesh visualization where each leaf face is colored according to the rate of photosynthesis over that face (brighter = higher rate). }

\label{fig:photosynthesis}
\end{table}
\paragraph{Biophysical Simulation Applications}
Demeter provides high-quality, realistic 3D meshes of plants that can be fed directly into simulation software to predict important agroecosystem variables. We showcase this capability by generating small crop fields by placing fitted Demeter models in a grid, and passing it to Helios~\cite{bailey2019helios} to simulate the response of the plant to weather variations over the course of a day. The weather variables were taken from data measured by a flux tower~\cite{pastorello2020fluxnet2015} and include temperature, humidity, radiation, and other environmental variables. In Fig.~\ref{fig:photosynthesis}, we visualize the output of the photosynthesis rate.










\paragraph{Limitations} 
We assume leaves are 2D shapes and stems are curves with uniform thickness. This excludes species with complex structures (e.g., cacti, algae, banyan trees) but applies to many common species.

\section{Conclusion}
We presented Demeter, a parametric shape model for plants using knowledge-based primitives and learning. Using a real-world soybean dataset, we show that our Demeter-Soybean model faithfully represents plant shapes and supports reconstruction and simulation applications. 


\section{Acknowledgment} This project is supported by NSF Awards \#1847334 \#2331878, \#2340254, \#2312102, \#2414227, and \#2404385. We greatly appreciate the NCSA for providing computing resources.

\clearpage
\setcounter{page}{1}
\maketitlesupplementary
\setcounter{section}{0}
\setcounter{footnote}{0}
\appendix

\begin{figure*}[h!]  
    \centering
    \includegraphics[width=\linewidth, trim=0 120 0 100, clip]{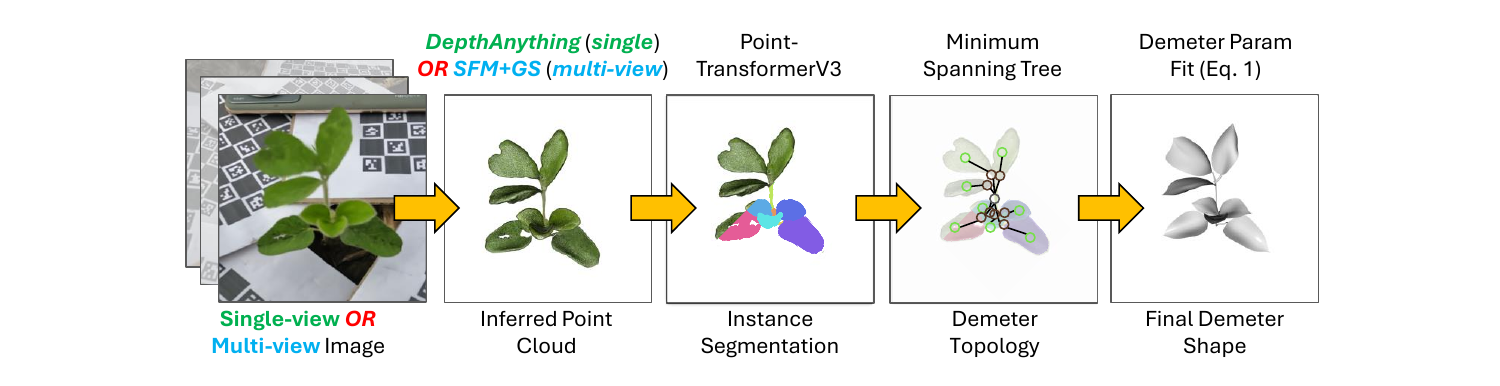}
    \caption{\textbf{Test-time Reconstruction Pipeline.} The pipeline transforms the input images to a segmented point cloud, extracts the topology and fits part templates, and finally outputs the parametric mesh.
    }
    \label{fig:infer_big}
\end{figure*}
\begin{figure}[t]
    \centering
    \includegraphics[width=0.8\linewidth]{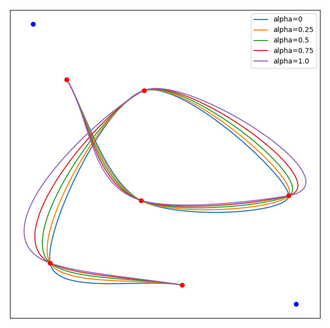}
    \caption{\textbf{Visualization of Catmull-Rom Curve.} The alpha parameter controls the curvature of the curve. We choose $\alpha=0.5$ in this paper. The red dots represent the control points and the blue ones represent the virtual control points, which is necessary to draw the first and last segment.}
    \label{fig:cm_curve}
\end{figure}
\begin{center}
     \includegraphics[width=.5\textwidth, trim=95 50 90 55, clip]{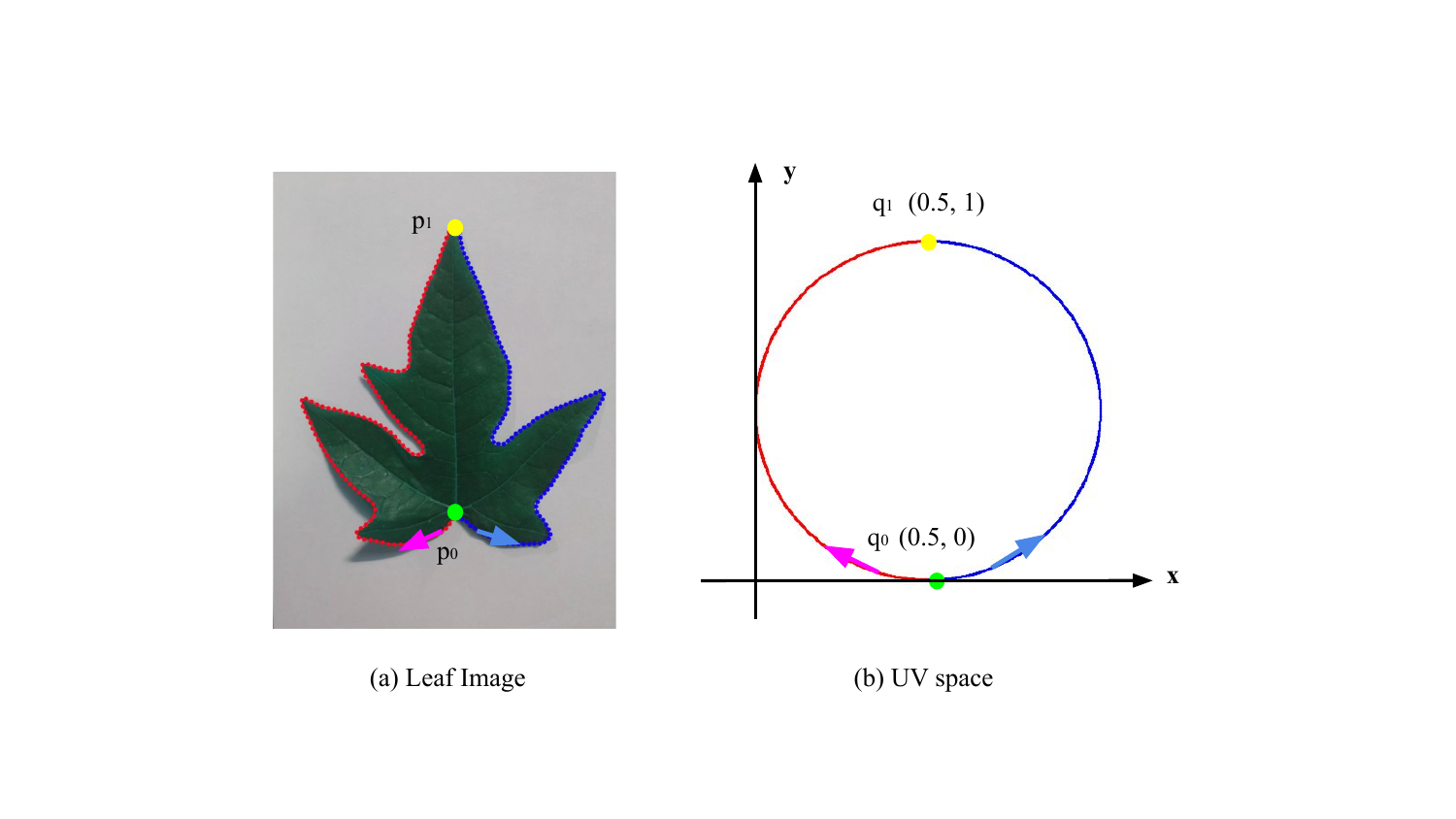}
    \captionof{figure}{\textbf{Initial value creation for Laplace equation.} We map the boundary value from the template to the leaf according to the curve length along the contour.
    }
    \label{fig:bound}
\end{center}

\begin{figure*}
    \centering
    \includegraphics[width=.9\textwidth]{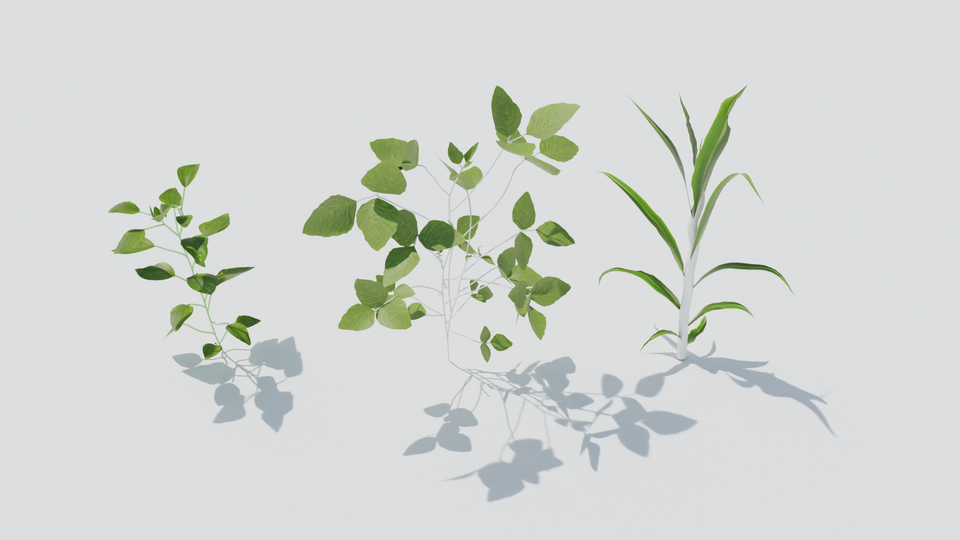}
    \captionof{figure}{\textbf{Render Demeter in Blender}. We could attach textures (Fig.~\ref{fig:square_texture}) to the leaf of our Demeter model and dump to rendering engine such as blender for photo-realistic rendering. From left to right, the image shows pepper, soybean and maize respectively.}
    \label{fig:blender}
\end{figure*}

\begin{figure*}[t]
\vspace{-3mm}
\centering
\def\arraystretch{0.3}
\setlength{\tabcolsep}{0.5pt}
\includegraphics[width=1.0\textwidth]{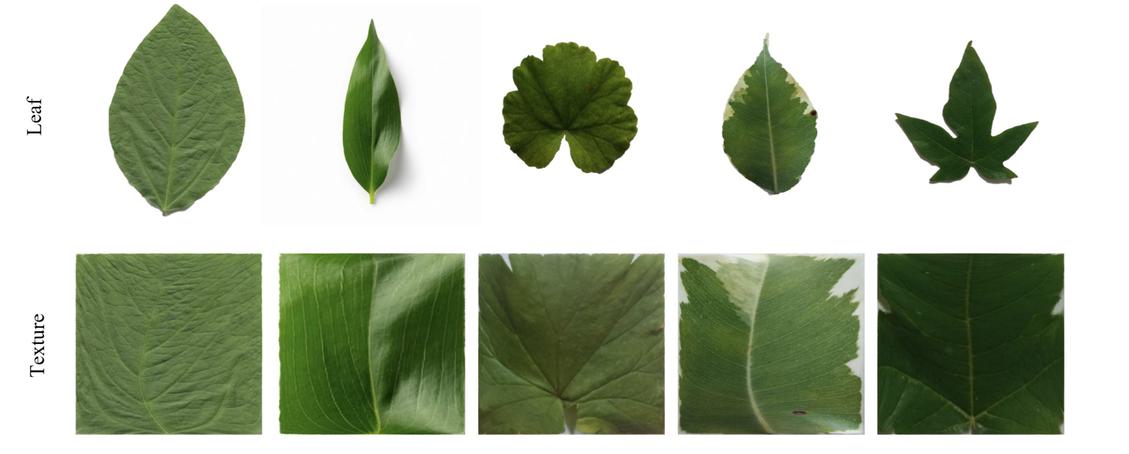}

\caption{\textbf{The texture extracted from leaf images using bijective mapping.} Similarly, we could map the leaf to a square UV space and get the texture, which can be easily applied to the grid points of our parametric leaf.}

\label{fig:square_texture}
\end{figure*}

\begin{table}[h]
\centering
\resizebox{0.9\linewidth}{!}{
\begin{tabular}{c|p{0.8\linewidth}}
\textbf{Parameter} & \textbf{Meaning} \\
\hline
$i$ & the index of node \\
$j$ & the index of control points within a node \\
\hline
$\mathcal{M}$ & the Demeter Model \\
$\mathbf{F}$ & the faces set of Demeter mesh \\
$\mathbf{V}$ & the vertices of Demeter mesh \\
$v_f$ & the vertices of the templates after shape offset and deformation, $v_f \in \mathbf{V}$ \\ 
$\mathbf{\Phi}$ & the parameter of Demeter model (PCA coefficient) \\
\hline
$\bGamma$ & topology \\
$\text{pa}(i)$ & the parent of the node $i$\\
$\text{ans}(j)$ & the parent of the control point $j$ (within node $i$)\\
\hline
$\btheta$ & articulation \\
$\btheta_i$ & the articulation of node $i$ \\
$\boldsymbol\tau_i$ & rotation of node $i$ \\
$d_i$ & length with respect to the parent stem of node $i$ \\
$s_i$ & scale of node $i$ \\
$\rm{T}$ & rigid + scale transformations \\
$\rm{\mathbf{T}}$ & accumulated rigid + scale transformations along the kinematic chain \\
\hline
$\bbeta$ & shape \\
$\bbeta_i$ & the shape parameter of node $i$ \\
$v_t$ & the vertices of the templates without deformation (canonical space) \\
$\mathcal{S}$ & shape operation to template vertices \\
$\mathbf{\Phi}_s$ & PCA of Shape \\
\hline
$\bgamma$ & deformation \\
$d_j$ & length with respect to parent of control point $j$ \\
$\boldsymbol\tau_j$ & rotation of control point $j$ \\
$v_j$ & vertex $j$ in node $i$ with shape offset \\
$\mathcal{D}$ & deformation operation to template vertices \\
$\mathbf{\Phi}_d$ & PCA of Deformation \\
\hline
$\text{tp}(i)$ & the type of the node $i$\\
$n$ & total number of nodes \\
$n_l$ & the number of leaf node \\
$n_s$ & the number of stem node \\
$n_o$ & the number of other node except leaf and stem \\
$m_{l_1}$ & the number of vertical control points of a leaf \\
$m_{l_2}$ & the number of horizontal control points of a leaf \\
$m_s$ & the number of control points of a stem \\
\hline
$\Omega$ & UV space\\
$\Omega'$ & leaf space\\
$\varphi$ & the mapping between UV space and leaf \\
$\partial$ & the boundary of domain \\
$b$ & the mapping between the boundary of UV space and leaf \\
\hline
$\mathcal{P}$ & the input point cloud \\

\end{tabular}
}
\caption{Explanation of Demeter model notation.}
\label{tab:parameters}
\end{table}

\begin{figure*}[h!]
\centering
\def\arraystretch{0.3}
\setlength{\tabcolsep}{0.5pt}
\resizebox{0.9\linewidth}{!}
{
\begin{tabular}{ccccccccc}
&{\small Sample 1}
&{\small Sample 2}
&{\small Sample 3}
&{\small Sample 4}
&{\small Sample 5}

\\
\raisebox{15mm}{\rotatebox{90}{{\small Input}}}
&\includegraphics[width=0.16\linewidth]{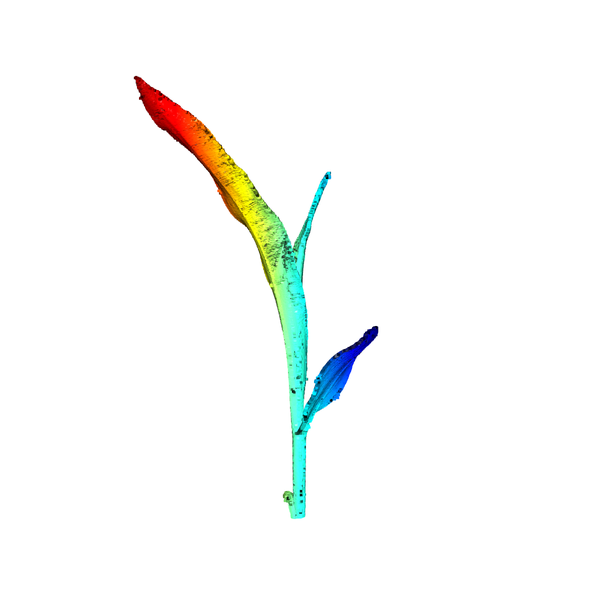}
&\includegraphics[width=0.16\linewidth]{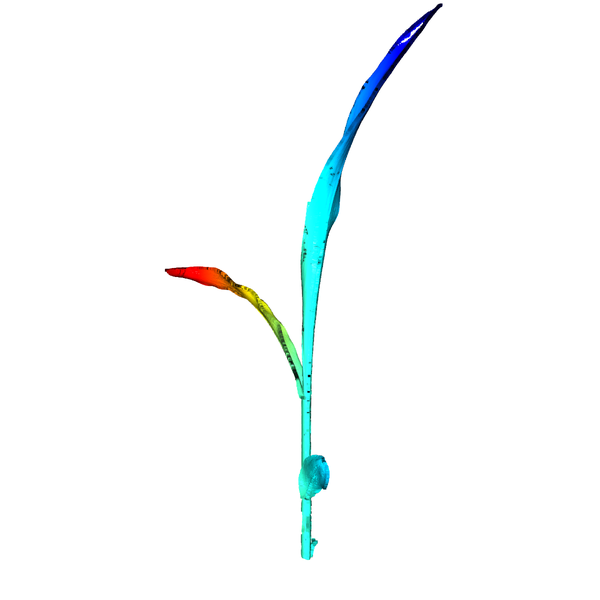}
&\includegraphics[width=0.16\linewidth]{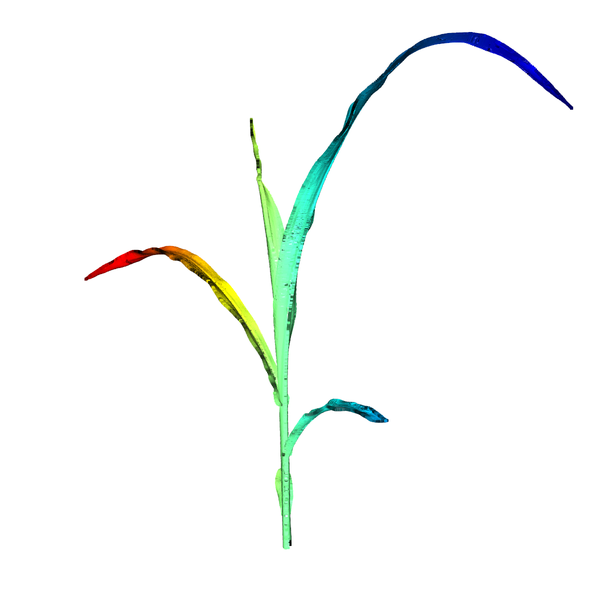}
&\includegraphics[width=0.25\linewidth]{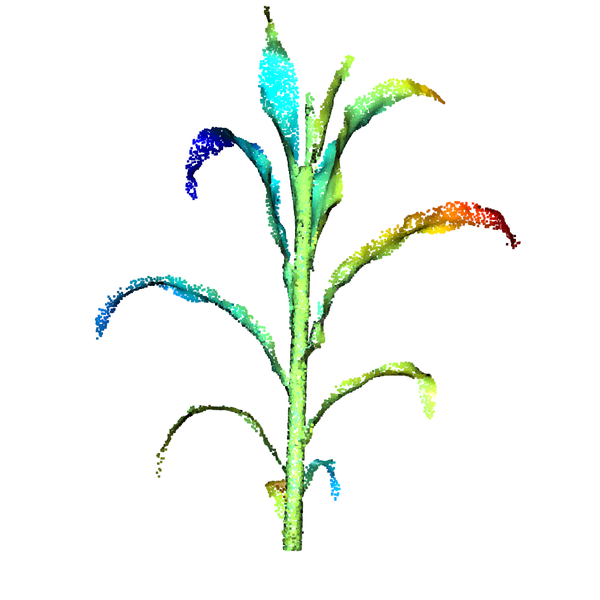}
&\includegraphics[width=0.25\linewidth]{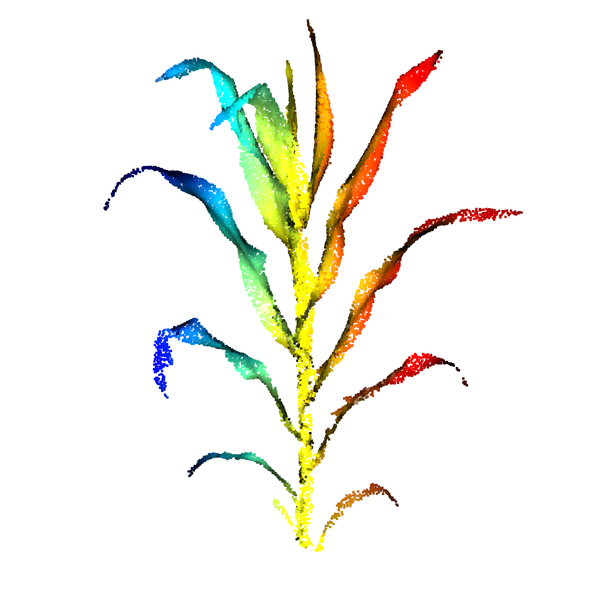}

\\
\raisebox{10mm}{\rotatebox{90}{{\small Pred Mesh}}}
&\includegraphics[width=0.16\linewidth]{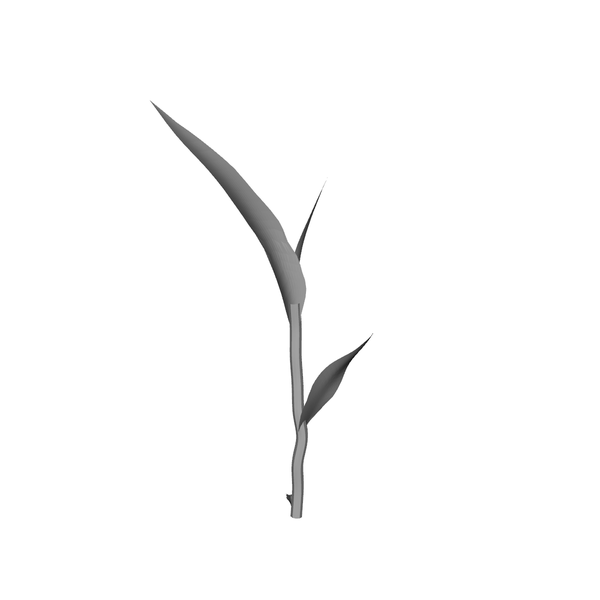}
&\includegraphics[width=0.16\linewidth]{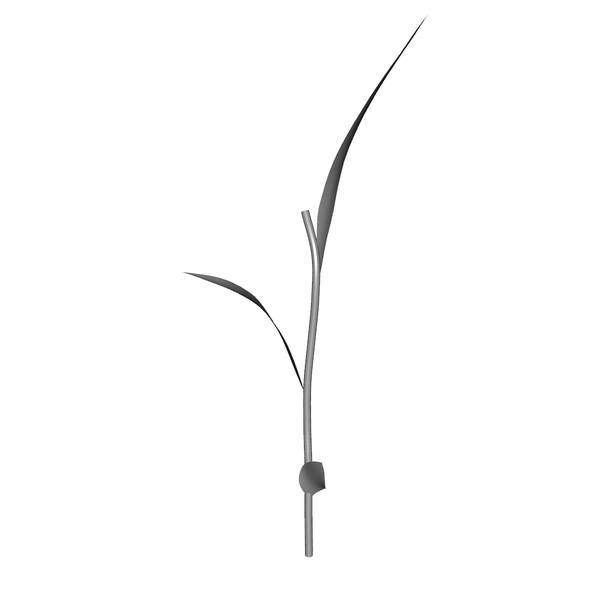}
&\includegraphics[width=0.16\linewidth]{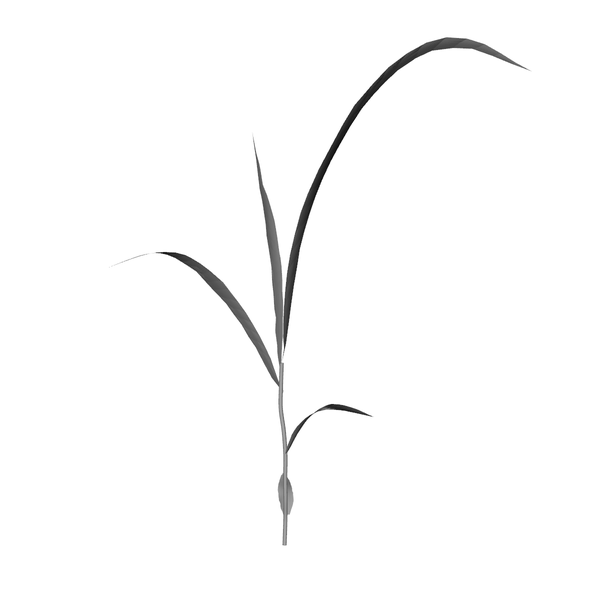}
&\includegraphics[width=0.25\linewidth]{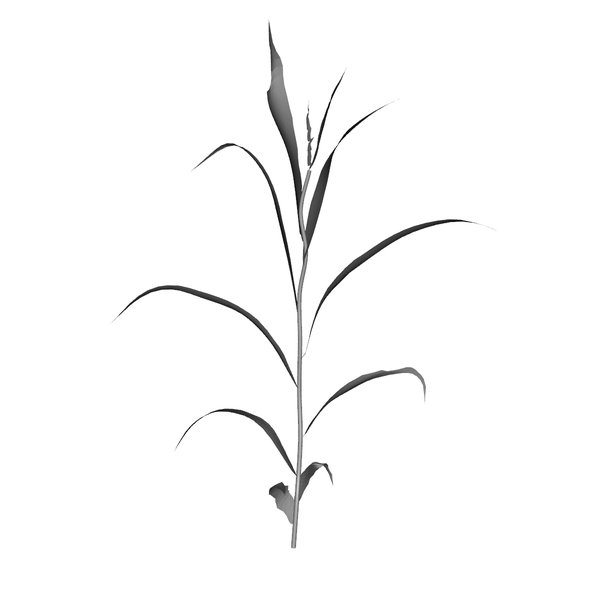}
&\includegraphics[width=0.25\linewidth]{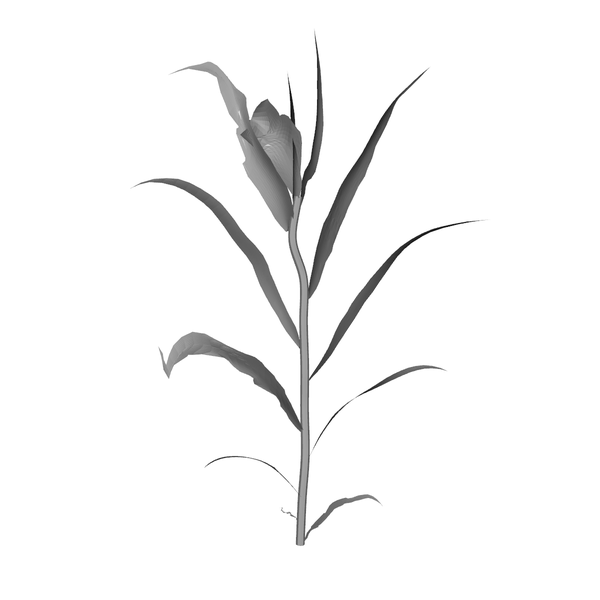}

\\
\raisebox{10mm}{\rotatebox{90}{{\small Pred Instances}}}
&\includegraphics[width=0.16\linewidth]{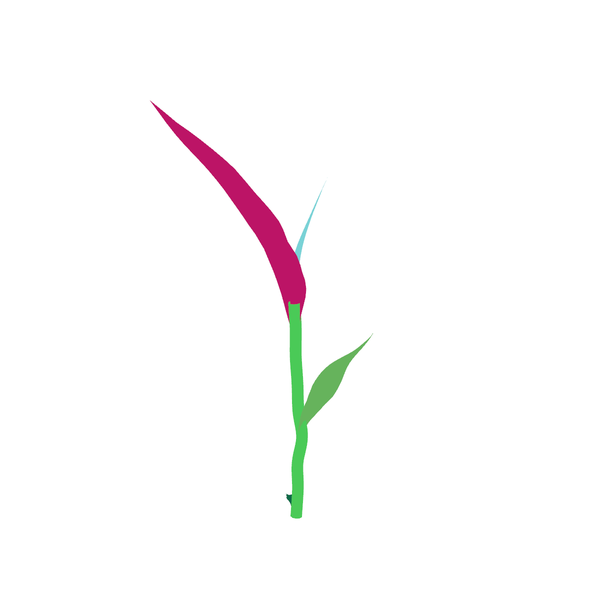}
&\includegraphics[width=0.16\linewidth]{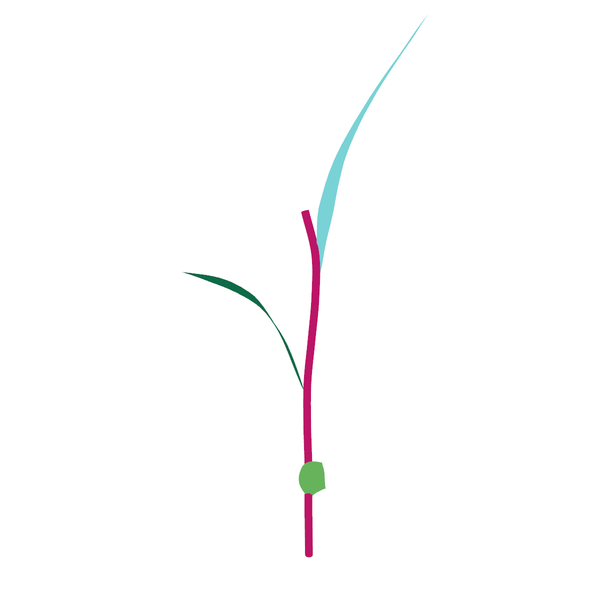}
&\includegraphics[width=0.16\linewidth]{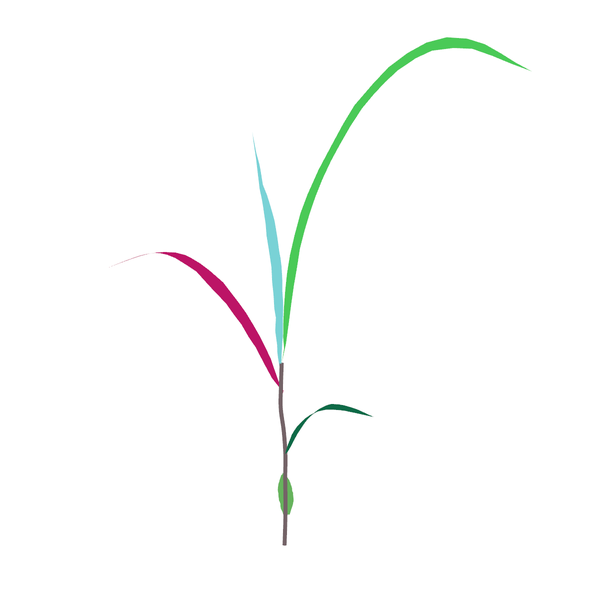}
&\includegraphics[width=0.25\linewidth]{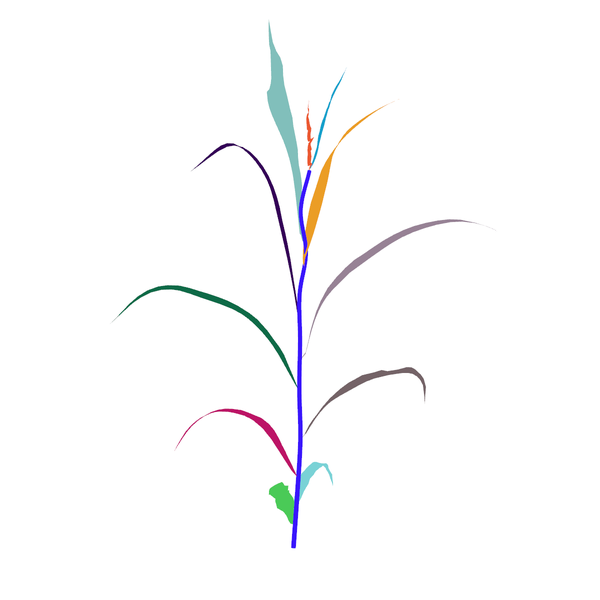}
&\includegraphics[width=0.25\linewidth]{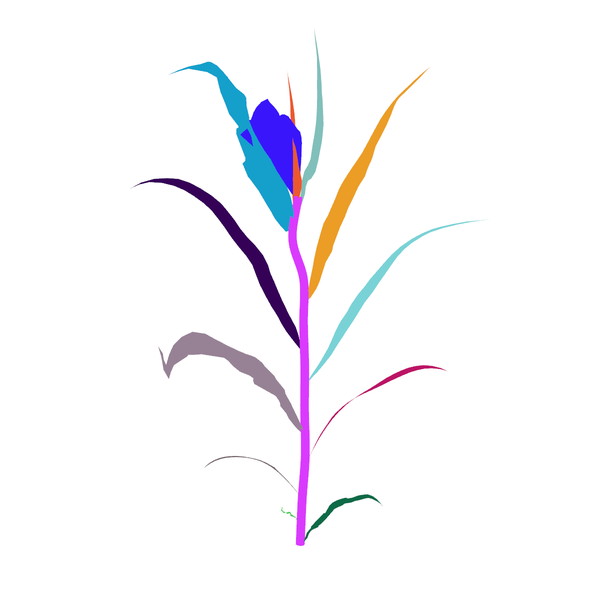}

\end{tabular}

}
\caption{\textbf{Qualitative results of reconstructing maize.} We show the results of fitting Demeter-Maize models to point clouds from the Pheno4D dataset (sample 1,2 and 3) and web dataset \cite{Zhang2022} (sample 4 and sample 5). Different colors represent different instances. The results show that our model can generalize to species beyond soybeans and capture shape details. 
}
\label{fig:3d_recon_maize}
\end{figure*}





\begin{figure}[t]
\centering
\def\arraystretch{0.3}
\setlength{\tabcolsep}{0.5pt}
\resizebox{\linewidth}{!}{
\begin{tabular}{ccccc}

&{\small 2D Comp. 1}
&{\small 2D Comp. 2}
&{\small 2D Comp. 3}
&{\small 2D Comp. 4} 

\\
\raisebox{5mm}{\rotatebox{90}{{\small $-3\sigma$}}}
&\includegraphics[width=0.26\linewidth]{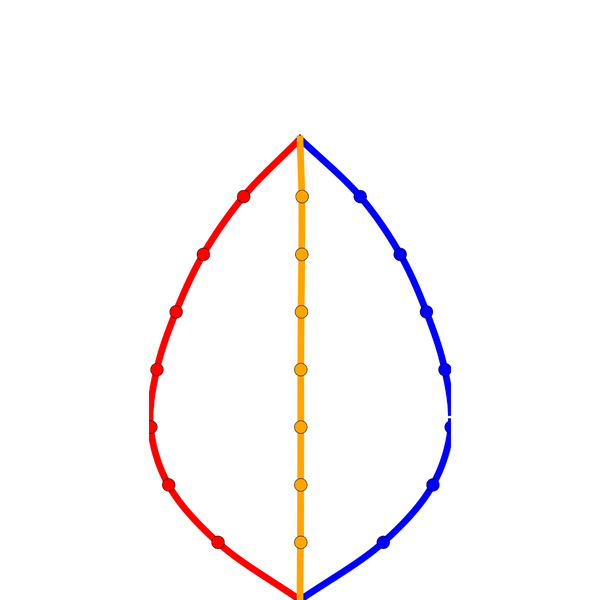}
&\includegraphics[width=0.26\linewidth]{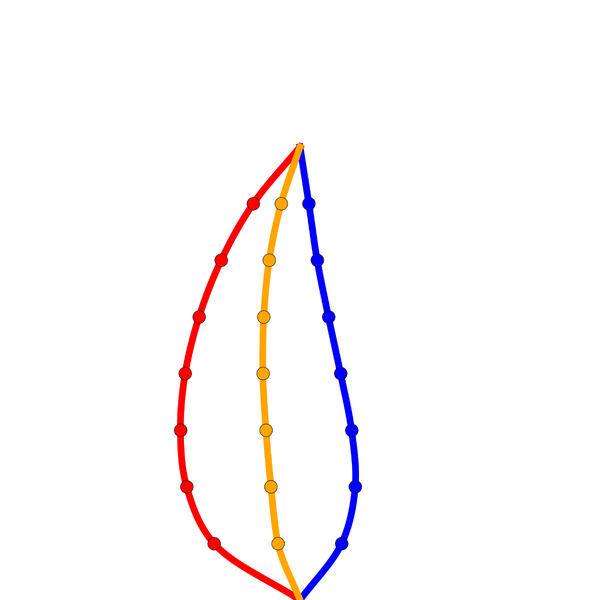}
&\includegraphics[width=0.26\linewidth]{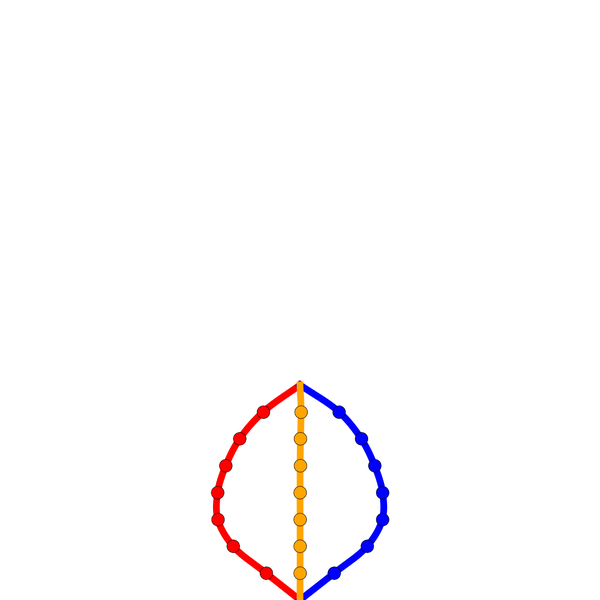}
&\includegraphics[width=0.26\linewidth]{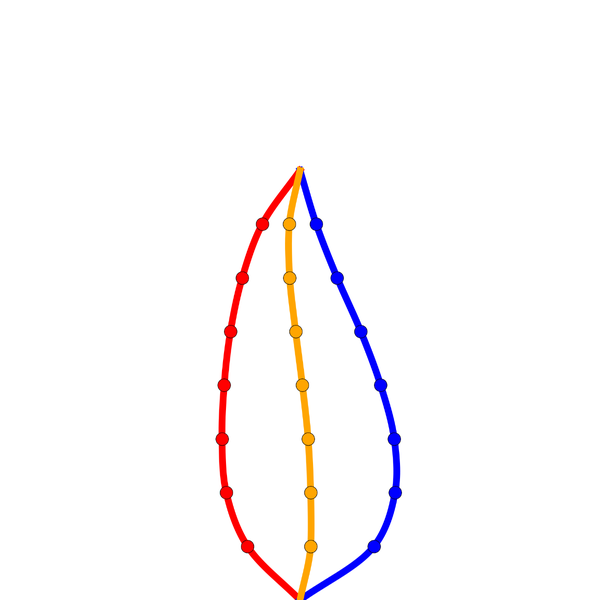}

\\
\raisebox{1mm}{\rotatebox{90}{{\small mean shape}}}
&\includegraphics[width=0.26\linewidth]{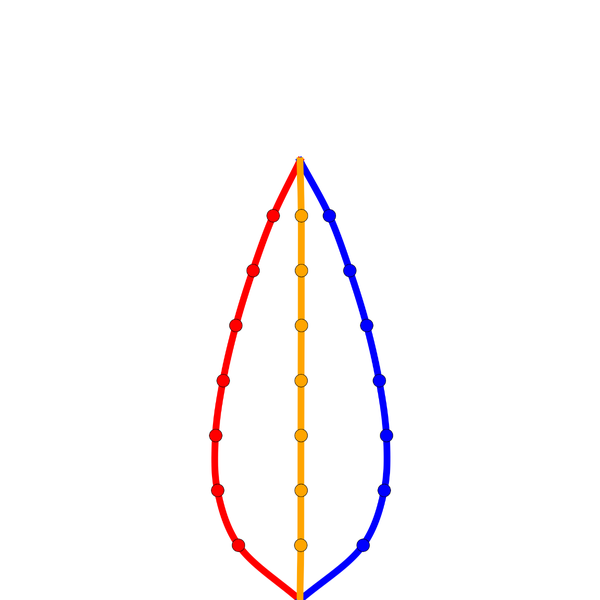}
&\includegraphics[width=0.26\linewidth]{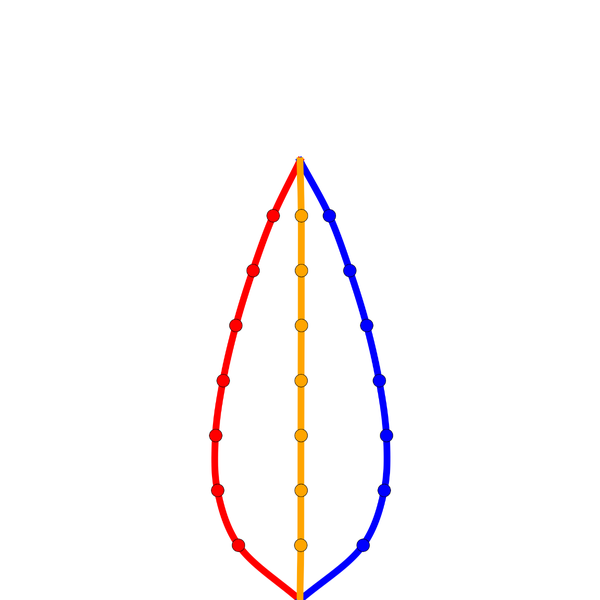}
&\includegraphics[width=0.26\linewidth]{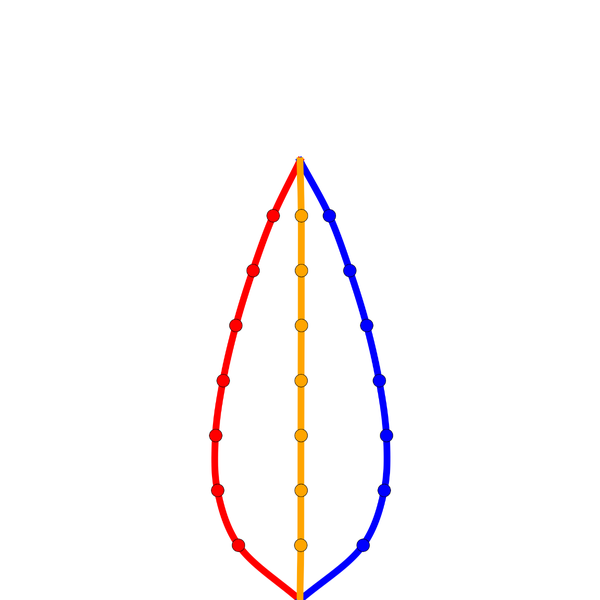}
&\includegraphics[width=0.26\linewidth]{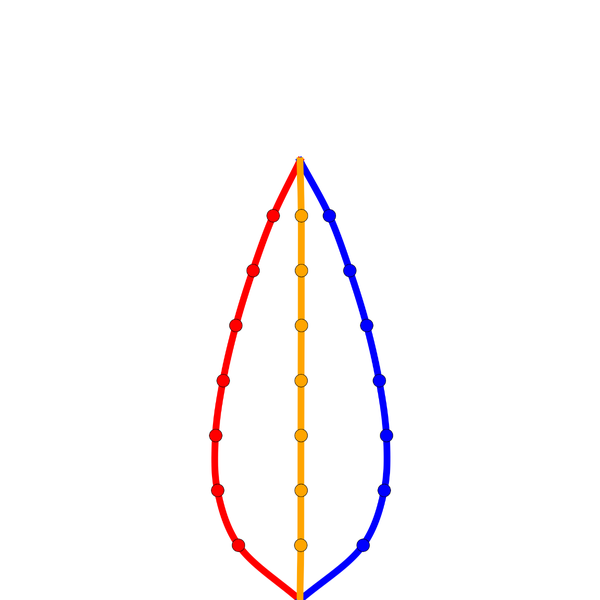}

\\
\raisebox{5mm}{\rotatebox{90}{{\small $+3\sigma$}}}
&\includegraphics[width=0.26\linewidth]{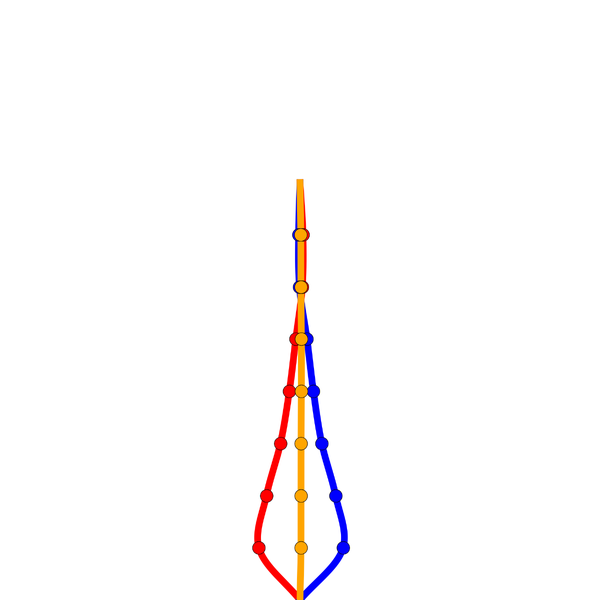}
&\includegraphics[width=0.26\linewidth]{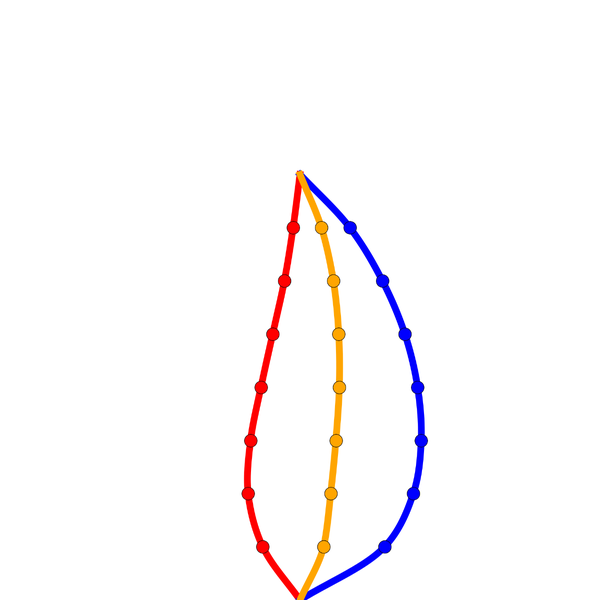}
&\includegraphics[width=0.26\linewidth]{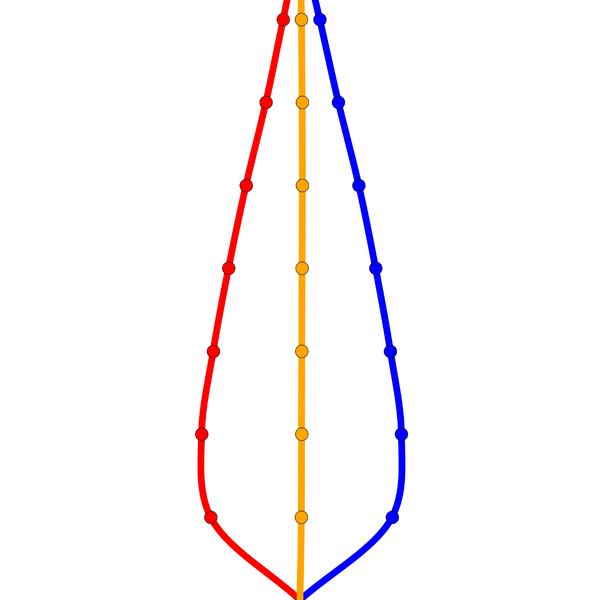}
&\includegraphics[width=0.26\linewidth]{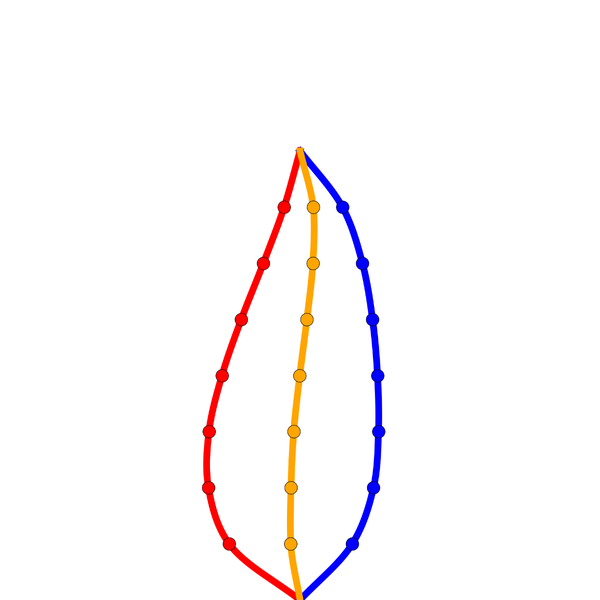}

\end{tabular}

}
\caption{\textbf{PCA coefficient of soybean leaf.} We visualize the first several principle component of leaf for soybean in $(-3\sigma, 3\sigma)$.}
\label{fig:leaf_pca_viz_soybean}
\end{figure}
\begin{figure}[t]
\centering
\def\arraystretch{0.3}
\setlength{\tabcolsep}{0.5pt}
\resizebox{\linewidth}{!}{
\begin{tabular}{ccccccc}

&{\small 2D Comp. 1}
&{\small 2D Comp. 2}
&{\small 2D Comp. 3}

\\
\raisebox{2mm}{\rotatebox{90}{{\small $-3\sigma$}}}
&\includegraphics[width=0.35\linewidth]{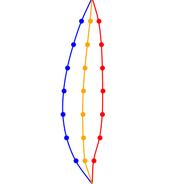}
&\includegraphics[width=0.35\linewidth]{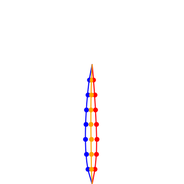}
&\includegraphics[width=0.35\linewidth]{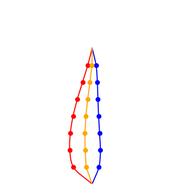}

\\
\raisebox{3mm}{\rotatebox{90}{{\small $0$}}}
&\includegraphics[width=0.35\linewidth]{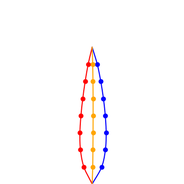}
&\includegraphics[width=0.35\linewidth]{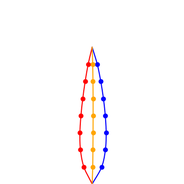}
&\includegraphics[width=0.35\linewidth]{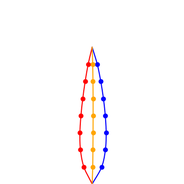}

\\
\raisebox{2mm}{\rotatebox{90}{{\small $+3\sigma$}}}
&\includegraphics[width=0.35\linewidth]{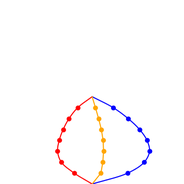}
&\includegraphics[width=0.35\linewidth]{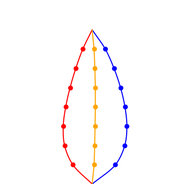}
&\includegraphics[width=0.35\linewidth]{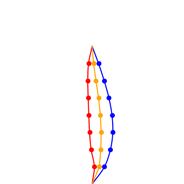}

\end{tabular}

}
\caption{\textbf{PCA coefficients of Demeter-Maize leaf shape.} We vary the PCA coefficients for maize leaf shape across $(-3\sigma, 3\sigma)$ with respect to the mean shape for all 3 components. The red and blue points represent the control points for the 2D contour and the orange curve represents the initial skeleton for 3D deformation.}
\label{fig:maize_leaf_pca_viz}
\end{figure}



\section{Catmull-Rom Curve}

Given a shape template with the skeleton and deformation, we generate a smooth geometry with Catmull-Rom curve. A general Catmull-Rom curve is composed of $K+2$ control points $\{{\boldsymbol{p}}\}_{i=0}^{K+2}$ and $K+1$ segments. The first and the last control point $p_0,p_{K+1}$ are virtual control points, which depend on other control points, i.e. $p_0=p_1-(p_2-p_1)$, $p_{K+1} =p_{K}+(p_K-p_{K-1})$, as shown in the blue dots in Fig. \ref{fig:cm_curve}. 

Each segment $\mathbf{p}(t)$ in time $t\in[0,1]$ is a cubic polynomial curve and defined by the consecutive four control points: $\mathbf{p}_{i-2}$, $\mathbf{p}_{i-1}$, $\mathbf{p}_{i}$, and $\mathbf{p}_{i+1}$, and a shape parameter $\alpha$. We set $\alpha = 0.5$ for all segments in this paper. Each segment satisfies $\mathbf{p}(0) = \mathbf{p}_{i-1}$ and $\mathbf{p}(1) = \mathbf{p}_{i}$. Assuming $\mathbf{p}(t) = \left[ \mathbf{p}_{i-2}, \mathbf{p}_{i-1}, \mathbf{p}_{i}, \mathbf{p}_{i+1} \right] \mathbf{C}$, the blending weights $\mathbf{C}(t)$ are given by:
\begin{equation}
    \mathbf{C}(t) =  \left[\begin{array}{c}
-\alpha t + 2 \alpha t^2 - \alpha t^3 \\
1 + (\alpha - 3) t^2 + (2 - \alpha) t^3 \\
\alpha t + (3 - 2 \alpha) t^2 + (\alpha - 2) t^3 \\
-\alpha t^2 + \alpha t^3
\end{array}\right].
\end{equation}

\section{Detailed explaination of reconstruction pipeline}

~Fig.~\ref{fig:infer_big} depicts our single- or multi-image inference pipeline. 
\paragraph{3D Geometry}
We first acquire a point cloud—either by estimating camera intrinsics and depth with PerspectiveField and DepthAnything on a single view and lifting to a partial cloud, or by running SfM+GSplats on multi-view images to obtain a full cloud. 

\paragraph{Segmentation}
We then perform instance segmentation—using MaskRCNN on the single view and unprojecting masks to 3D, or using PointTransformer-v3 (PT-v3) on the multi-view cloud to predict per-point categories and inverse-distance scores, removing high-score points and clustering with DBSCAN. 

\paragraph{Tree-structure recovery}
We construct a dense directed graph $G$ over the estimated instances. The edge weights \( W_{ij} \) enforce the constraint that leaves have no children: $W_{ij} = 0$ if $i = j$; $W_{ij} = \infty$ if cluster $i$ is leaf; otherwise $W_{ij} = r_{ij}$, where $r_{ij}$ is the min distance of any points from cluster $i$ to $j$. Then, we compute the minimum spanning tree (MST) over $G$ starting from the root stem, which yields the plant skeleton with the minimum connection cost.

\paragraph{Demeter Fitting}
Finally, given the estimated topology, seg, and point cloud, we fit each cluster with a leaf or stem template, then run a global optimization over all parts parameters by minimizing Chamfer distance between the parametric shape and input cloud, producing the final Demeter parametric mesh.

\section{Additional Model Details}

\paragraph{Topology} 
Not every arbitrary tree structure corresponds to a valid plant.  
Most plants exhibit strong morphological constraints. For example, each plant may have a maximum depth or width and exhibit some first-order constraints, including the following: leaves cannot possess child nodes, and for soybeans, top-canopy stems often have a triplet of leaves as children. We enforce the former constraint by pruning edges with leaves as parents before constructing the minimal spanning tree. Incorporation of further constraints is left for future work.  

\paragraph{Articulation}

\begin{figure}[t]
  \centering
  \includegraphics[width=0.6\linewidth]{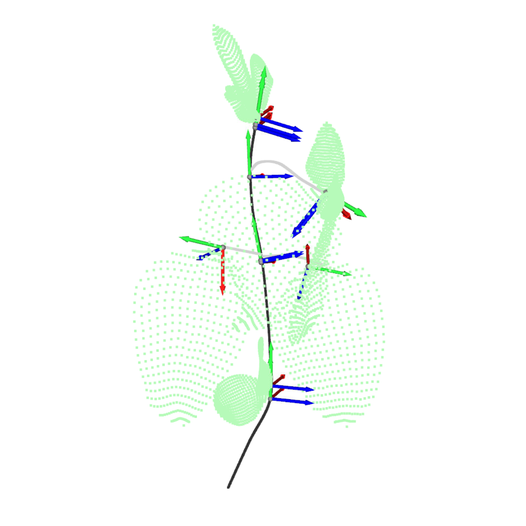}
  \caption{\textbf{Local coordinate system on the stem.} Each point on the curve is defined in a local coordinate system determined by the curve's intrinsic geometry at that specific point.}
  \label{fig:parent_coord_sys}
\end{figure}

Each node, except the root, is connected to its parent stem. The articulation of each node determines the rigid transformation relative to the local coordinate system at the connection point, as shown in Fig. \ref{fig:parent_coord_sys}. Given an arbitrary stem, the coordinate system depends only on the stem shape and the position of the connection point, and is independent of the global 6 DoF pose of the stem.

Given $m_s$ control points $\{v\}_{j=1}^{m_s}$ on the stem, we calculate the relative rotation $\{R_j\}_{j=1}^{m_s-1}$ of control points from each segment $v_{j+1}-v_j$, where $j=1,2,...,m_s-1$, and the y-axis of $R_0$ is aligned with the first segment. The points in each segment will have the same local coordinate as their preceding control point.

\paragraph{Shape}

We solve Eq. 6 in the main paper to get the bijective mapping between the leaf and template. To get the initial boundary value, we assign the leaf bottom $p_0$ to $q_0 =(0.5, 0)$, and the leaf tip $p_1$ to $q_1 =(0.5, 1.0)$. For other points along the contour, we uniformly map the boundary value according to the curve length of the point, as shown in Fig.~\ref{fig:bound}.

\begin{figure}[t]
\centering
\def\arraystretch{0.3}
\setlength{\tabcolsep}{0.5pt}
\resizebox{\linewidth}{!}{
\begin{tabular}{ccccccccc}
&{\small Input pcd} 
&{\small Dist.}
&{\small Inv Dist.}

\\
&\includegraphics[width=0.35\linewidth]{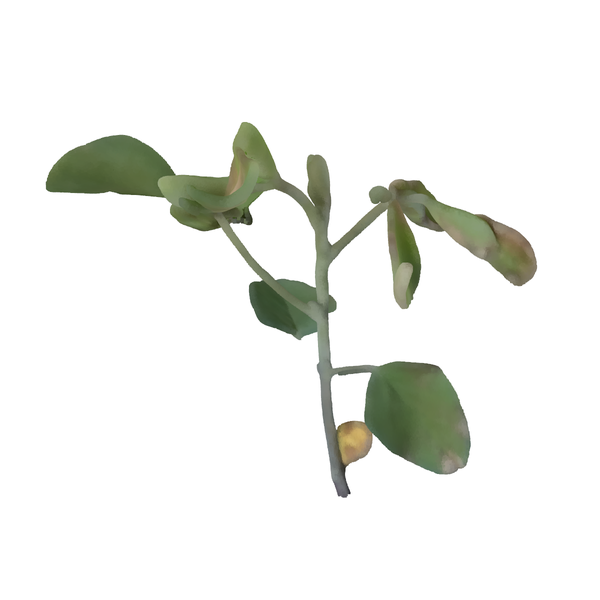}
&\includegraphics[width=0.35\linewidth]{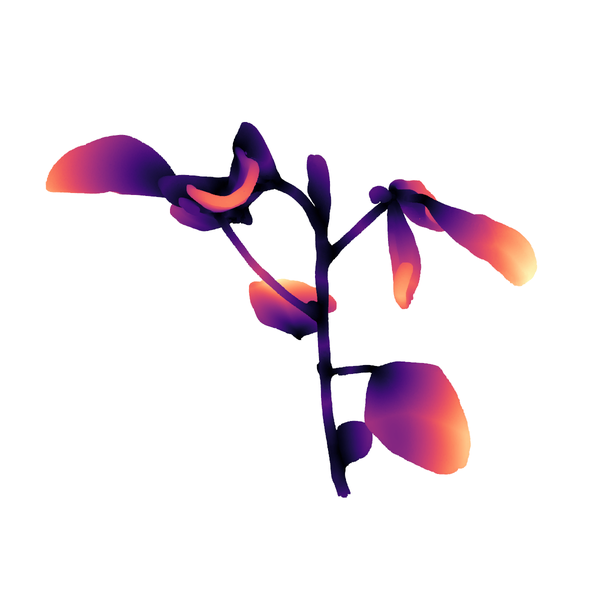}
&\includegraphics[width=0.35\linewidth]{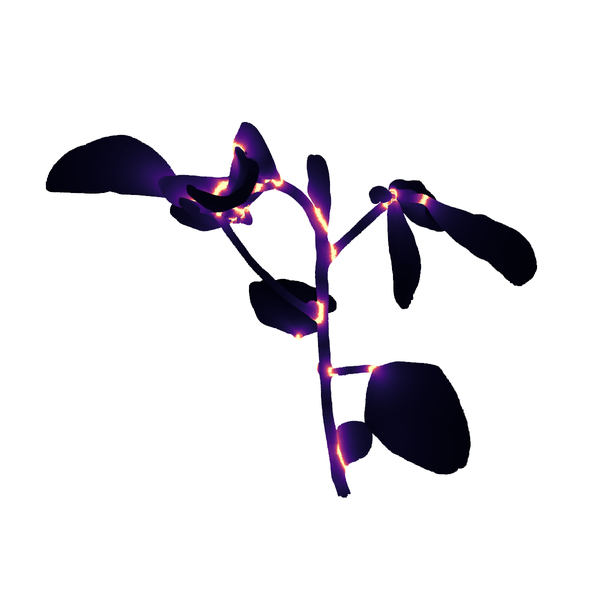}

\\
&\includegraphics[width=0.35\linewidth]{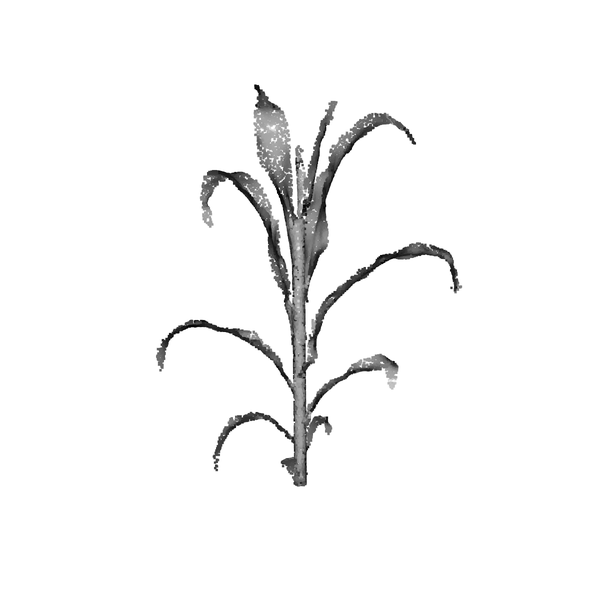}
&\includegraphics[width=0.35\linewidth]{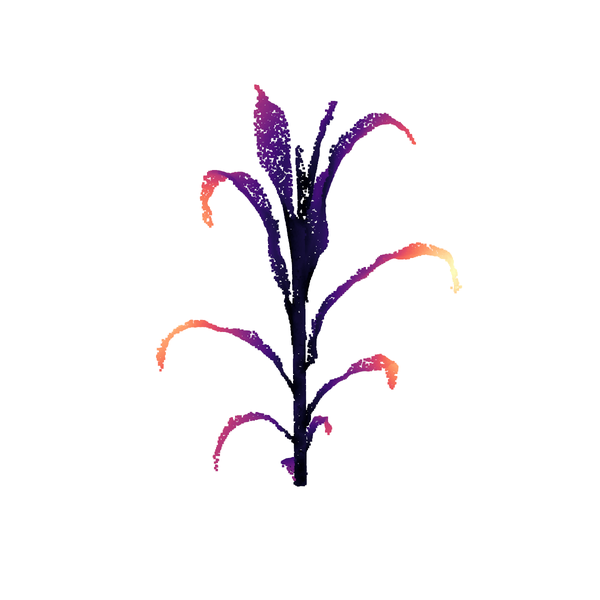}
&\includegraphics[width=0.35\linewidth]{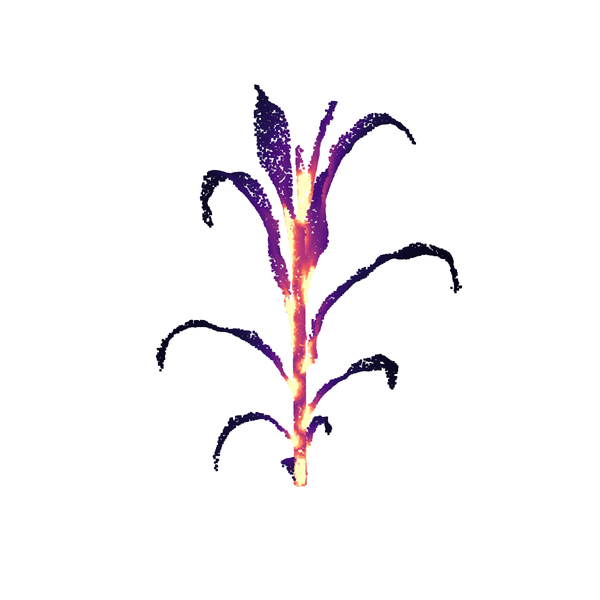}

\end{tabular}

}
\caption{\textbf{Truncated Inverse Distance.} Given a input point cloud with instance segmentation, we select the point in every instance and calculate the distance to all other instances for that point (Middle). Afterwards, we select a threshold and calculate the inverse distance and train PointTransformer to predict this value.
}
\label{fig:ptv3_inv_dist}
\end{figure}

\section{Additional Results and Visualizations}

\begin{figure*}[t]
\centering
\def\arraystretch{0.3}
\setlength{\tabcolsep}{0.5pt}
\resizebox{\linewidth}{!}{
\begin{tabular}{ccccccccc}
&{\small Input} 
&{\small Mesh} 
&{\small Semantics}
&{\small Instances}
&{\small Topology}

\\
\raisebox{7mm}{\rotatebox{90}{{\small Ground truth}}}
&\includegraphics[width=0.25\linewidth]{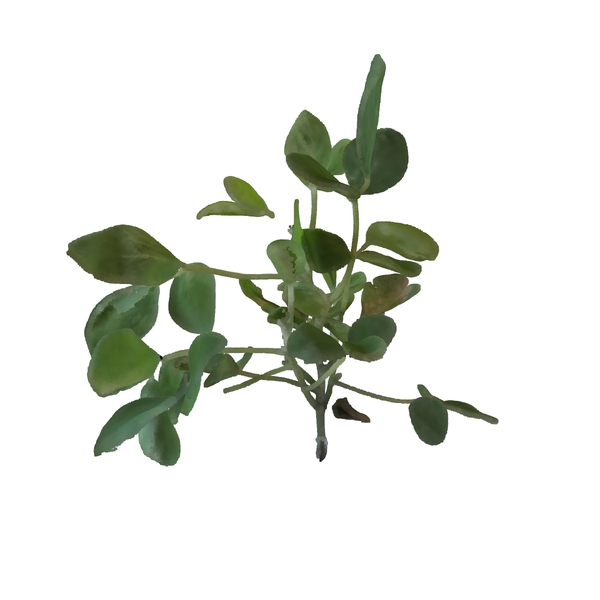}
&\includegraphics[width=0.25\linewidth]{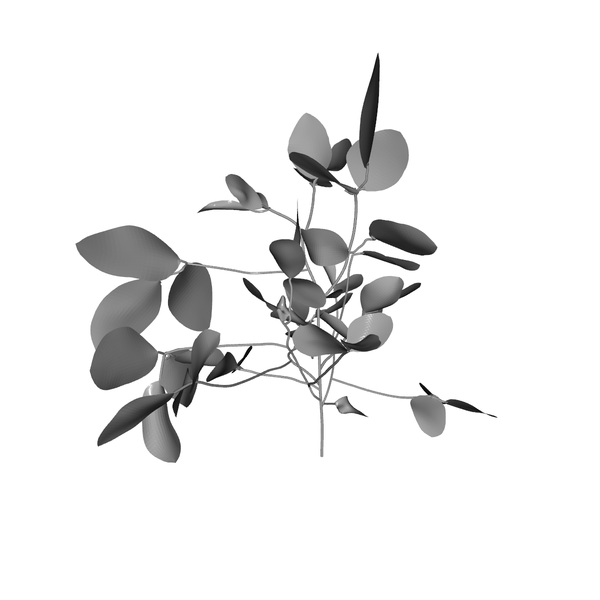}
&\includegraphics[width=0.25\linewidth]{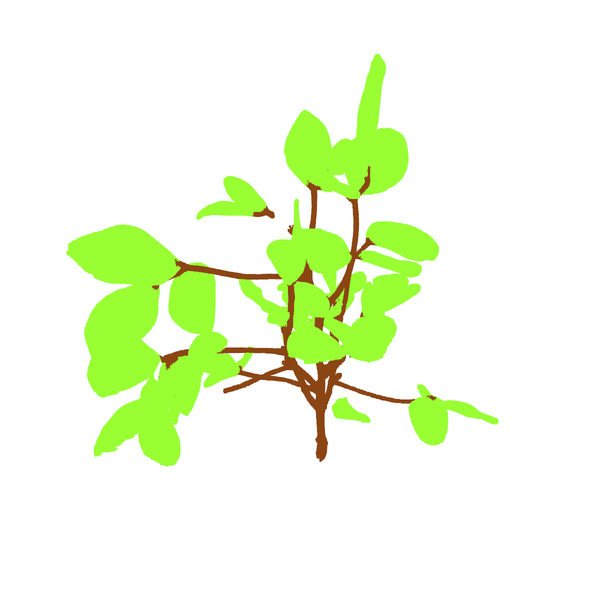}
&\includegraphics[width=0.25\linewidth]{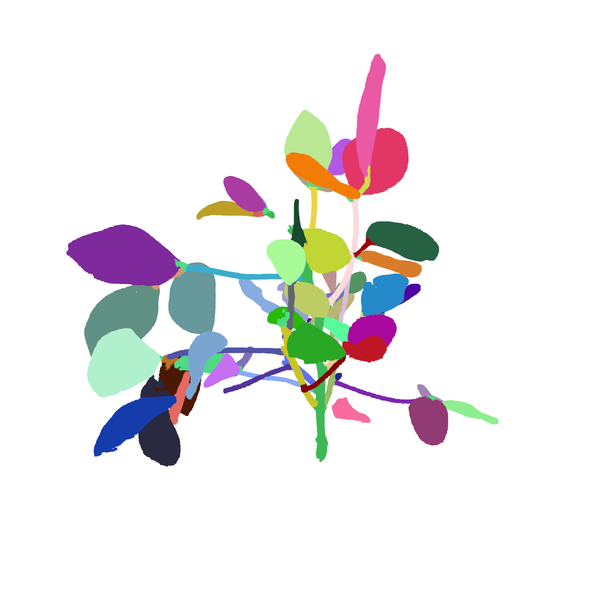}
&\includegraphics[width=0.25\linewidth]{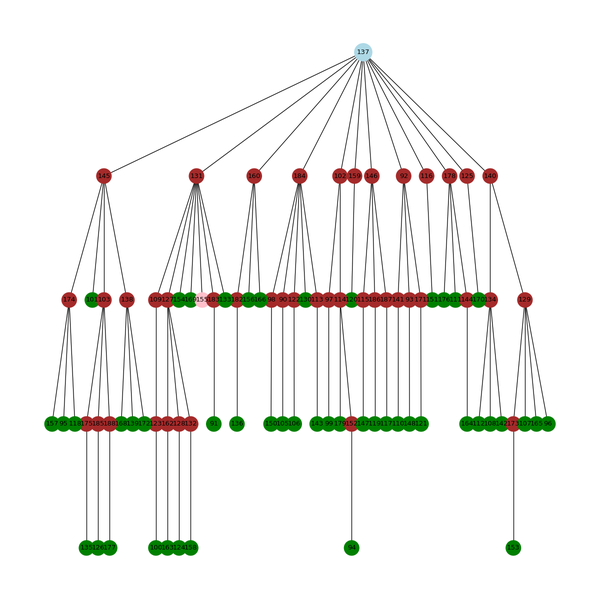}

\\
\raisebox{7mm}{\rotatebox{90}{{\small Prediction}}}
&\includegraphics[width=0.25\linewidth]{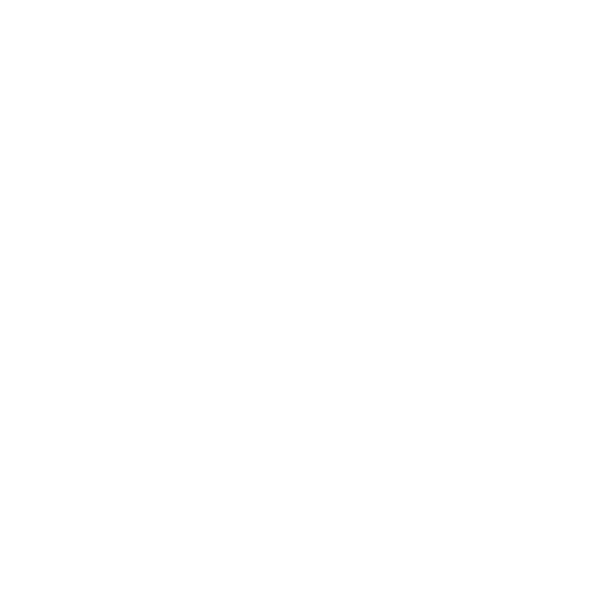}
&\includegraphics[width=0.25\linewidth]{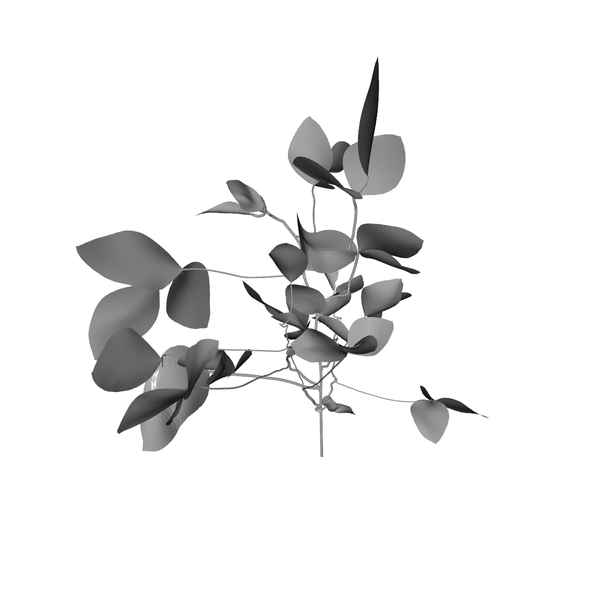}
&\includegraphics[width=0.25\linewidth]{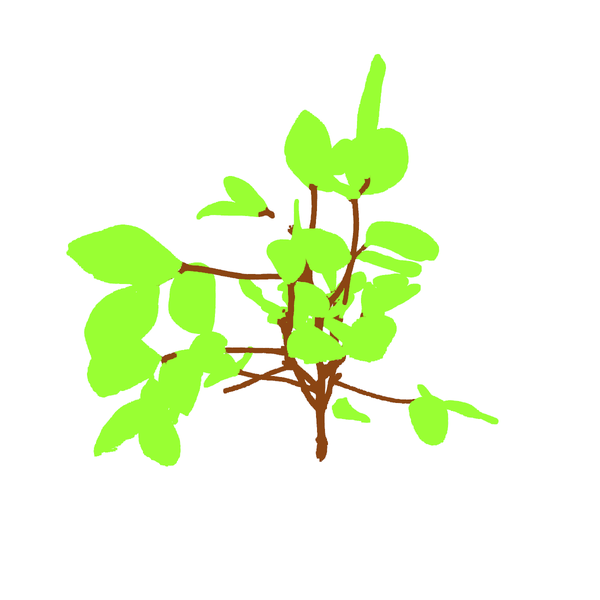}
&\includegraphics[width=0.25\linewidth]{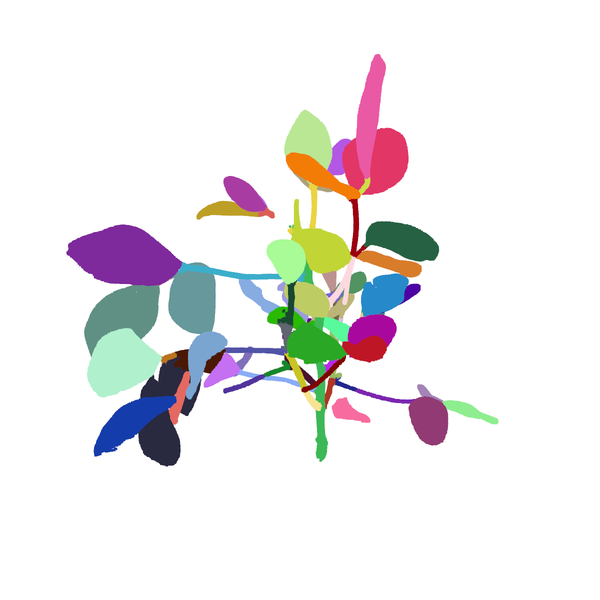}
&\includegraphics[width=0.25\linewidth]{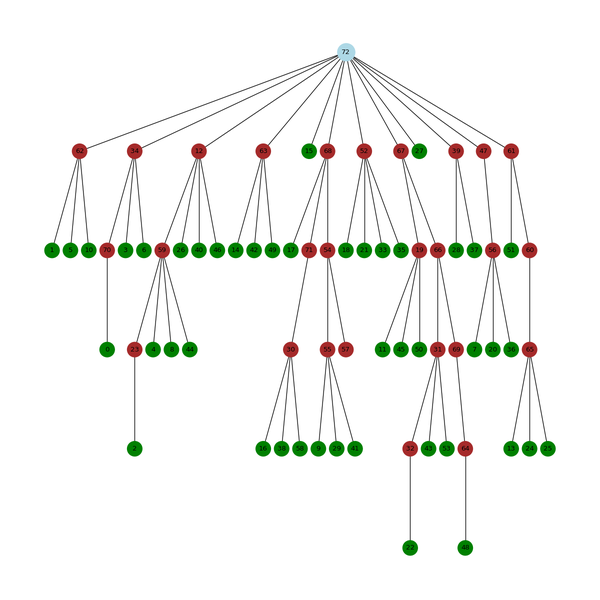}

\\
\raisebox{7mm}{\rotatebox{90}{{\small Ground truth}}}
&\includegraphics[width=0.25\linewidth]{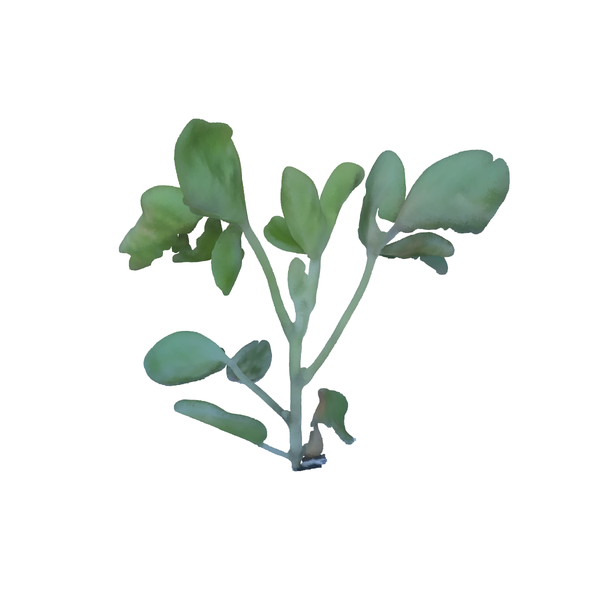}
&\includegraphics[width=0.25\linewidth]{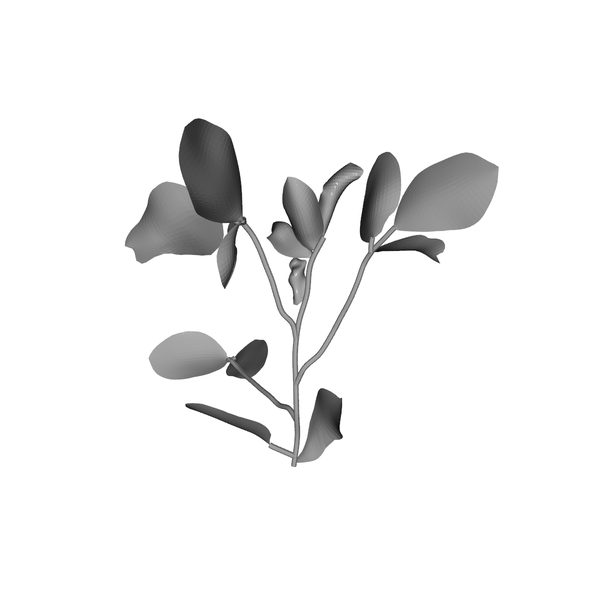}
&\includegraphics[width=0.25\linewidth]{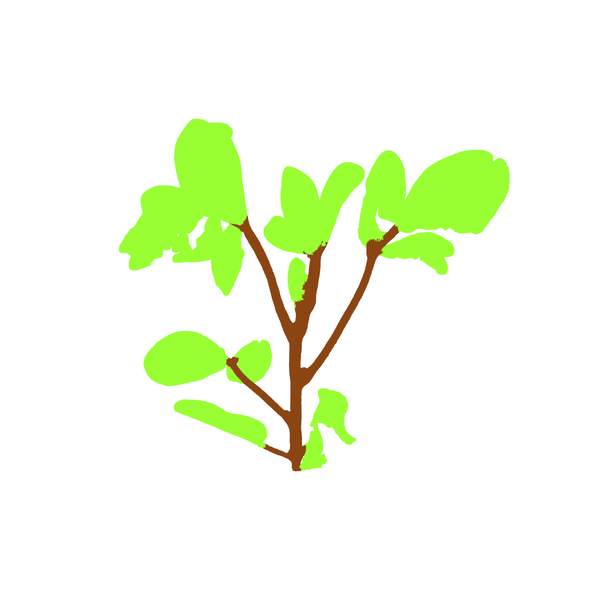}
&\includegraphics[width=0.25\linewidth]{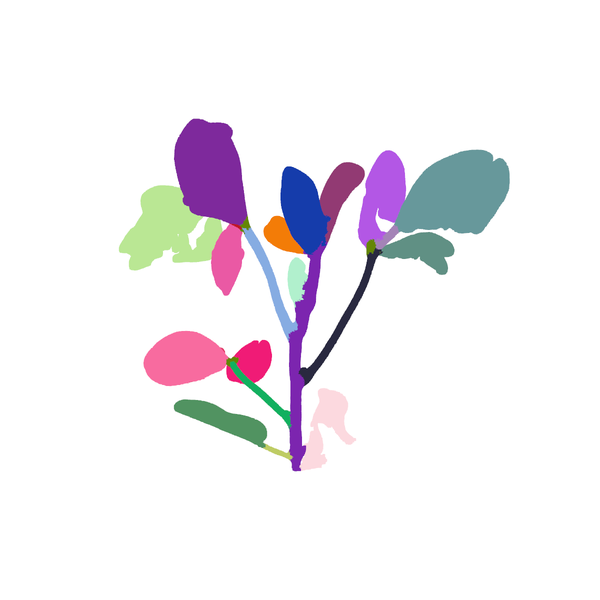}
&\includegraphics[width=0.25\linewidth]{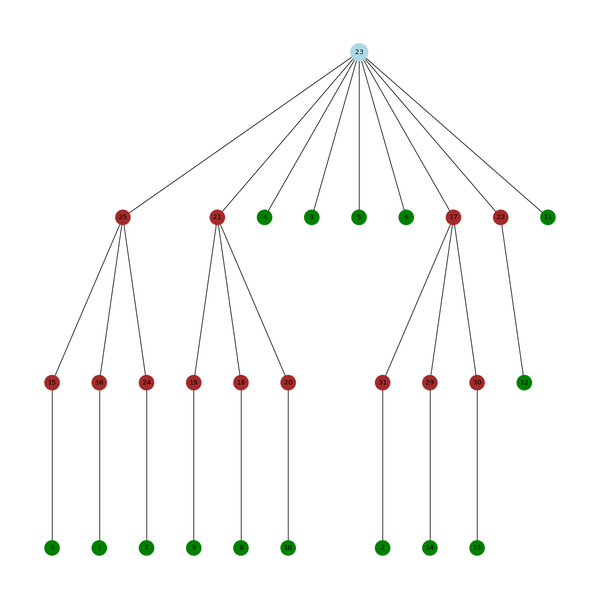}

\\
\raisebox{7mm}{\rotatebox{90}{{\small Prediction}}}
&\includegraphics[width=0.25\linewidth]{figure/empty.png}
&\includegraphics[width=0.25\linewidth]{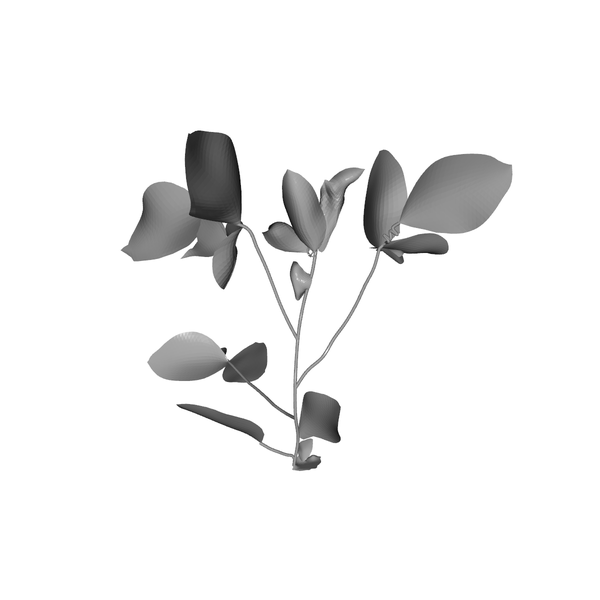}
&\includegraphics[width=0.25\linewidth]{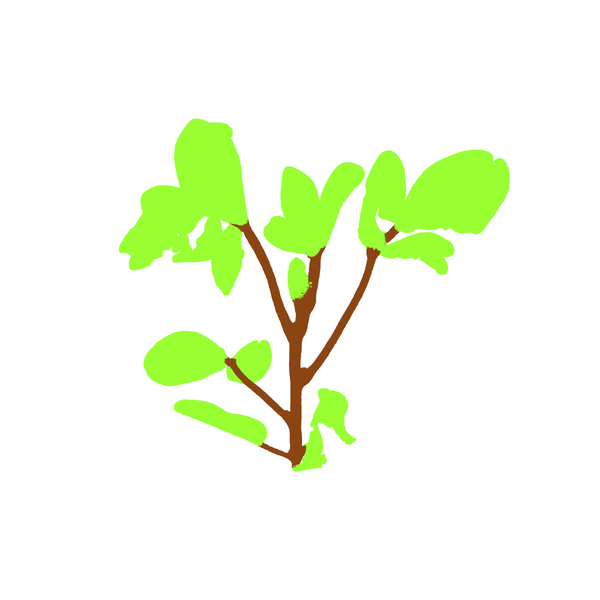}
&\includegraphics[width=0.25\linewidth]{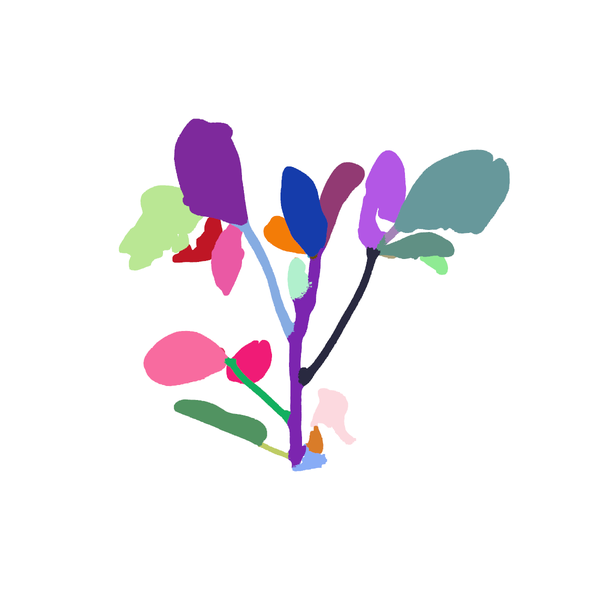}
&\includegraphics[width=0.25\linewidth]{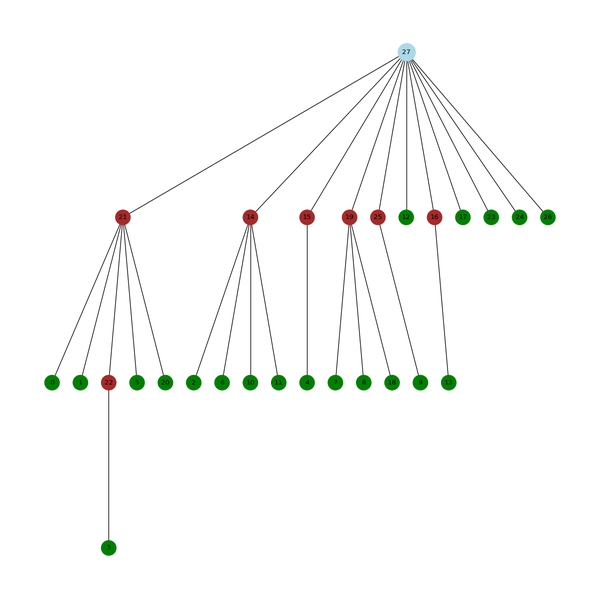}

\\
\raisebox{7mm}{\rotatebox{90}{{\small Ground truth}}}
&\includegraphics[width=0.25\linewidth]{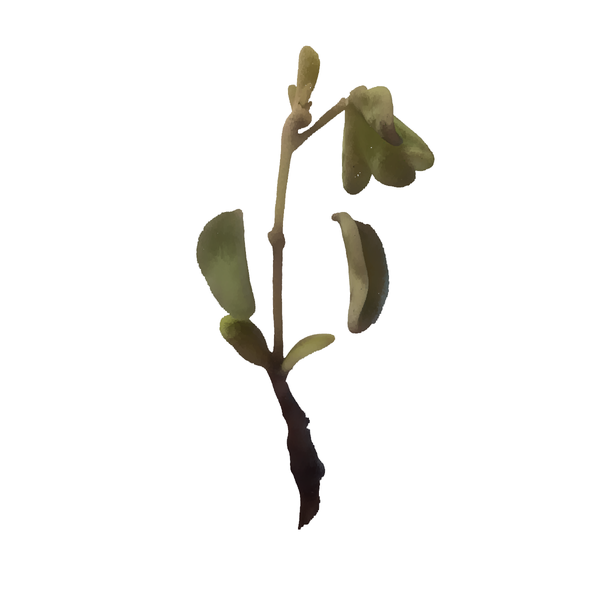}
&\includegraphics[width=0.25\linewidth]{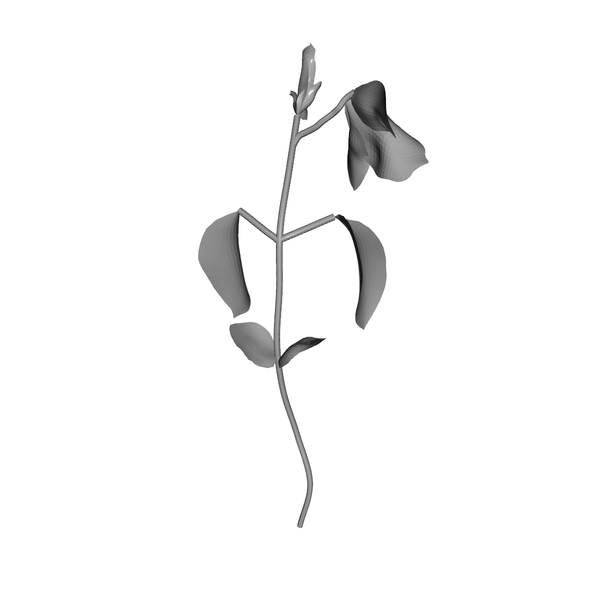}
&\includegraphics[width=0.25\linewidth]{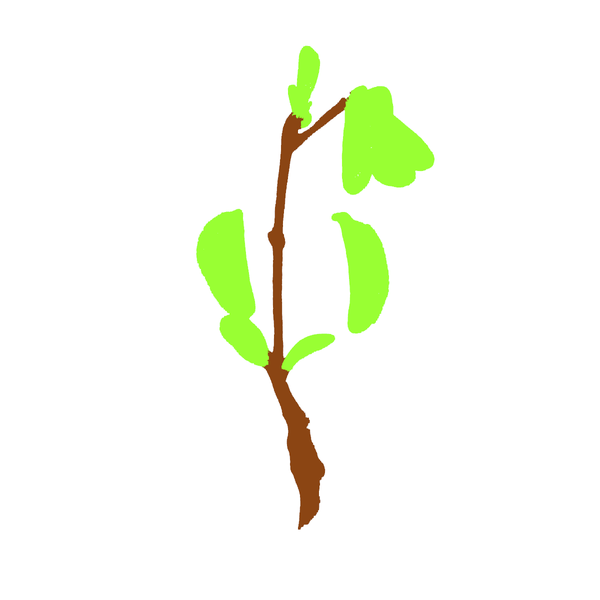}
&\includegraphics[width=0.25\linewidth]{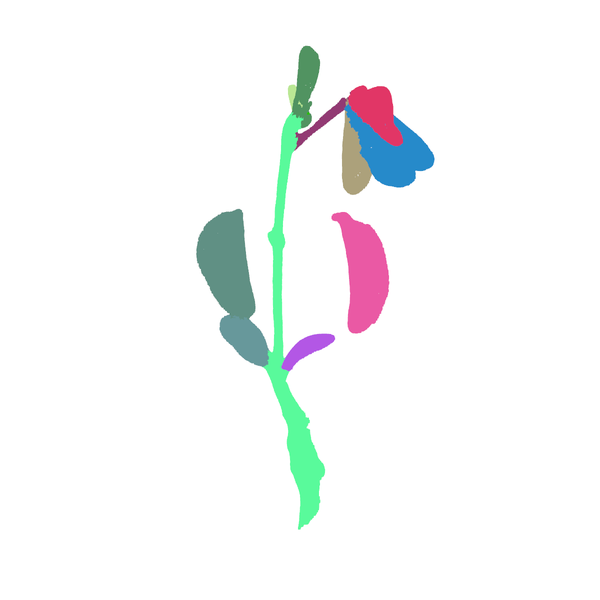}
&\includegraphics[width=0.25\linewidth]{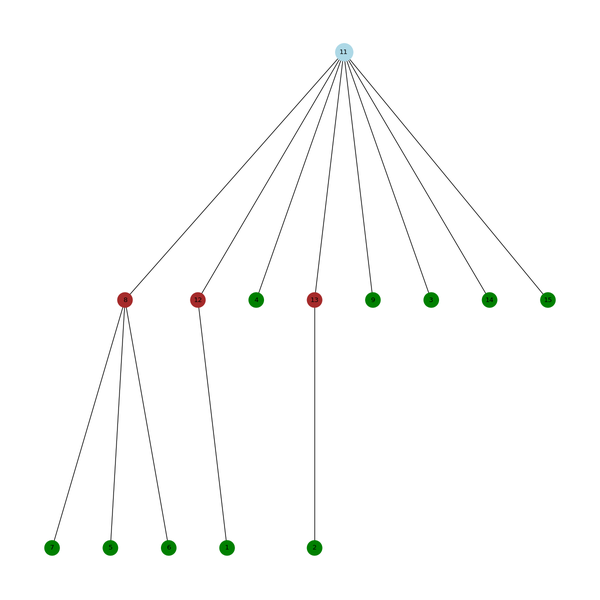}

\\
\raisebox{7mm}{\rotatebox{90}{{\small Prediction}}}
&\includegraphics[width=0.25\linewidth]{figure/empty.png}
&\includegraphics[width=0.25\linewidth]{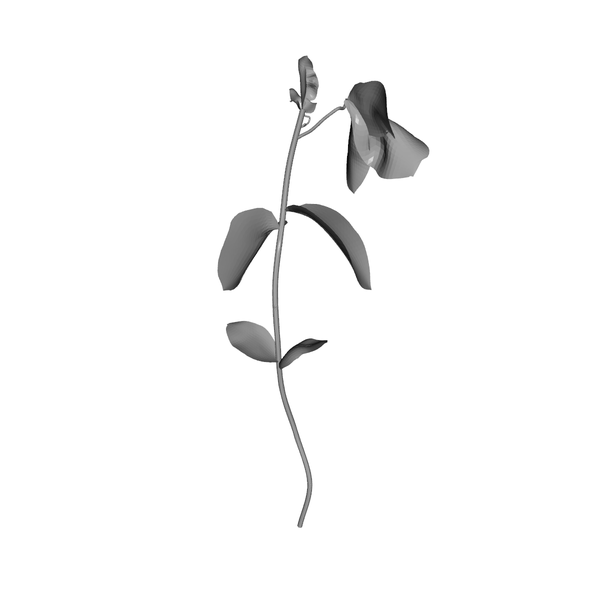}
&\includegraphics[width=0.25\linewidth]{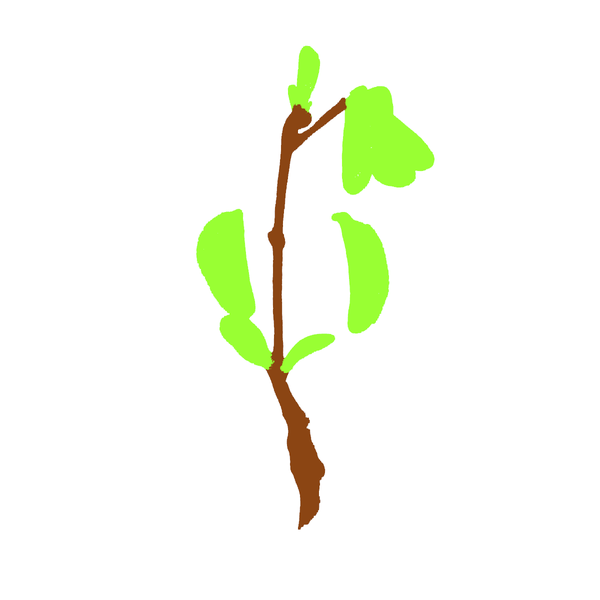}
&\includegraphics[width=0.25\linewidth]{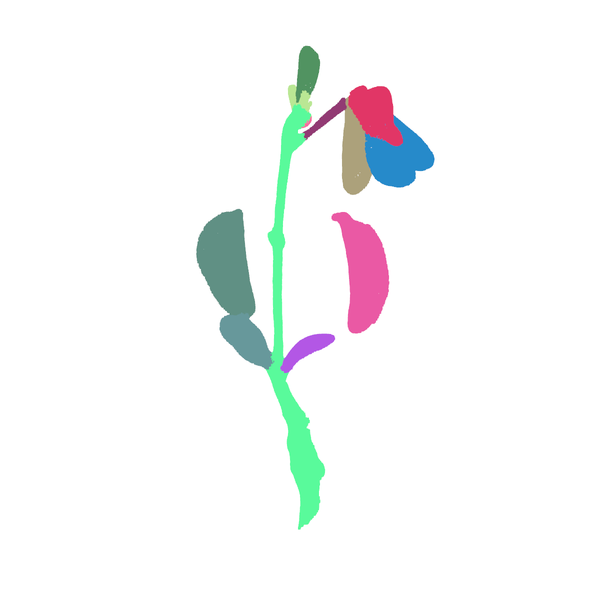}
&\includegraphics[width=0.25\linewidth]{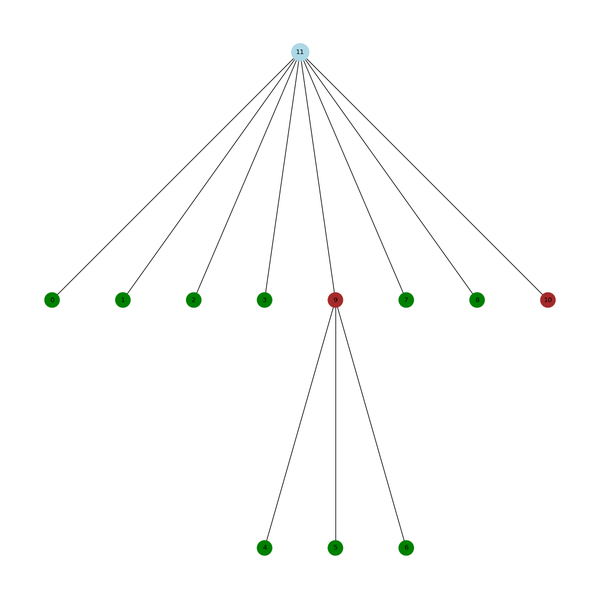}

\end{tabular}

}
\caption{\textbf{Qualitative result of point-based reconstruction.} Given the unlabeled 3D point cloud as input, our model could faithfully recover the mesh, semantics, instances and topology. 
}
\label{fig:3d_recon_detailed}
\end{figure*}
\begin{figure*}[t]
\centering
\def\arraystretch{0.3}
\setlength{\tabcolsep}{0.5pt}
\resizebox{\linewidth}{!}{
\begin{tabular}{ccccccccc}
&{\small Input} 
&{\small Mesh} 
&{\small Semantics}
&{\small Instances}
&{\small Topology}

\\
\raisebox{7mm}{\rotatebox{90}{{\small Ground truth}}}
&\includegraphics[width=0.25\linewidth]{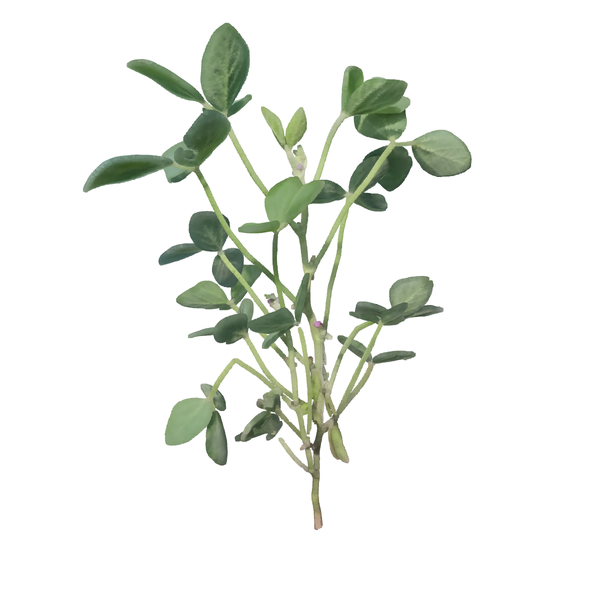}
&\includegraphics[width=0.25\linewidth]{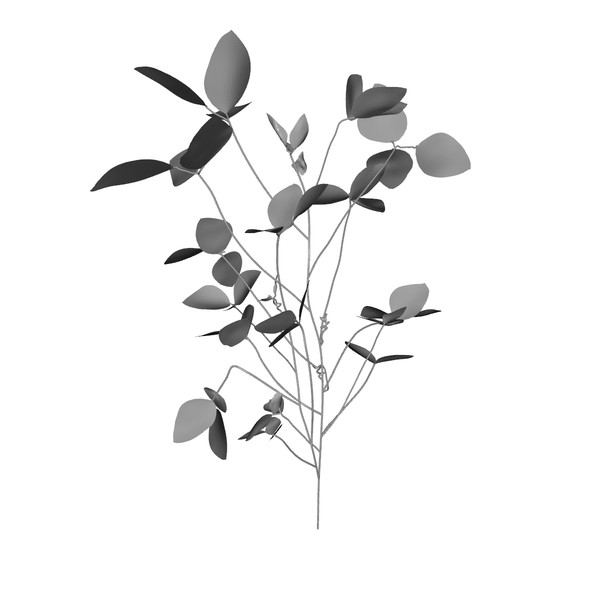}
&\includegraphics[width=0.25\linewidth]{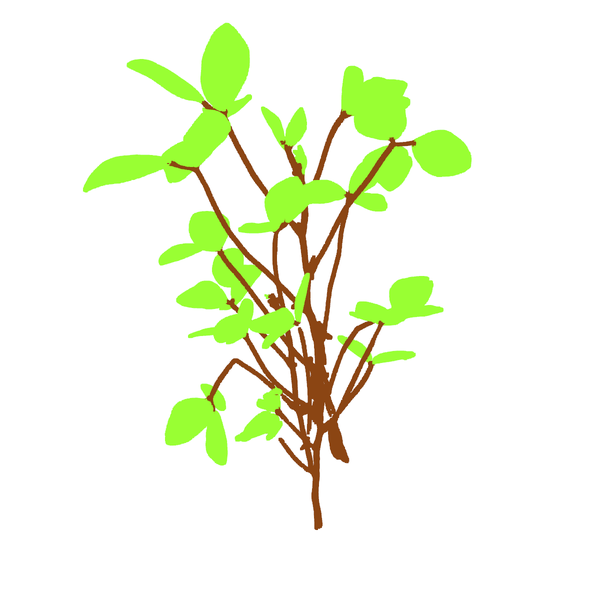}
&\includegraphics[width=0.25\linewidth]{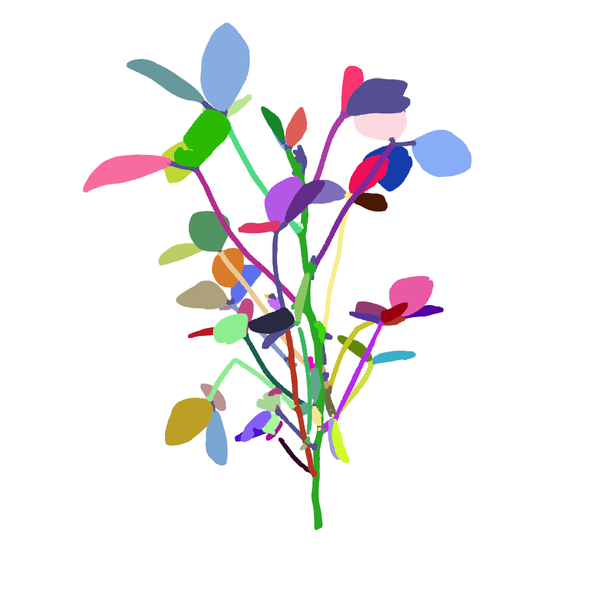}
&\includegraphics[width=0.25\linewidth]{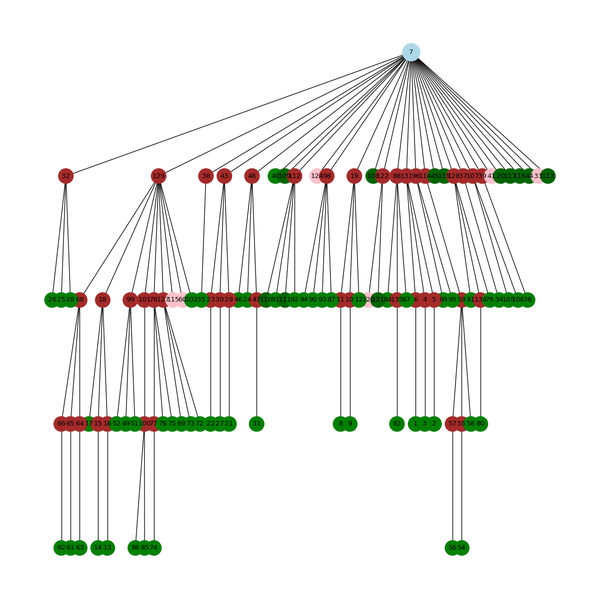}
\vspace{-6pt}

\\
\raisebox{7mm}{\rotatebox{90}{{\small Prediction}}}
&\includegraphics[width=0.25\linewidth]{figure/empty.png}
&\includegraphics[width=0.25\linewidth]{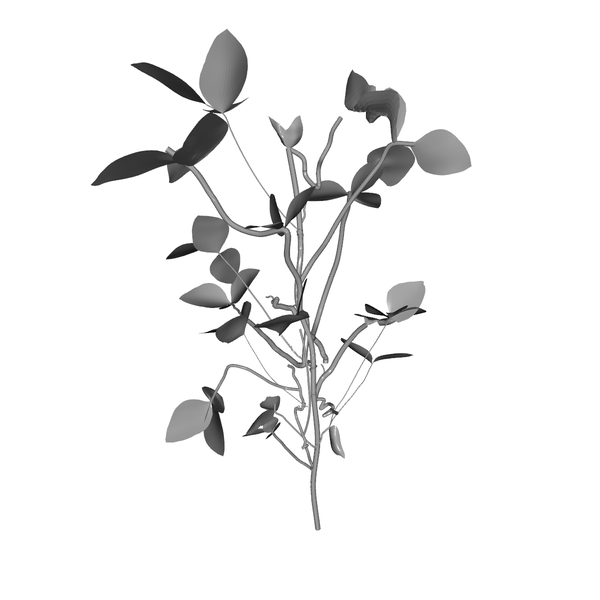}
&\includegraphics[width=0.25\linewidth]{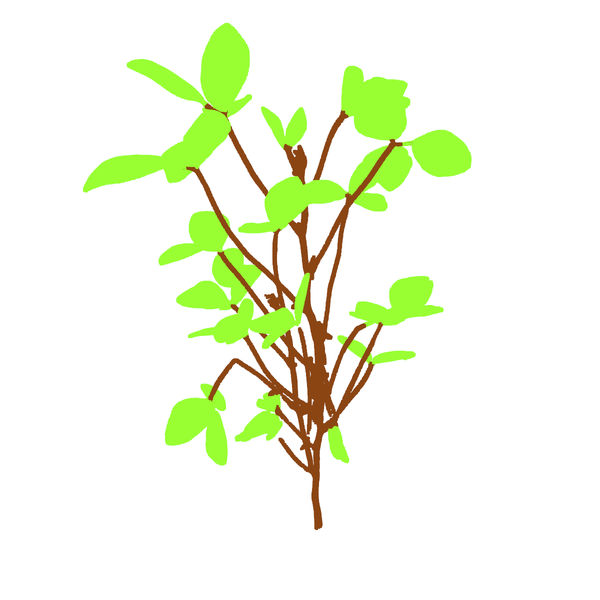}
&\includegraphics[width=0.25\linewidth]{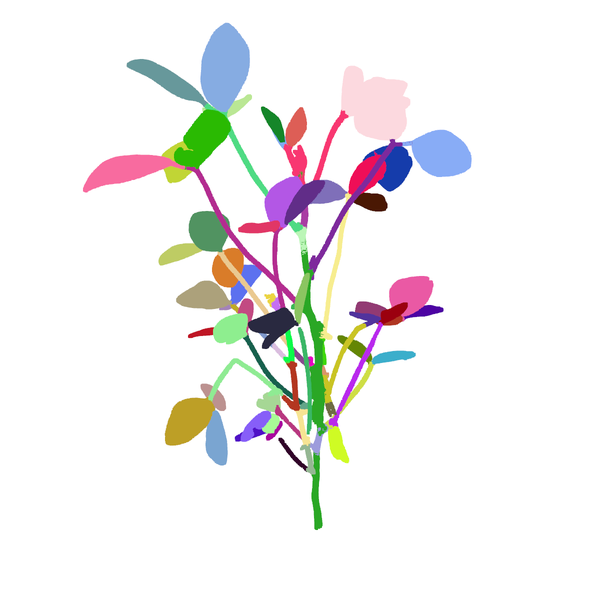}
&\includegraphics[width=0.25\linewidth]{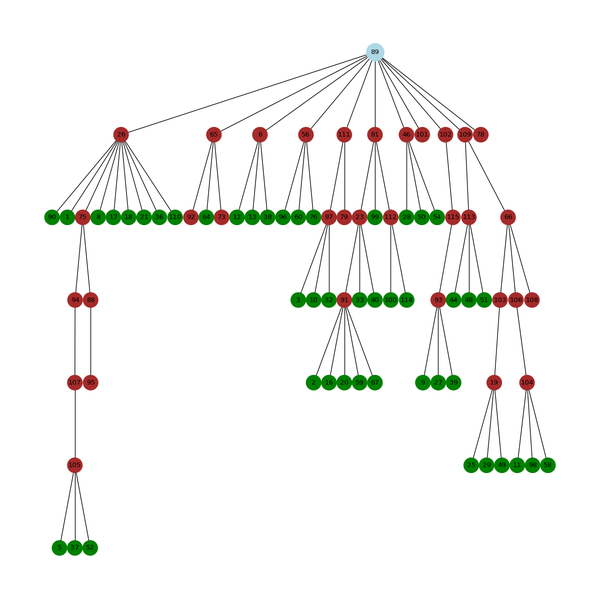}
\vspace{-12pt}

\\
\raisebox{7mm}{\rotatebox{90}{{\small Ground truth}}}
&\includegraphics[width=0.25\linewidth]{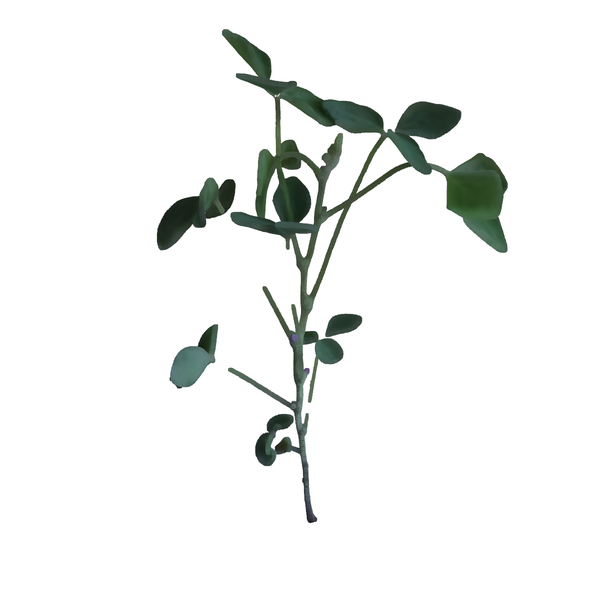}
&\includegraphics[width=0.25\linewidth]{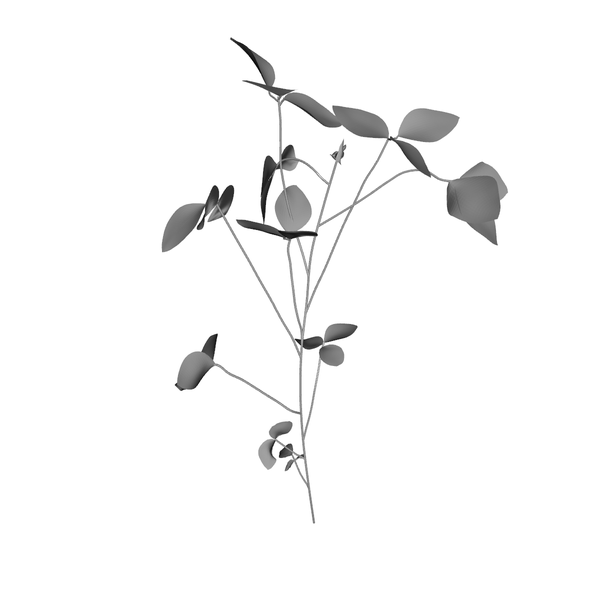}
&\includegraphics[width=0.25\linewidth]{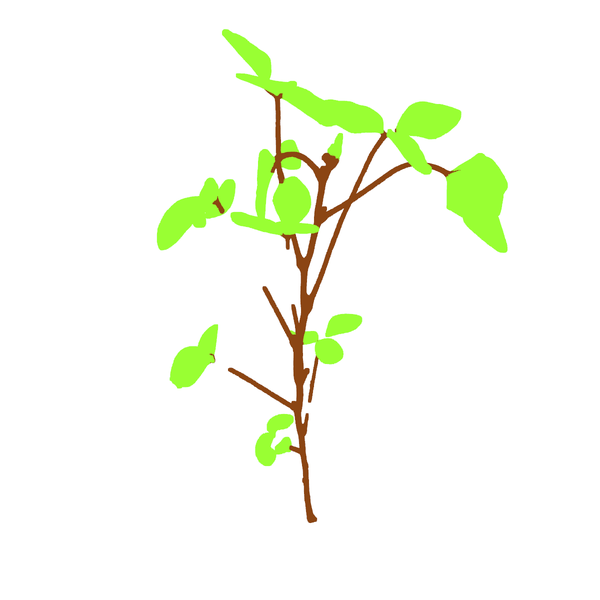}
&\includegraphics[width=0.25\linewidth]{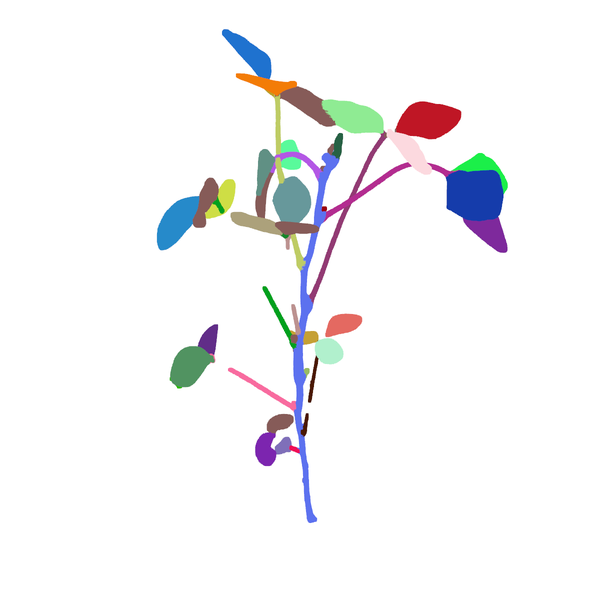}
&\includegraphics[width=0.25\linewidth]{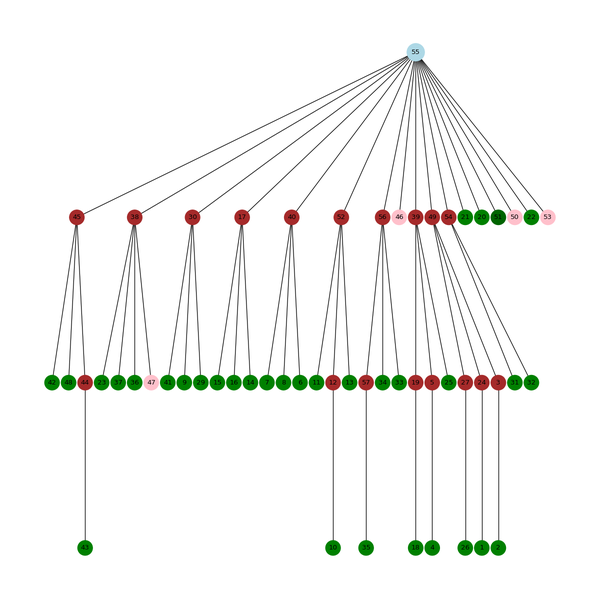}
\vspace{-6pt}

\\
\raisebox{7mm}{\rotatebox{90}{{\small Prediction}}}
&\includegraphics[width=0.25\linewidth]{figure/empty.png}
&\includegraphics[width=0.25\linewidth]{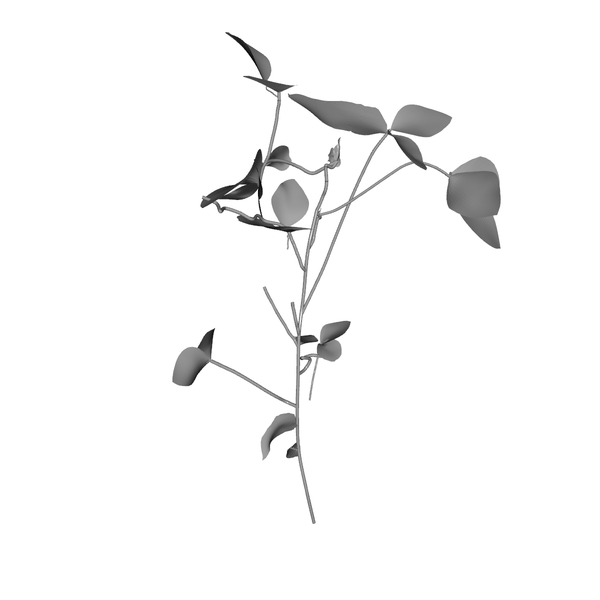}
&\includegraphics[width=0.25\linewidth]{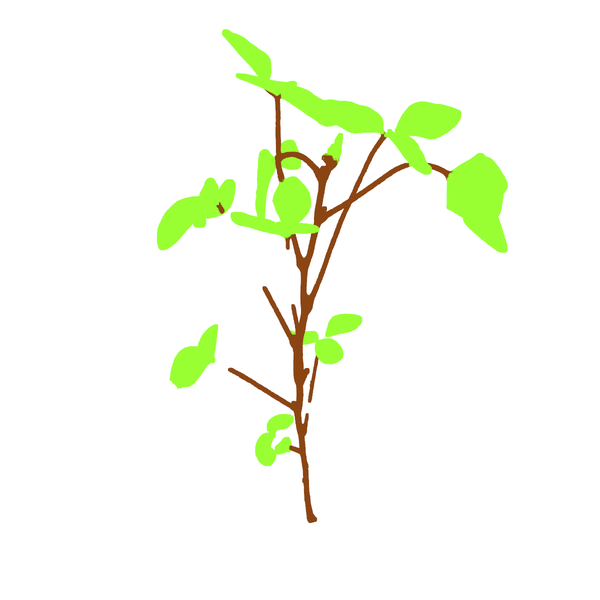}
&\includegraphics[width=0.25\linewidth]{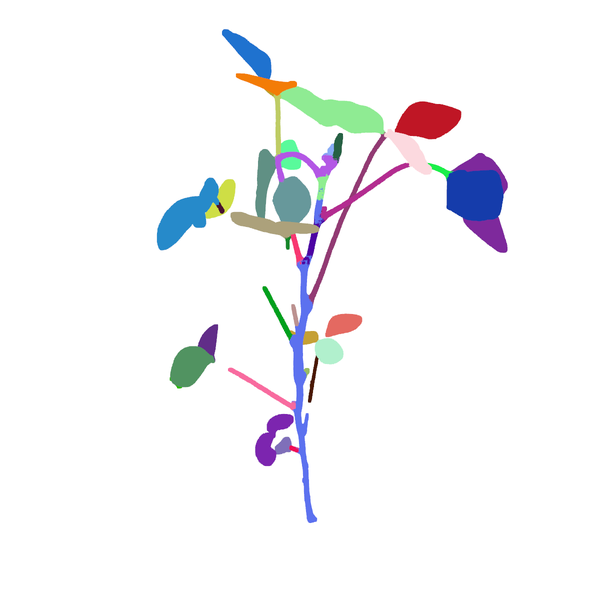}
&\includegraphics[width=0.25\linewidth]{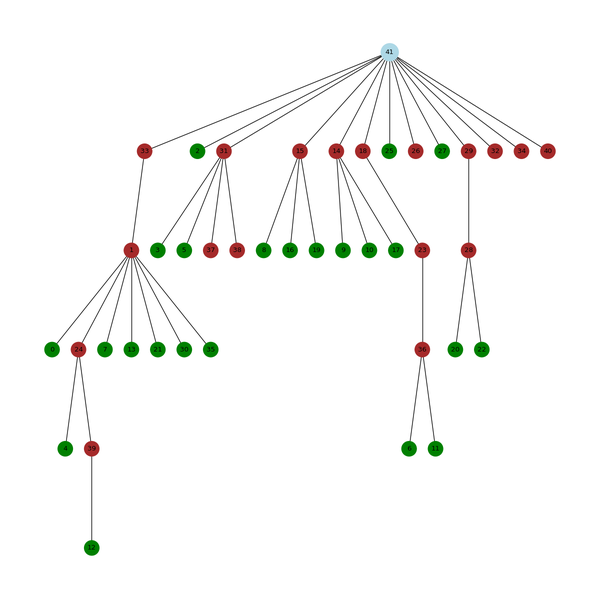}
\vspace{-12pt}

\\
\raisebox{7mm}{\rotatebox{90}{{\small Ground truth}}}
&\includegraphics[width=0.25\linewidth]{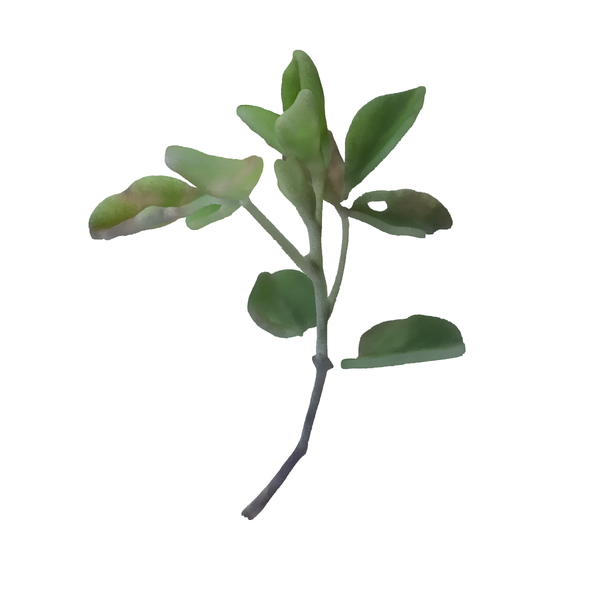}
&\includegraphics[width=0.25\linewidth]{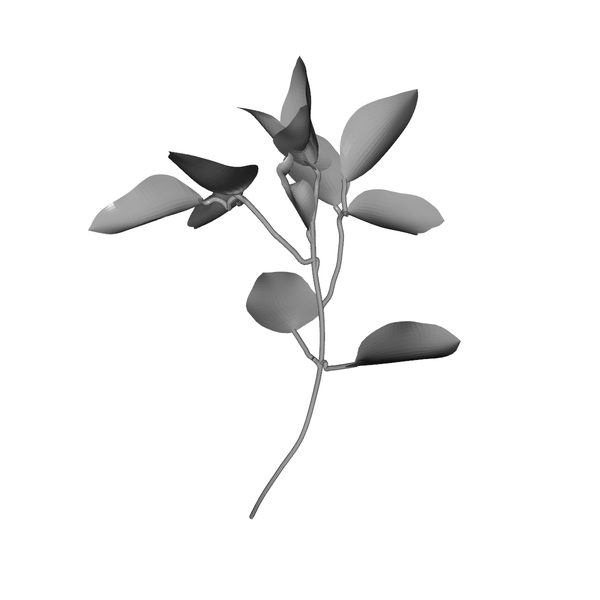}
&\includegraphics[width=0.25\linewidth]{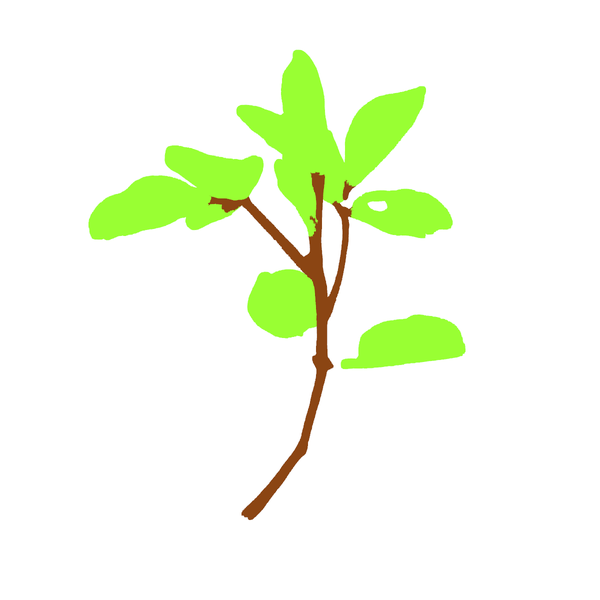}
&\includegraphics[width=0.25\linewidth]{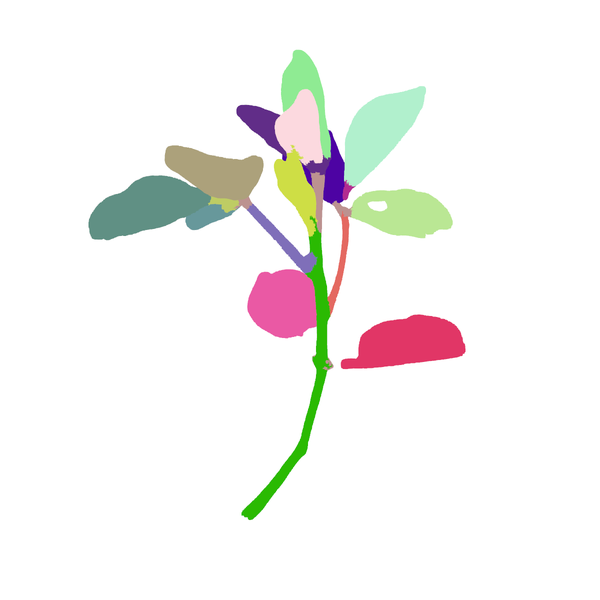}
&\includegraphics[width=0.25\linewidth]{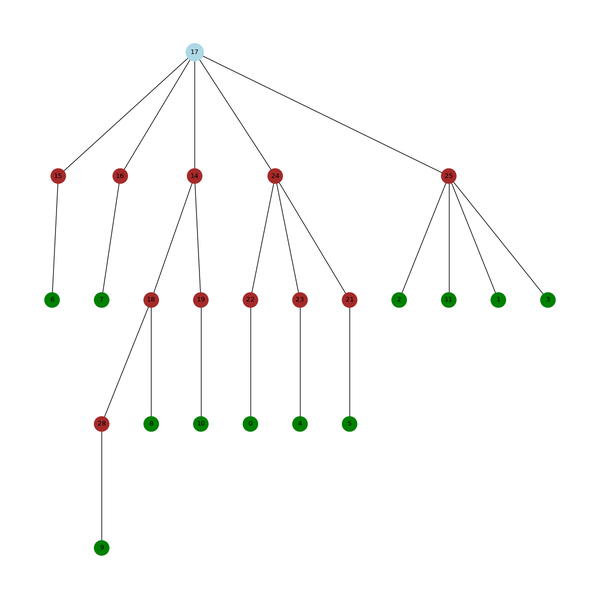}
\vspace{-6pt}

\\
\raisebox{7mm}{\rotatebox{90}{{\small Prediction}}}
&\includegraphics[width=0.25\linewidth]{figure/empty.png}
&\includegraphics[width=0.25\linewidth]{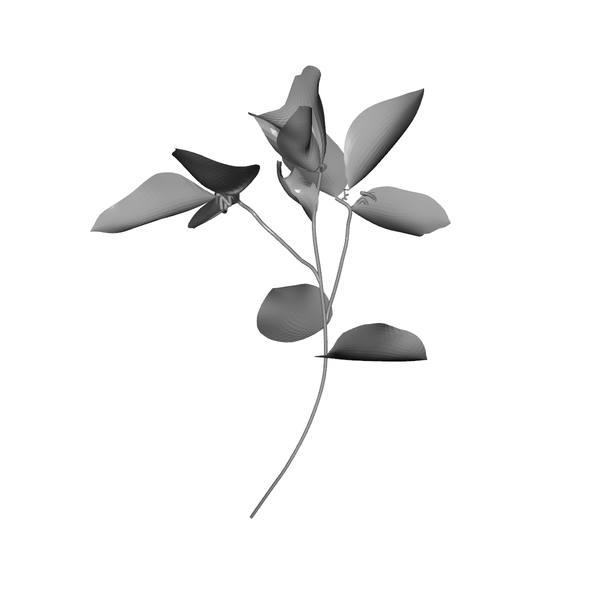}
&\includegraphics[width=0.25\linewidth]{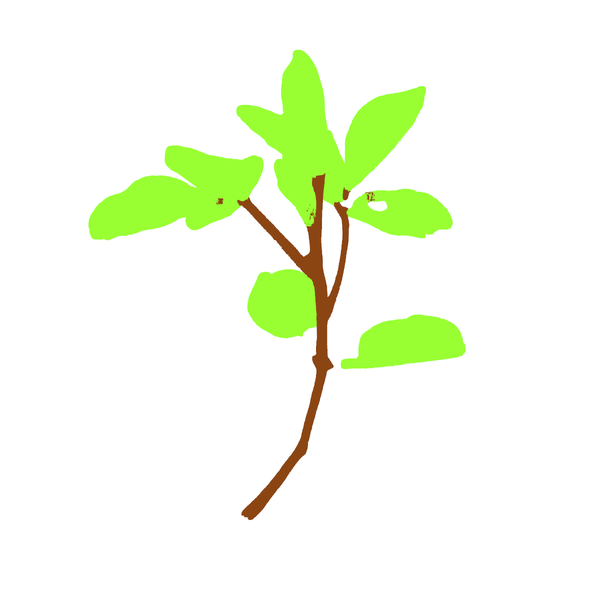}
&\includegraphics[width=0.25\linewidth]{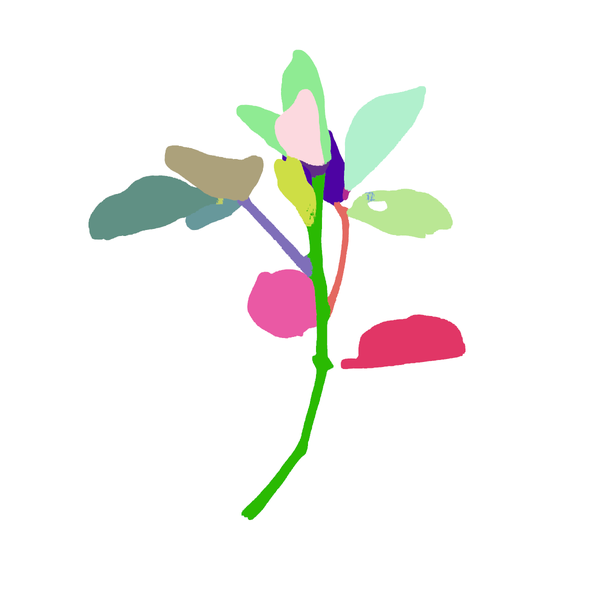}
&\includegraphics[width=0.25\linewidth]{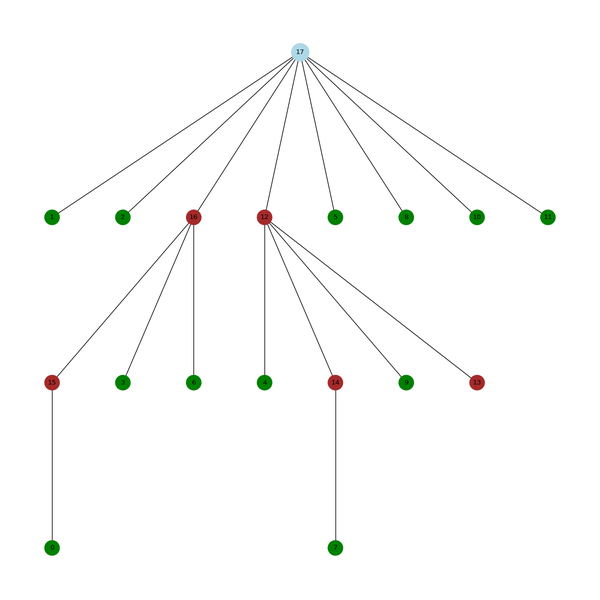}
\vspace{-12pt}

\end{tabular}

}
\vspace{-2mm}
\caption{\textbf{Qualitative result of point-based reconstruction.} Given the unlabeled 3D point cloud as input, our model could faithfully recover the mesh, semantics, instances and topology. 
}
\vspace{-12pt}
\label{fig:3d_recon_detailed2}
\end{figure*}

\begin{table*}[h!]
\centering
\scalebox{1}{

\begin{tabular}{ccccccc|ccc}
\hline
       & Ribes                                                & Rose                                                 & Pepper                                               & Tobacco                                              & Soybean                                              & Maize                                                & Smooth       & Learnable    & Disentangle  \\ \hline
NKSR   & \cellcolor[HTML]{FD9B9A}{\color[HTML]{333333} 0.084} & 0.095                                                & 0.087                                                & \cellcolor[HTML]{FD9B9A}{\color[HTML]{333333} 0.081} & 0.174                                                & \cellcolor[HTML]{FD9B9A}{\color[HTML]{333333} 0.332} &              & $\checkmark$ &              \\
Bezier & 0.134                                                & \cellcolor[HTML]{FECD9E}{\color[HTML]{333333} 0.066} & \cellcolor[HTML]{FECD9E}{\color[HTML]{333333} 0.082} & 0.252                                                & \cellcolor[HTML]{FECD9E}{\color[HTML]{333333} 0.159} & 0.978                                                & $\checkmark$ &              &              \\
Ours   & \cellcolor[HTML]{FECD9E}{\color[HTML]{333333} 0.121} & \cellcolor[HTML]{FD9B9A}{\color[HTML]{333333} 0.041} & \cellcolor[HTML]{FD9B9A}{\color[HTML]{333333} 0.054} & \cellcolor[HTML]{FECD9E}{\color[HTML]{333333} 0.095} & \cellcolor[HTML]{FD9B9A}{\color[HTML]{333333} 0.154} & \cellcolor[HTML]{FECD9E}{\color[HTML]{333333} 0.649} & $\checkmark$ & $\checkmark$ & $\checkmark$ \\ \hline
\end{tabular}

}
\caption{\textbf{Leaf fitting error for different species.} We show normalized CD($\times$ 100). We highlight the \colorbox{best_color}{best} and \colorbox{second_color}{second best} values.}
\label{Tab:fitting_compare_big}
\end{table*}

\paragraph{Overfitting over different species}

To further show the generalization ability, we report the overfitting result on different species using our model, NKSR and Bézier surfaces (most common NURBS shape) and report results in Tab.~\ref{Tab:fitting_compare_big}. We didn't compare with CropCraft\cite{zhai2024cropcraft} since it only works for Soybean. Our model yields superior reconstruction plus two key benefits: (1) a PCA‐based, learnable shape prior; and (2) disentangled, interpretable parameters—unlike Bézier curves, whose control points lack physical meaning and are not even on-surface, our on‐leaf and on‐vein control points and PCA basis correspond to biophysical/phenotypical traits.

\paragraph{3D Reconstruction}

Given the ground-truth instance segmentation, we calculate the minimal distance from each point to other instances, as shown in Fig. \ref{fig:ptv3_inv_dist}.

For training, we follow the default configuration in PointTransformer-V3 but without the mixing strategy. We set the batch size as 7 and use AdamW optimizer with a learning rate 0.0025, weight decay 0.02, and trained in 300 epochs. The loss function is composed of the cross-entropy loss and distance regression loss, i.e. $l_{total}=l_{ce}+15l_{dist}$.


The predicted graph differs from the ground truth due to both errors in prediction and the inherent ambiguity in stem connectivity labeling. Additionally, our model cannot automatically fill in missing stems if the input point cloud has too many missing parts, as shown in Fig.~\ref{fig:3d_recon_detailed}.

\paragraph{2D Reconstruction}

\begin{figure*}[t]
\centering
\def\arraystretch{0.3}
\setlength{\tabcolsep}{0.5pt}
\resizebox{\linewidth}{!}{
\begin{tabular}{ccccccccc}
&{\small Input} 
&{\small Input View} 
&{\small Instances}
&{\small Novel View}

\\
\raisebox{4mm}{\rotatebox{90}{{\small Ground truth}}}
&\includegraphics[width=0.25\linewidth]{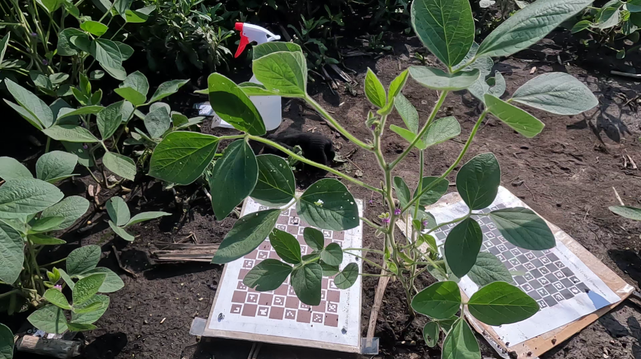}
&\includegraphics[width=0.25\linewidth]{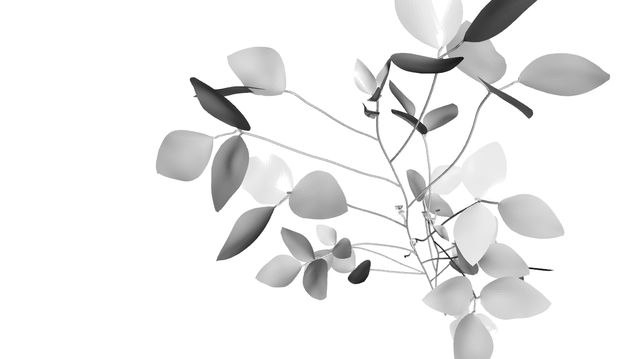}
&\includegraphics[width=0.25\linewidth]{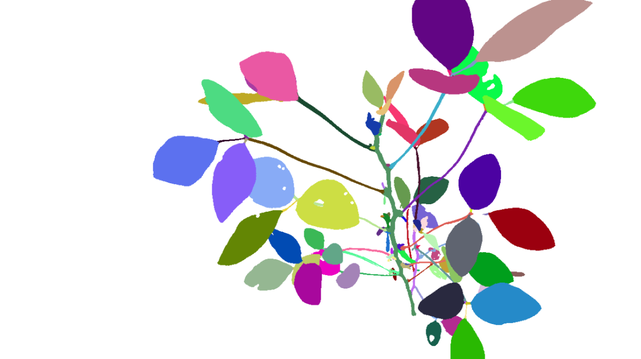}
&\includegraphics[width=0.25\linewidth]{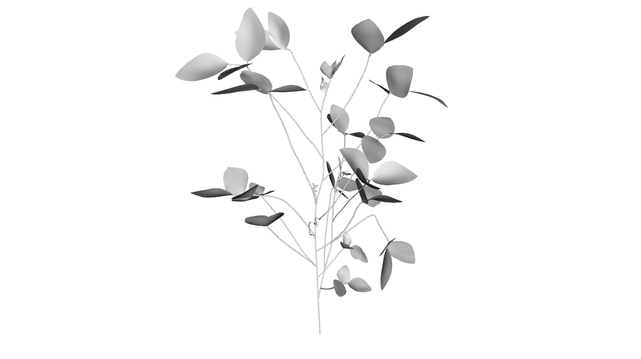}

\\
\raisebox{7mm}{\rotatebox{90}{{\small Prediction}}}
&\includegraphics[width=0.25\linewidth]{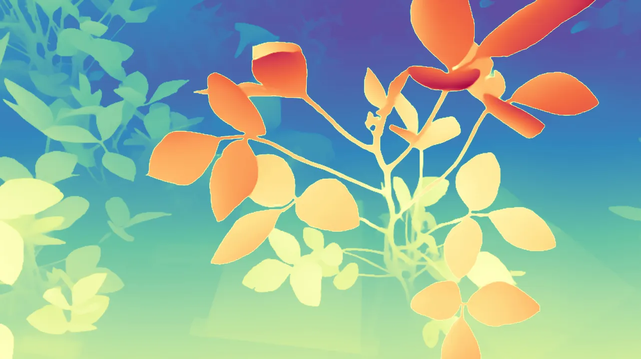}
&\includegraphics[width=0.25\linewidth]{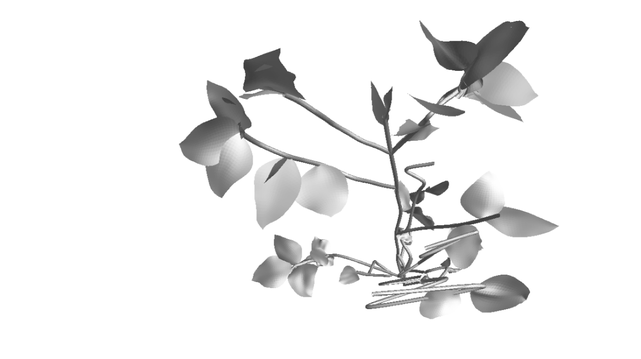}
&\includegraphics[width=0.25\linewidth]{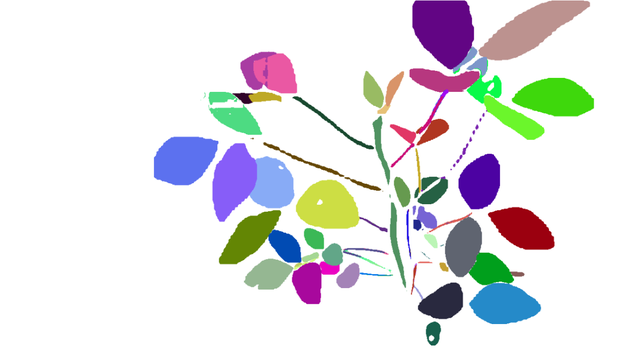}
&\includegraphics[width=0.25\linewidth]{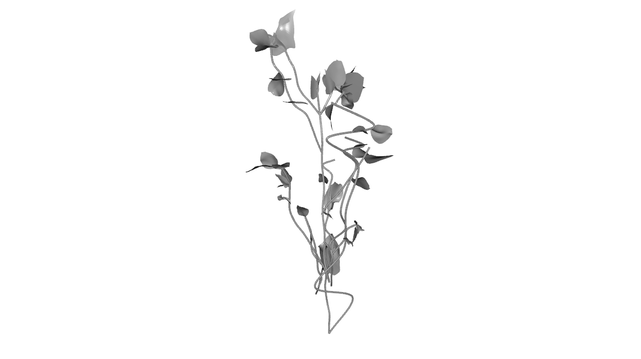}

\\
\raisebox{4mm}{\rotatebox{90}{{\small Ground truth}}}
&\includegraphics[width=0.25\linewidth]{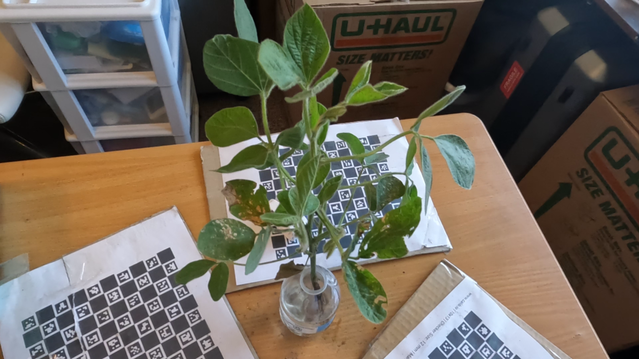}
&\includegraphics[width=0.25\linewidth]{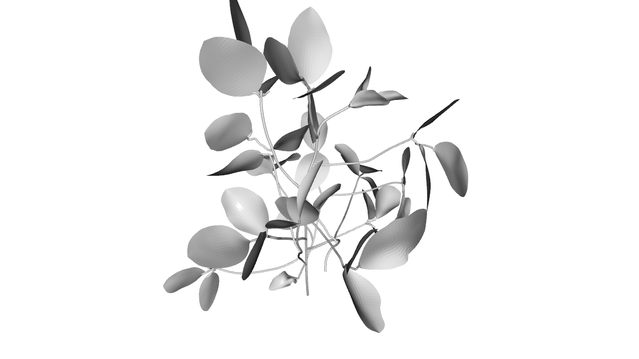}
&\includegraphics[width=0.25\linewidth]{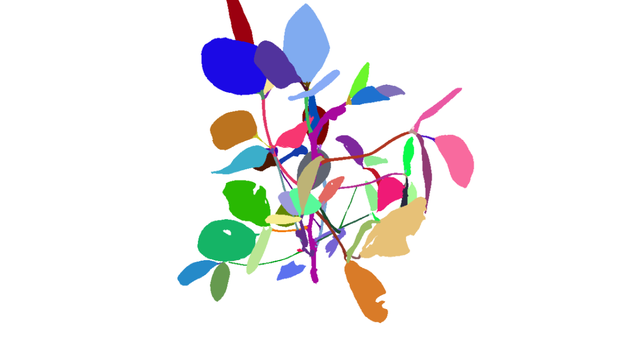}
&\includegraphics[width=0.25\linewidth]{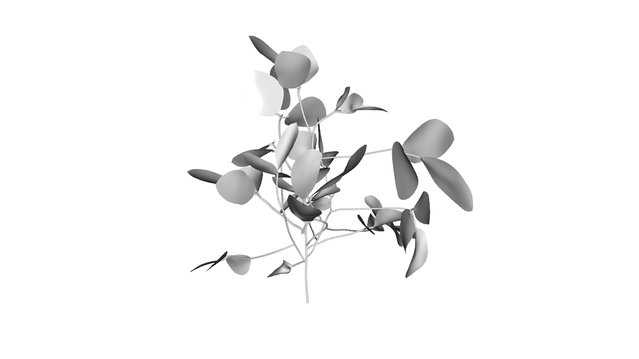}

\\
\raisebox{7mm}{\rotatebox{90}{{\small Prediction}}}
&\includegraphics[width=0.25\linewidth]{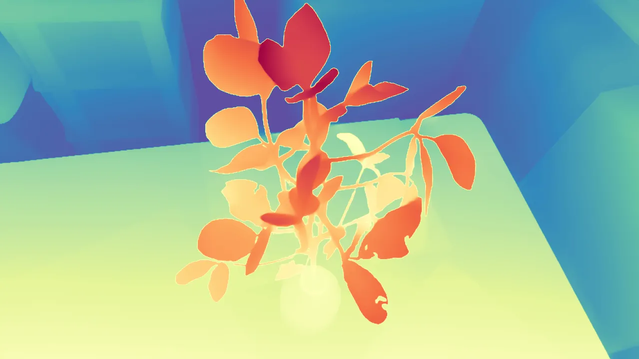}
&\includegraphics[width=0.25\linewidth]{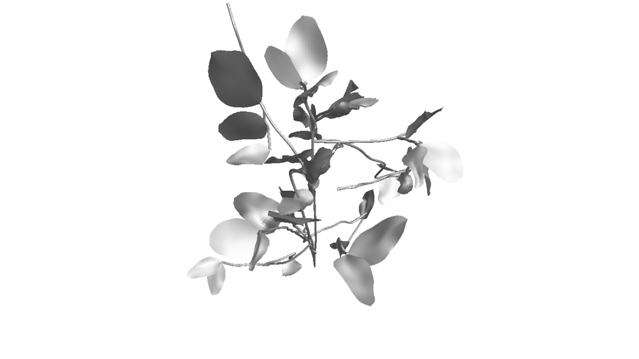}
&\includegraphics[width=0.25\linewidth]{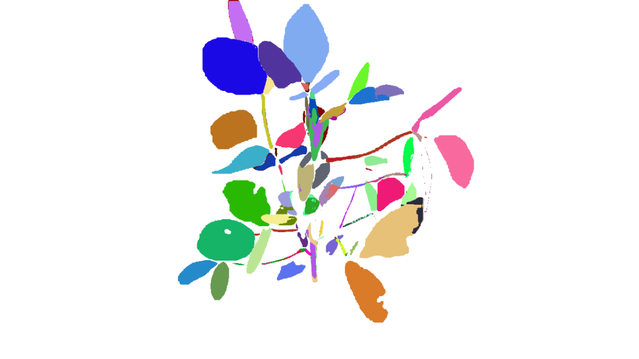}
&\includegraphics[width=0.25\linewidth]{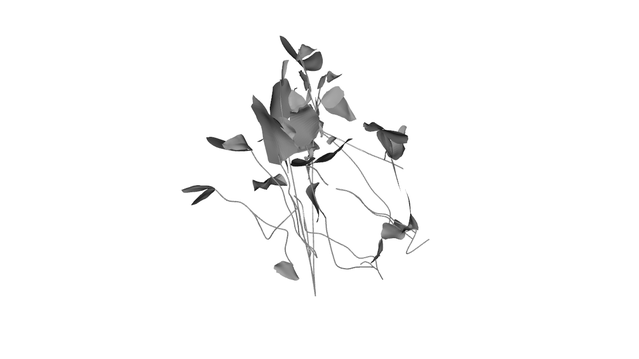}

\\
\raisebox{4mm}{\rotatebox{90}{{\small Ground truth}}}
&\includegraphics[width=0.25\linewidth]{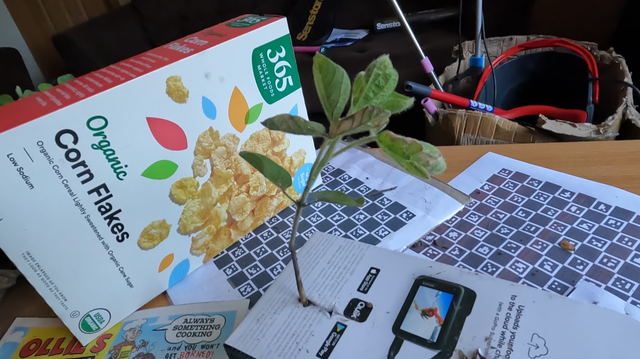}
&\includegraphics[width=0.25\linewidth]{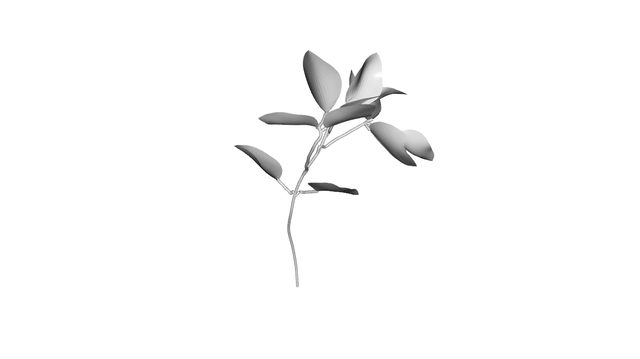}
&\includegraphics[width=0.25\linewidth]{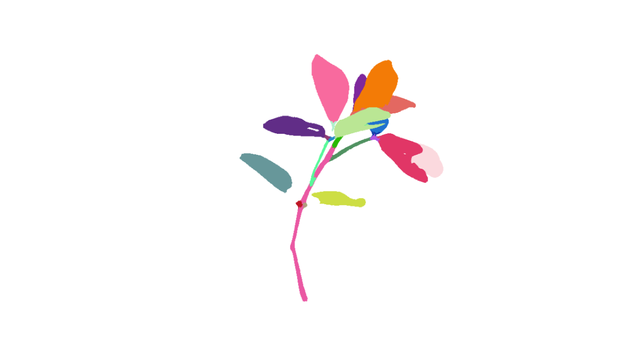}
&\includegraphics[width=0.25\linewidth]{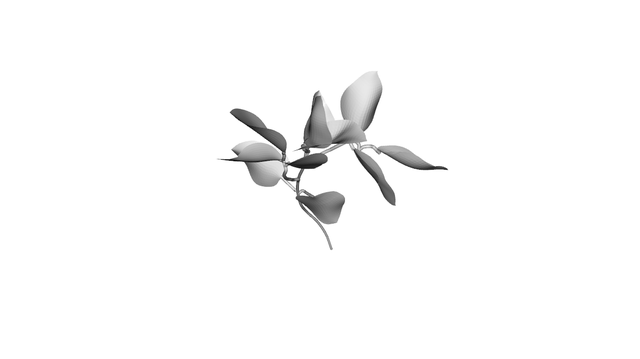}

\\
\raisebox{7mm}{\rotatebox{90}{{\small Prediction}}}
&\includegraphics[width=0.25\linewidth]{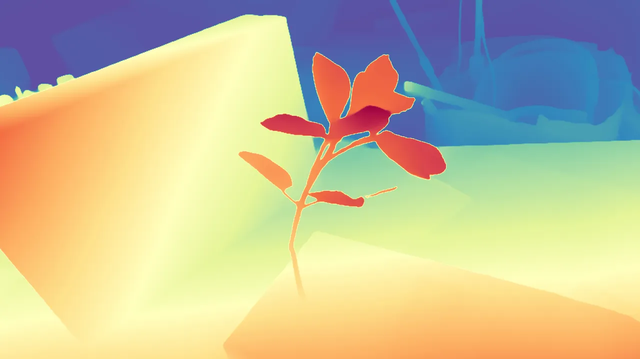}
&\includegraphics[width=0.25\linewidth]{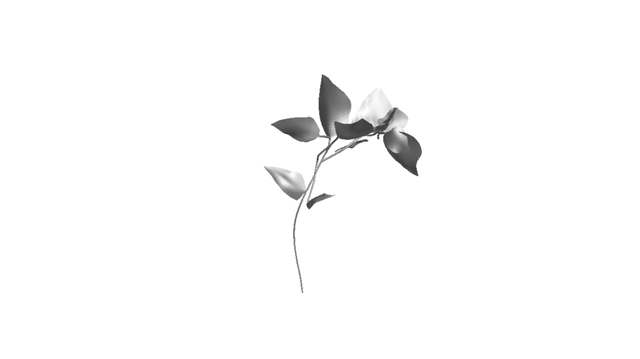}
&\includegraphics[width=0.25\linewidth]{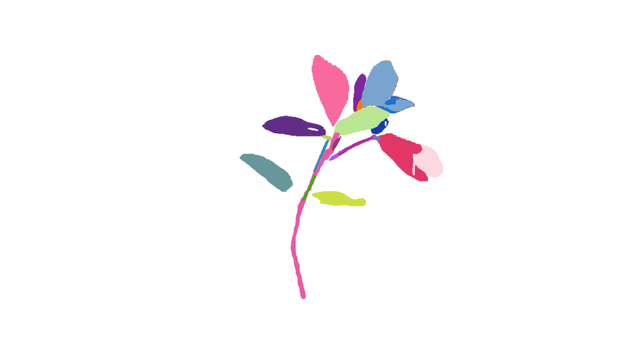}
&\includegraphics[width=0.25\linewidth]{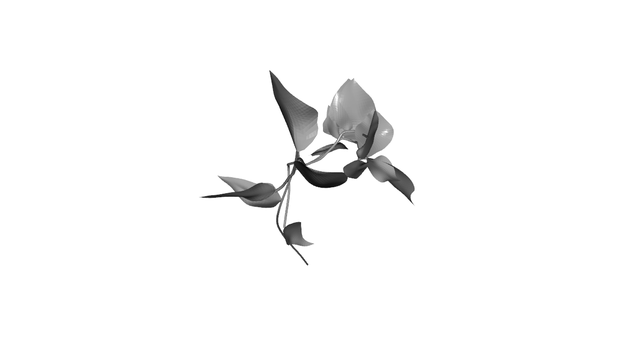}

\\
\raisebox{7mm}{\rotatebox{90}{{\small Ground truth}}}
&\includegraphics[width=0.25\linewidth]{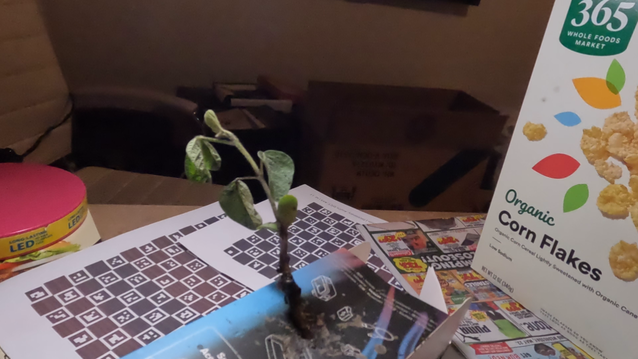}
&\includegraphics[width=0.25\linewidth]{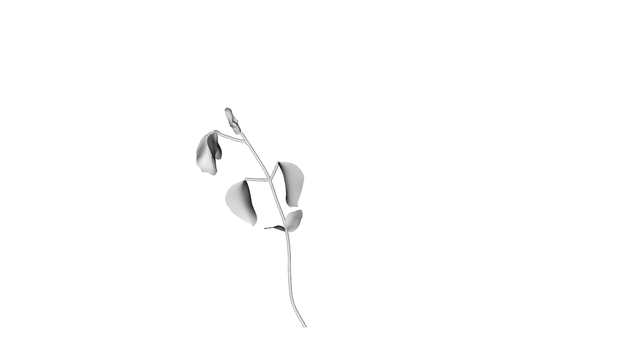}
&\includegraphics[width=0.25\linewidth]{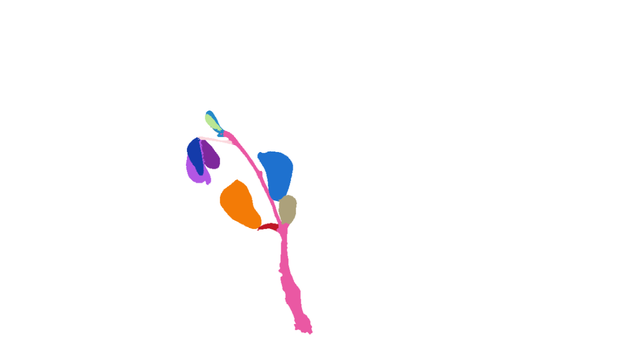}
&\includegraphics[width=0.25\linewidth]{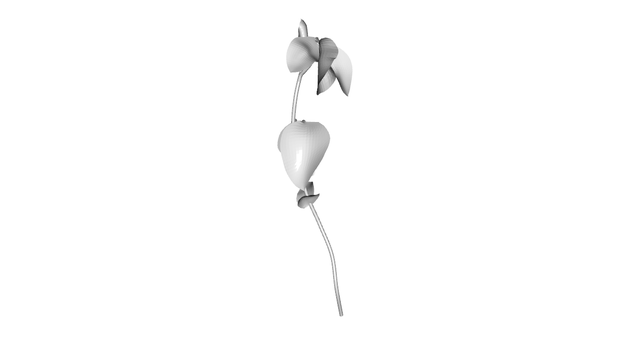}

\\
\raisebox{7mm}{\rotatebox{90}{{\small Prediction}}}
&\includegraphics[width=0.25\linewidth]{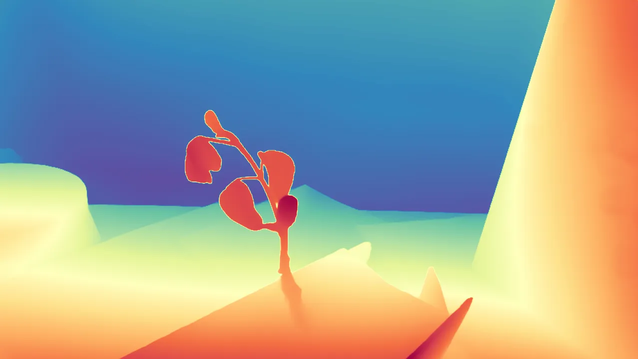}
&\includegraphics[width=0.25\linewidth]{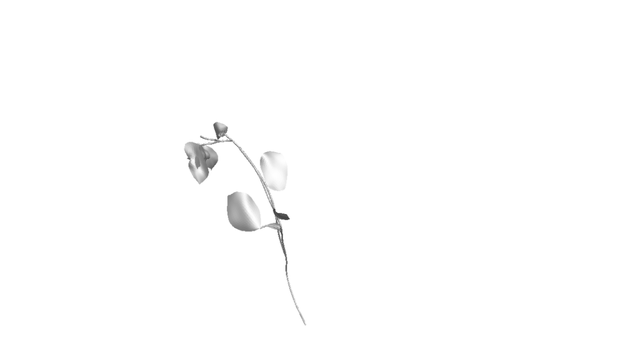}
&\includegraphics[width=0.25\linewidth]{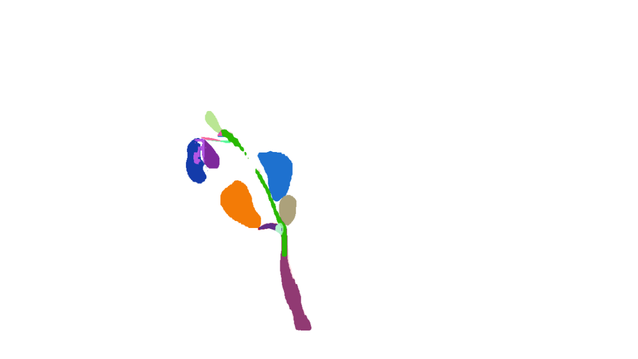}
&\includegraphics[width=0.25\linewidth]{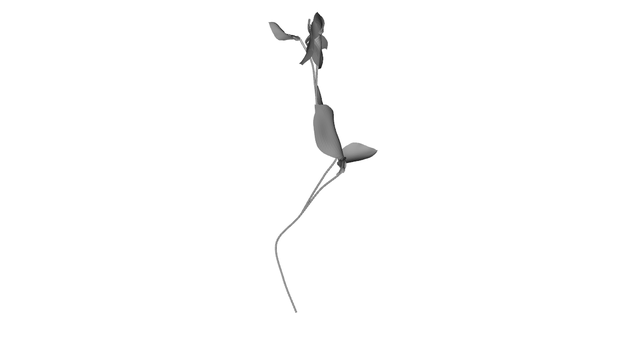}

\end{tabular}

}
\caption{\textbf{Qualitative result of image-based reconstruction.} The depth prediction comes from off-the-shelf DepthAnything~\cite{yang2024depth}
}
\label{fig:2d_recon_detailed}
\end{figure*}

We train Mask-RCNN with a ResNet50-FPN backbone starting from the COCO-pretrained model in detectron2~\cite{wu2019detectron2} for 20K iterations. We use the default configuration but with a batch size of 4 and lowering the learning rate at 8K and 12K iterations. For inference, we use a confidence threshold of 0.8.

We apply an off-the-shelf depth estimator~\cite{yang2024depth} to lift all the instances to 3D. However, the lifted partial 3D point clouds are usually noisy. Therefore we remove instances if the total number of points is less than 30 for stems and 100 for leaves, and remove leaves based on the proportions of their rotated bounding boxes. We also filter out the points with a high gradient of depth. We infer the topology from the partial point cloud by building a minimal spanning tree, and removing the "bare stems" (stems without any children) to reduce noise. Afterwards, we fit the Demeter model parameters in the same way as before. During fitting, we apply the linear model to constrain the shape parameter $\boldsymbol \beta$ into $(-2\sigma,2\sigma)$ and adapt the L1 chamfer distance for robustness to outliers. As a result, our model achieves a 2D IoU of 0.9028 between the mask of predicted mesh and the mask of ground-truth mesh.

Our model may produce noisy outputs in novel views because we only apply constraints at the node level, not at the global topology level. We also did not constrain the leaf deformation or stem deformation. We leave these as directions for future improvements.

\paragraph{Remark} Both the multi-view and single-image reconstruction we proposed here is a preliminary exploration of 3D reconstruction using Demeter. Despite showing great potential, we believe there is significant potential for the community to develop better reconstruction algorithms in the future using our parametric model.

\section{Detail about Maize Species}

We show that the Demeter methodology generalizes to other plant species by developing a {\bf Demeter-Maize} prototype using the Pheno4D~\cite{schunck2021pheno4d} dataset, which consists of 84 maize point clouds capturing different growth stages of 6 maize plants. We also reconstruct maize from a web dataset~\cite{Zhang2022}, which contains larger maize plants compared to Pheno4D. To acquire the 2D shape parameterization, we simply append an additional scaling axis to the soybean leaf shape parameters (Fig.~\ref{fig:maize_leaf_pca_viz}), since there is no existing maize leaf scan dataset. Afterward, we apply the same pipeline as for soybeans to learn the 3D deformation basis and fit other parameters. The result (Fig.~\ref{fig:3d_recon_maize}) shows that Demeter can faithfully capture maize plant shape.

\section{Other Application}

\paragraph{Agriculture}

\begin{table}[]
    \centering
    \resizebox{\linewidth}{!}{
\setlength{\tabcolsep}{0.1em} 
\renewcommand{\arraystretch}{1.}
    \begin{tabular}{cccc}
    \footnotesize Net Photosynthesis & \footnotesize Early Morning (6 AM) & \footnotesize Noon (12 PM)  \\
          
        \includegraphics[width=.3\linewidth, trim={0 0 0 0.5cm}, clip]{figure/photosynthesis/timeseries_102_o.pdf}
        &
        \includegraphics[width=.35\linewidth]{figure/photosynthesis/early_5_102_o.png}
        & 
        \includegraphics[width=.35\linewidth]{figure/photosynthesis/noon_5_102_o.png}

        \\

       \includegraphics[width=.3\linewidth, trim={0 0 0 0.2cm}, clip]{figure/photosynthesis/timeseries_74_o.pdf}
        &
        \includegraphics[width=.35\linewidth]{figure/photosynthesis/early_5_74_o.png}
        & 
        \includegraphics[width=.35\linewidth]{figure/photosynthesis/noon_5_74_o.png}

        \\
    \end{tabular}
    
    }
\captionof{figure}{\textbf{Photosynthesis simulation results}. We perform simulations using Helios~\cite{bailey2019helios} on two soybean canopies generated by repeating Demeter-Soybean models. Left: timeseries of the net photosynthesis rate for the crop canopy over the course of a day, in units of $\mu$molCO$_2$/\SI{}{\meter}$^2$/\SI{}{\second}. Other columns: mesh visualization where each leaf face is colored according to the rate of photosynthesis over that face (brighter = higher rate). }

\label{fig:photosynthesis}
\end{table}
\begin{table}[]
    \centering
    \resizebox{\linewidth}{!}{
\setlength{\tabcolsep}{0.1em} 
\renewcommand{\arraystretch}{1.}
    \begin{tabular}{cccc}
    \footnotesize Net Gas Exchange Rate & \footnotesize Early Morning (6 AM) & \footnotesize Noon (12 PM)  \\
          
        \includegraphics[width=.3\linewidth, trim={0 0 0 0.5cm}, clip]{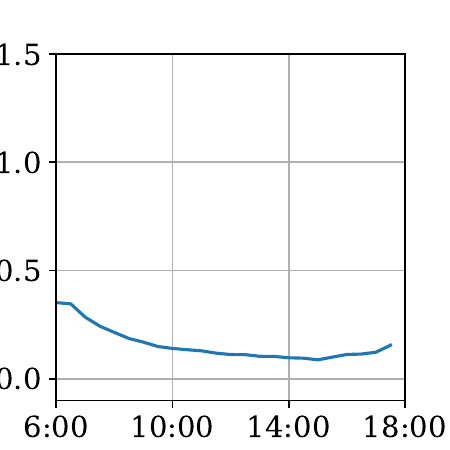}
        &
        \includegraphics[width=.35\linewidth]{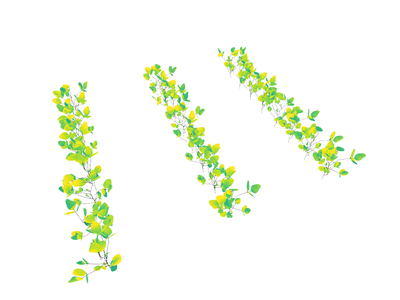}
        & 

        \includegraphics[width=.35\linewidth]{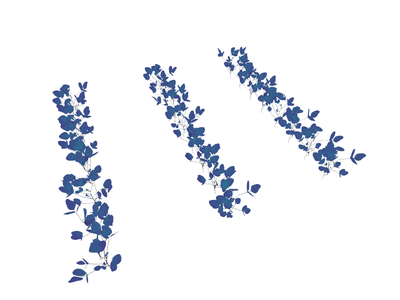}
        \\

       \includegraphics[width=.3\linewidth, trim={0 0 0 0.2cm}, clip]{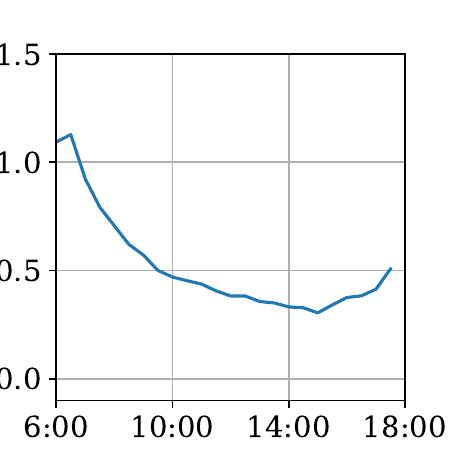}
        &
       \includegraphics[width=.35\linewidth]{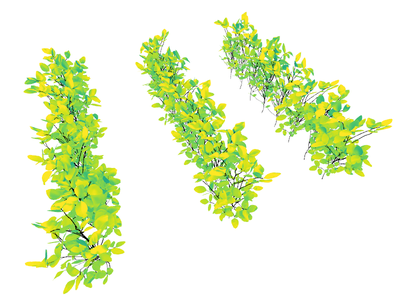}
        & 

        \includegraphics[width=.35\linewidth]{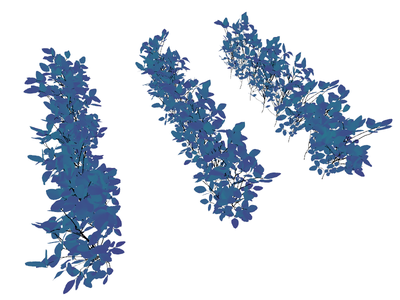}
        \\
    \end{tabular}
    
    }
\captionof{figure}{\textbf{Stomatal conductance simulation results}. Stomatal conductance is a measure of the degree of opening of a plant's stomata, which determines the rate of gas exchange (including carbon dioxide and water) between the plant and the air. Left: timeseries of the net gas exchange per unit ground area, in units of mol/\SI{}{\meter}$^2$/\SI{}{\second}. Other columns: per-face rate-colored visualization.
}
\label{fig:water_transpiration}
\end{table}

We showcase this capability by generating small crop fields by placing fitted Demeter models in a grid, and passing them to Helios~\cite{bailey2019helios} to simulate the response of the plant to weather variations over the course of a day. The weather variables were taken from data measured by a flux tower~\cite{pastorello2020fluxnet2015} and include temperature, humidity, radiation, and other environmental variables. In Fig.~\ref{fig:photosynthesis} and Fig.~\ref{fig:water_transpiration}, we visualize two outputs of the simulation: photosynthesis rate and stomatal conductance, which are both directly related to crop productivity.

\paragraph{Rendering}

By extracting texture from the bijective mapping mentioned in Eq. 6 in the main paper, we could add texture to the parametric model and using ray-tracing to obtain realistic rendering.

\section{Discussion}

\paragraph{The importance of disentanglement}

In agriculture, disentangled and interpretable shape parameters are vital for phenotyping \& genotyping, biophysical simulation, and process‐models, as they serve as controllable, measurable variables in crop scientists’ workflows. In graphics, disentangled shapes allows controllable procedural generation, texture mapping, and physically driven deformations. 

\paragraph{Sensitivity to dataset size}

The learnable parameters are PCA components for leaf and stem shape and deformation. Training these PCA requires only a handful of diverse plants per species—each plant contributes multiple leaves and stems—so long as the total \# of leaves/stems exceeds the \# of PCA control points per template. We can also adapt an initial template from related species (e.g., soybean to cowpea or tobacco) and finetune on even smaller datasets when annotated 3D scans are scarce. For example, our soybean model uses custom scans and FGLIR leaf data; Papaya is trained on PLANesT3D; tobacco adapts the soybean template and retrains on Plant3D, producing a viable model from just 3–5 full 3D samples (with small fidelity trade-offs). 

And training an instance‐segmentation network from scratch typically requires more data (e.g., 50 plants), but fine-tuning a pretrained model on just a few examples generalizes well to new species.

\paragraph{Limitation}

Although our model achieves realistic modeling, there are still some limitations. For example, the Demeter does not model skinning, so each part has uniform thickness and the connections between parts are unnatural. 

Additionally, we have only demonstrated very basic capabilities for sample generation, such as copying the sub-tree from existing soybeans and pasting to random position and changing its topology. We believe that learning the distribution of the plant graph in latent space can better handle this task in the future.
{
    \small
    \bibliographystyle{plainnat}
    \bibliography{main}
}

\end{document}